\documentclass[10pt,twocolumn,letterpaper]{article}

\usepackage{cvpr}              % To produce the CAMERA-READY version
\definecolor{cvprblue}{rgb}{0.21,0.49,0.74}
\usepackage[pagebackref,breaklinks,colorlinks,allcolors=cvprblue]{hyperref}

\usepackage{algorithm}
\usepackage{algorithmicx}
\usepackage{algpseudocode}
\usepackage{bbm}
\usepackage{tcolorbox}
\usepackage{booktabs}
\usepackage{tabularx}
\usepackage{caption}
\usepackage[shortlabels,inline]{enumitem}

\usepackage{wrapfig}
\usepackage{subcaption}

\usepackage{todonotes}
\usepackage{colortbl}
\usepackage{microtype}      % microtypography
\usepackage{xcolor}         % colors
\usepackage{makecell}       % For multi-line headers
\usepackage{multirow}
\usepackage{threeparttable}
\usepackage{hyperref}
\usepackage{url}

\makeatletter
\renewcommand\@makefntext[1]{%
  \noindent\makebox[1in][r]{\@makefnmark\,}#1}
\makeatother
\def\paperID{11875} % *** Enter the Paper ID here
\def\confName{CVPR}
\def\confYear{2026}

\title{Time Blindness: Why Video-Language Models Can’t See What Humans Can?}

\author{Ujjwal Upadhyay$^{1,3,}$\thanks{\footnotesize Equal contribution; Author ordering determined by a coin toss.}~, 
~~ Mukul Ranjan$^{2,*}$, 
~~ Zhiqiang Shen$^{2,}$\thanks{Corresponding authors} ~, ~~  Mohamed Elhoseiny$^{1,\dagger}$ \\
$^1$KAUST~~ $^{2}$VILA Lab, MBZUAI ~~ $^{3}$DocPanel Technologies\\
\small Project page: \url{https://timeblindness.github.io/} 
}

\begin{document}
\maketitle

\begin{abstract}
    Recent advances in vision-language models (VLMs) have made impressive strides in understanding spatio-temporal relationships in videos. However, when spatial information is obscured, these models struggle to capture purely temporal patterns. We introduce \textbf{SpookyBench}, a benchmark where information is encoded solely in temporal sequences of noise-like frames, mirroring natural phenomena from biological signaling to covert communication. Interestingly, while humans can recognize shapes, text, and patterns in these sequences with over 98\% accuracy, state-of-the-art VLMs achieve 0\% accuracy. This highlights a critical limitation: an over-reliance on frame-level spatial features and an inability to extract meaning from temporal cues. Overcoming this limitation will require novel architectures or training paradigms that decouple spatial dependencies from temporal processing. Our systematic analysis shows that this issue persists across model scales and architectures. We release SpookyBench to catalyze research in temporal pattern recognition and bridge the gap between human and machine video understanding. Dataset and code are available on the project website. 
    % We also provide examples and analysis based on motion segmentation using optimal flow which can be easily detected by these VLMS.
    % Dataset is available at this anonymous link: \href{https://tinyurl.com/spooky-bench}{\textcolor{magenta}{https://tinyurl.com/spooky-bench}}
\end{abstract}  
% \vspace{-0.28in}
\section{Introduction}
Large multimodal models have revolutionized visual understanding in both images~\citep{liu2023llavavisual, wang2024qwen2vl, bai2025qwen2.5vl, chen2024intern2vl, deitke2024molmo, dai2024nvlm} and videos~\citep{zhang2024llavavideo, maaz2023videochatgpt, ataallah2024minigpt4video, weng2024longvlm, wang2025internvideo2}. Recent Video-Vision Language Models (Video-VLMs) demonstrate impressive capabilities across action recognition~\citep{wu2023revisiting, kahatapitiya2024victr, zhao2023learning}, visual question answering~\citep{yu2023self, min2024morevqa, zhong2022video, ayyubi2025enter, park2024too}, dense captioning~\citep{qasim2023dense, yang2023vid2seq, xu2024pllava, kim2024you, chen2024panda, chen2025vidcapbench, chen2024sharegpt4video}, and temporal grounding~\citep{chen2024timemarker, wang2024grounded, xu2024vtg}. Yet even in tasks labeled as temporal, strong per-frame spatial cues often allow models to succeed without genuinely reasoning over time~\citep{cores2024tvbench, cai2024temporalbench, li2024mvbench}.
\noindent This paper introduces \textbf{\texttt{SpookyBench}}, a benchmark that closes this loophole. Every stimulus is constructed so that individual frames appear as noise; the signal exists only in the temporal relationship between frames. Existing temporal benchmarks~\citep{cai2024temporalbench, li2024vitatecs, yang2025svbench, li2024vidhalluc} still entangle spatial and temporal cues, making it difficult to diagnose where a model's understanding breaks down. \textbf{\texttt{SpookyBench}} eliminates this confound entirely. In spirit it parallels ARC-AGI~\citep{chollet2026arcprize2025technical}, which uses synthetic puzzles to isolate core reasoning; here, controlled stimuli isolate a single fundamental capability, extracting meaning from change over time.
\begin{figure*}
    \centering
    \includegraphics[width=0.9\linewidth]{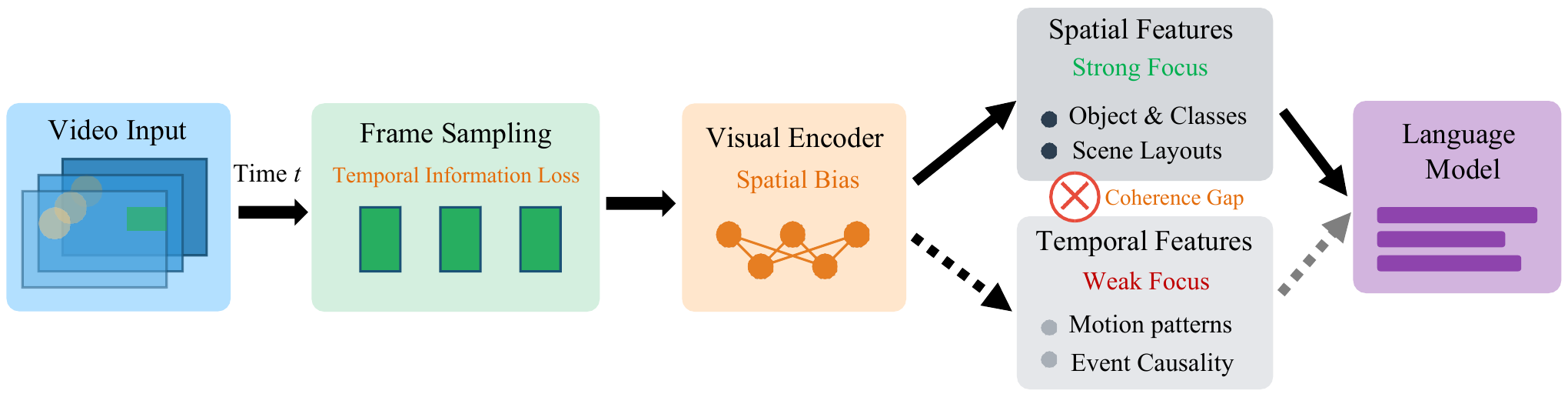}
    \vspace{-0.10in}
    \caption{Illustration of the current video-language models' limitations: over reliance on spatial visual features within individual frame. Frame sampling results in temporal information loss, while the visual encoder exhibits a strong spatial bias. This creates a coherence gap ($\times$) between well-represented spatial features (objects, scene layouts) and under-represented temporal features (motion patterns, causality), limiting video understanding capabilities.} 
    \label{fig:spatial-reliance-vlm}
    \vspace{-0.2in}
\end{figure*}

Current approaches to video understanding~\citep{tang2023video, nguyen2024video} typically follow a hierarchical paradigm: extract frame-level features using ViTs~\citep{bertasius2021space, radford2021learning, dosovitskiy2020image}, integrate these features temporally, and fuse them with language for downstream tasks~\citep{zhang2024mm1.5, li2024llavanextinterleave, wu2024janus, wang2024emu3}. This paradigm has yielded significant advances in general video understanding~\citep{li2024llavaonevision, dubey2024llama, tang2023video, nguyen2024video}. However, our findings reveal a critical blind spot: when information exists purely in the temporal domain without reliable frame-level features, state-of-the-art models fail catastrophically (Figure \ref{fig:spatial-reliance-vlm}). The inability to decode temporal patterns has significant implications for real-world applications. In nature, organisms such as fireflies communicate through precise temporal sequences of bioluminescence~\citep{carlson1985flash, owens2022behavioral, ramirez2018fireflies}, encoding information exclusively through timing rather than spatial arrangements. Similarly, in medical domains, pre-ictal EEG patterns for epileptic seizure prediction emerge only through temporal analysis~\citep{slimen2020epileptic}. These natural examples demonstrate how temporal patterns can carry rich information even when individual observations contain minimal static content. Similarly, various human technologies from Morse code to digital communication protocols rely on temporal encoding, yet current Video-VLMs lack the fundamental mechanisms to process such information.

The human visual system has evolved mechanisms for processing temporal information without relying solely on spatial cues~\citep{tangemann2024object,johansson1973visual,arstila2016theories}. Neuroscience research has revealed that temporal processing is distributed across neural structures rather than centralized in a single area~\citep{mauk2004neural}, and the brain uses intrinsic network dynamics to perform temporal computations~\citep{paton2018neural}. Areas such as the parietal cortex integrate temporal information along with spatial and numeric magnitudes~\citep{bueti2009parietal}. Our experiments confirm humans' remarkable temporal perception: participants achieve over 98\% accuracy on \textbf{\texttt{SpookyBench}} tasks without training. In stark contrast, our evaluation of 15 state-of-the-art Video-VLMs, including closed-source commercial systems such as GPT-4o~\citep{hurst2024gpt4o}, and Gemini 2.0 Flash~\citep{deepmind2023gemini2.0flash}, reveals near-zero accuracy on these same tasks.

This performance gap persists across model architectures, parameter scales, and pre-training strategies, from compact systems like VideoLLaMA3-2B~\citep{zhang2025videollama} to large-scale ones such as GPT-4o~\citep{hurst2024gpt4o} and Qwen-VL~\citep{wang2024qwen2vl}, and extends to models designed specifically for video understanding including LongVLM~\citep{weng2024longvlm}, LLaVA-NeXT-Interleave~\citep{li2024llavanextinterleave}, and InternVideo2.5~\citep{wang2025internvideo2}. Efforts to enhance temporal reasoning through specialized temporal modeling~\citep{ren2024timechat, qian2024momentor, wang2024videollm} and fine-grained temporal localization~\citep{liu2024st, chen2024timemarker, wang2024grounded} have not addressed the challenge of extracting meaning from purely temporal patterns without reliable spatial features.

Our findings suggest that achieving human-like video understanding requires rethinking how neural architectures process temporal information. Rather than treating temporal integration as secondary to spatial feature extraction, future models may need dedicated mechanisms for temporal pattern recognition, drawing inspiration from cognitive neuroscience research on distributed neural timing mechanisms~\citep{paton2018neural, mauk2004neural} and specialized brain regions for temporal processing~\citep{bueti2009parietal, merchant2013neural}. The gap between human and machine performance on \textbf{\texttt{SpookyBench}} indicates that current architectures remain ``time-blind'' despite strong performance on standard benchmarks. We release SpookyBench to catalyze research into temporal reasoning in Video-VLMs, with applications ranging from medical diagnostics to autonomous systems that must interpret temporal cues in complex environments.

\vspace{-0.05in}
\section{Related Work} \label{sec:related}

\vspace{-0.05in} 
\subsection{Temporal Reasoning in Video-VLMs} 
\vspace{-0.05in} 
Video-VLMs have advanced through several architectural families, including LLaVA variants~\citep{liu2023llavavisual, zhang2024llavavideo, li2024llavanextinterleave, li2024llavaonevision, maaz2023videochatgpt, liu2024llavanext}, the Qwen series~\citep{wang2024qwen2vl, bai2025qwen2.5vl}, InternVL models~\citep{chen2024internvl, chen2024intern2vl, wang2025internvideo2}, dual encoders~\citep{maaz2024videogpt+}, interleaved tokens~\citep{ataallah2024minigpt4video, zhu2023minigpt4}, compression techniques~\citep{shen2024longvu}, and multimodal fusion~\citep{zhang2024mm1.5, wu2024janus, wang2024emu3}. All of these exhibit limited temporal reasoning, with hallucinations~\citep{li2024vidhalluc}, grounding failures~\citep{wang2024grounded}, and reliance on linguistic shortcuts~\citep{ko2023large} persisting across action recognition~\citep{wu2023revisiting, kahatapitiya2024victr, zhao2023learning}, question answering~\citep{yu2023self, min2024morevqa, ayyubi2025enter}, and captioning~\citep{yang2023vid2seq, kim2024you, chen2024sharegpt4video}. Targeted interventions such as timestamp-aware encoding~\citep{ren2024timechat}, segment-level reasoning~\citep{qian2024momentor}, direct token processing~\citep{liu2024st}, temporal separation tokens~\citep{chen2024timemarker}, specialized temporal streams~\citep{wang2024grounded, wang2024videollm}, and synthetic training paradigms~\citep{zhang2024llavavideo, yu2023self, tang2023video, nguyen2024video} have not closed this gap, and dedicated video architectures like VideoGPT+~\citep{maaz2024videogpt+}, TimeChat~\citep{ren2024timechat}, LinVT~\citep{gao2024linvt}, LongVLM~\citep{weng2024longvlm}, and Baichuan-Omni~\citep{li2024baichuan} extract spatial features first and treat temporal integration as an afterthought.  

These limitations are further corroborated by temporal understanding benchmarks; TemporalBench~\citep{cai2024temporalbench} reveals a significant gap between model and human performance, TVBench~\citep{cores2024tvbench}, VITATECS~\citep{li2024vitatecs}, and Fateh et al.~\citep{fateh2024video} confirm that many datasets reward spatial analysis over temporal reasoning, and targeted evaluations surface specific failures in temporal hallucinations~\citep{li2024vidhalluc}, streaming video reasoning~\citep{yang2025svbench}, and temporal location, object tracking, and anomaly detection~\citep{li2024videovista}. Across these analyses, models such as LLaVA-Video~\citep{zhang2024llavavideo}, Video-ChatGPT~\citep{maaz2023videochatgpt}, TemporalVLM~\citep{fateh2024video}, and VidChain~\citep{lee2025vidchain} exploit spatial shortcuts instead of temporal reasoning~\citep{wang2024grounded, chen2024timemarker, li2024vidhalluc, ko2023large}. SpookyBench addresses this by obscuring spatial information and forcing models to extract meaning from temporal dynamics alone, exposing a ``time-blindness'' that conventional benchmarks leave undetected.

\begin{table}[t]
\centering
\vspace{0.1in}
\resizebox{\linewidth}{!}{
\begin{tabular}{lccc}
\toprule
\rowcolor{gray!20}
\textbf{Model} & \textbf{Direct Prompt} & \textbf{CoT} & \textbf{Params} \\
\hline
\textbf{Human Performance} & \bf 98.0\% $\pm$ 0.6 & N/A & N/A \\
\hline
\multicolumn{4}{c}{\textbf{Open-Source Models}} \\
\hline
VideoLLaMA3-7B~\citep{zhang2025videollama} & 0\% $\pm$ 0.0 & 0\% $\pm$ 0.0 & 7B \\
VideoLLaMA3-2B~\citep{zhang2025videollama} & 0\% $\pm$ 0.0 & 0\% $\pm$ 0.0 & 2B \\
TimeChat-7B~\citep{ren2024timechat} & 0\% $\pm$ 0.0 & 0\% $\pm$ 0.0 & 7B \\
MiniGPT4-Video~\citep{ataallah2024minigpt4video} & 0\% $\pm$ 0.0 & 0\% $\pm$ 0.0 & 7B \\
MovieChat~\citep{song2024moviechat} & 0\% $\pm$ 0.0 & 0\% $\pm$ 0.0 & 7B \\
Video-ChatGPT-7B~\citep{maaz2023videochatgpt} & 0\% $\pm$ 0.0 & 0\% $\pm$ 0.0 & 7B \\
VideoGPT-plus-Phi3-mini-4k~\citep{maaz2024videogpt+} & 0\% $\pm$ 0.0 & 0\% $\pm$ 0.0 & 7B \\
VILA1.5-13b\cite{lin2023vila} & 0\% $\pm$ 0.0 & 0\% $\pm$ 0.0 & 13B \\
ShareGPT4Video-8B~\citep{chen2024sharegpt4video} & 0\% $\pm$ 0.0 & 0\% $\pm$ 0.0 & 8B \\
VideoLLaMA2-7B~\citep{cheng2024videollama} & 0\% $\pm$ 0.0 & 0\% $\pm$ 0.0 & 7B \\
Video-LLaVA~\citep{zhang2024llavavideo} & 0\% $\pm$ 0.0 & 0\% $\pm$ 0.0 & 7B \\
LLaVA-NeXT-Video~\citep{li2024llavanextinterleave} & 0\% $\pm$ 0.0 & 0\% $\pm$ 0.0 & 8B \\
\hline
InternVL2-40B~\citep{chen2024intern2vl} & 0\% $\pm$ 0.0 & 0\% $\pm$ 0.0 & 40B \\
InternVL2-8B~\citep{chen2024intern2vl} & 0\% $\pm$ 0.0 & 0\% $\pm$ 0.0 & 8B \\
InternVL2.5-78B~\citep{chen2024expanding2.5internvl} & 0\% $\pm$ 0.0 & 0\% $\pm$ 0.0 & 78B \\
InternVL2.5-8B~\citep{chen2024expanding2.5internvl} & 0\% $\pm$ 0.0 & 0\% $\pm$ 0.0 & 8B \\
InternVideo2.5-Chat-8B~\citep{wang2025internvideo2} & 0\% $\pm$ 0.0 & 0\% $\pm$ 0.0 & 8B \\
InternVideo2-Chat-8B ~\citep{wang2024internvideo2} & 0\% $\pm$ 0.0 & 0\% $\pm$ 0.0 & 8B \\
\hline
Qwen2-VL-2B-Instruct~\citep{wang2024qwen2vl} & 0\% $\pm$ 0.0 & 0\% $\pm$ 0.0 & 2B \\
Qwen2-VL-7B-Instruct~\citep{wang2024qwen2vl} & 0\% $\pm$ 0.0 & 0\% $\pm$ 0.0 & 7B \\
Qwen2-VL-72B-Instruct~\citep{wang2024qwen2vl} & 0\% $\pm$ 0.0 & 0\% $\pm$ 0.0 & 72B \\
Qwen2.5-VL-3B-Instruct~\citep{bai2025qwen2.5vl} & 0\% $\pm$ 0.0 & 0\% $\pm$ 0.0 & 3B \\
Qwen2.5-VL-7B-Instruct~\citep{bai2025qwen2.5vl} & 0\% $\pm$ 0.0 & 0\% $\pm$ 0.0 & 7B \\
Qwen2.5-VL-72B-Instruct~\citep{bai2025qwen2.5vl} & 0\% $\pm$ 0.0 & 0\% $\pm$ 0.0 & 72B \\
Qwen3-VL-8B-Instruct~\citep{qwen3technicalreport} & 0\% $\pm$ 0.0 & 0\% $\pm$ 0.0 & 8B \\
\hline
\multicolumn{4}{c}{\textbf{Closed-Source Models}} \\
\hline
Gemini 2.5 Pro~\citep{comanici2025gemini} & 0\% $\pm$ 0.0 & 0\% $\pm$ 0.0 & N/A \\
Gemini 1.5 Pro~\citep{team2024gemini1.5pro} & 0\% $\pm$ 0.0 & 0\% $\pm$ 0.0 & N/A \\
Gemini 2.0 Flash\cite{deepmind2023gemini2.0flash} & 0\% $\pm$ 0.0 & 0\% $\pm$ 0.0 & N/A \\
GPT-4o~\citep{hurst2024gpt4o} & 0\% $\pm$ 0.0 & 0\% $\pm$ 0.0 & N/A \\
% GPT-5-mini~\citep{OpenAI2025GPT5SystemCard} & 0\% $\pm$ 0.0 & 0\% $\pm$ 0.0 & N/A \\
% GPT-5~\citep{OpenAI2025GPT5SystemCard} & 0\% $\pm$ 0.0 & 0\% $\pm$ 0.0 & N/A \\
\bottomrule
\end{tabular}
}
% \vspace{-0.1in}
\caption{Benchmark results comparing model performance on \textbf{\texttt{SpookyBench}} across different prompting strategies along with model size. Human accuracy (98.0\%) is the weighted average of accuracy across 3 different categories.}
\label{tab:model_performance}
\vspace{-0.2in}
\end{table}

\vspace{-0.08in} \subsection{Neuroscience Insights on Temporal Processing} 
\vspace{-0.05in} 
Human perception of temporally-defined objects relies on the Gestalt principle of common fate~\citep{wertheimer2012perceived}, whereby coherently moving elements are grouped together, enabling figure-ground segregation from motion alone~\citep{johansson1973visual}. Tangemann et al.~\citep{tangemann2024object} showed that a cortical motion energy model~\citep{simoncelli1998model} matches human performance on random dot segmentation while 40 deep optical flow models fail, and relatedly, recent works~\citep{burgert2025go, chang2025warped} encode motion into structured noise for video diffusion control.  Beyond these perceptual grouping mechanisms, neuroscience research offers insights for addressing temporal limitations in Video-VLMs. Mauk and Buonomano~\citep{mauk2004neural} established that temporal processing is distributed across neural structures through intrinsic circuit properties, contrasting with current Video-VLMs' sequential spatial processing. The human brain processes time at multiple granularities, with the cerebellum handling millisecond-to-second timing~\citep{merchant2013neural}, the parietal cortex integrating temporal, spatial and numerical magnitudes~\citep{bueti2009parietal}, and neural patterns encoding time through ``population clocks''~\citep{paton2018neural}. Distributed temporal representations that evolve over time~\citep{wittmann2013neural, paton2018neural} offer an alternative to treating temporal integration as secondary. The performance gap on temporal tasks~\citep{cai2024temporalbench, cores2024tvbench, li2024vidhalluc} and our SpookyBench findings demonstrate that current architectures lack mechanisms for processing purely temporal patterns, a natural capability in humans through neural systems representing time as intrinsic dynamics.

% \begin{algorithm}[t]
% \caption{Content Mask Animation}
% \label{alg:mask_animation}
% \begin{algorithmic}[1]
% \State \textbf{Input:} Content mask $M$, velocity $v$
% \State \textbf{Output:} Animated frame $F_t$

% \State Generate noise patterns $N_{bg}$, $N_{fg}$
% \For{each pixel $(x,y)$}
%     \If{$M(x,y)$}
%         \State $F_t(x,y) \gets N_{fg}(x,y+vt \bmod h)$ \Comment{Foreground}
%     \Else
%         \State $F_t(x,y) \gets N_{bg}(x,y-vt \bmod h)$ \Comment{Background}
%     \EndIf
% \EndFor
% \end{algorithmic}
% \end{algorithm}

% \begin{algorithm}[t]
% \caption{Video Depth Map Animation}
% \label{alg:video_depth}
% \begin{algorithmic}[1]
% \State \textbf{Input:} Depth map $D$, thresholds $(t_l, t_u)$, velocity $v$
% \State \textbf{Output:} Animated frame $F_t$

% \State Generate noise pattern $N$
% \For{each pixel $(x,y)$}
%     \State $d \gets$ brightness from depth map $D(x,y)$
%     \If{$t_l \leq d \leq t_u$}
%         \State $F_t(x,y) \gets N(x,y+vt \bmod h)$ \Comment{Moving noise}
%     \Else
%         \State $F_t(x,y) \gets N(x,y)$ \Comment{Static noise}
%     \EndIf
% \EndFor
% \end{algorithmic}
% \end{algorithm}
\vspace{-0.05in}
\section{SpookyBench}
\vspace{-0.05in}
We introduce \textbf{\texttt{SpookyBench}}, a novel synthetic dataset specifically designed to isolate and evaluate pure temporal understanding in video language models. The key innovation of our benchmark lies in its unique design: All meaningful information is encoded exclusively in the temporal domain through dynamic patterns of texts, images and video depth maps, while individual frames contain only structured noise. Our dataset is fundamentally different from the existing datasets used for training, fine-tuning, and evaluation of video-VLMs. Many state-of-the-art video language models employ advanced techniques, such as dynamic resolution strategies~\citep{bai2025qwen2.5vl, wang2024qwen2vl, chen2024intern2vl}, specialized temporal encoding methods~\citep{ren2024timechat, wang2024qwen2vl, bai2025qwen2.5vl}, hierarchical token merging~\citep{weng2024longvlm, wang2025internvideo2}, and joint video-motion training frameworks~\citep{chen2024motionllm} to capture temporal dynamics. However, these methods still rely on spatial representations extracted from individual frames, which currently remain the only viable mechanism for inferring temporal information. In contrast, \textbf{\texttt{SpookyBench}} forces models to depend only on temporal cues, thereby creating the first benchmark that exclusively evaluates a model's ability to process and understand pure temporal information.
%%%%%%%%%%%%%%%%%%%%%%%%%%
\begin{figure*}[!htbp]
    \centering
    \includegraphics[width=\textwidth]{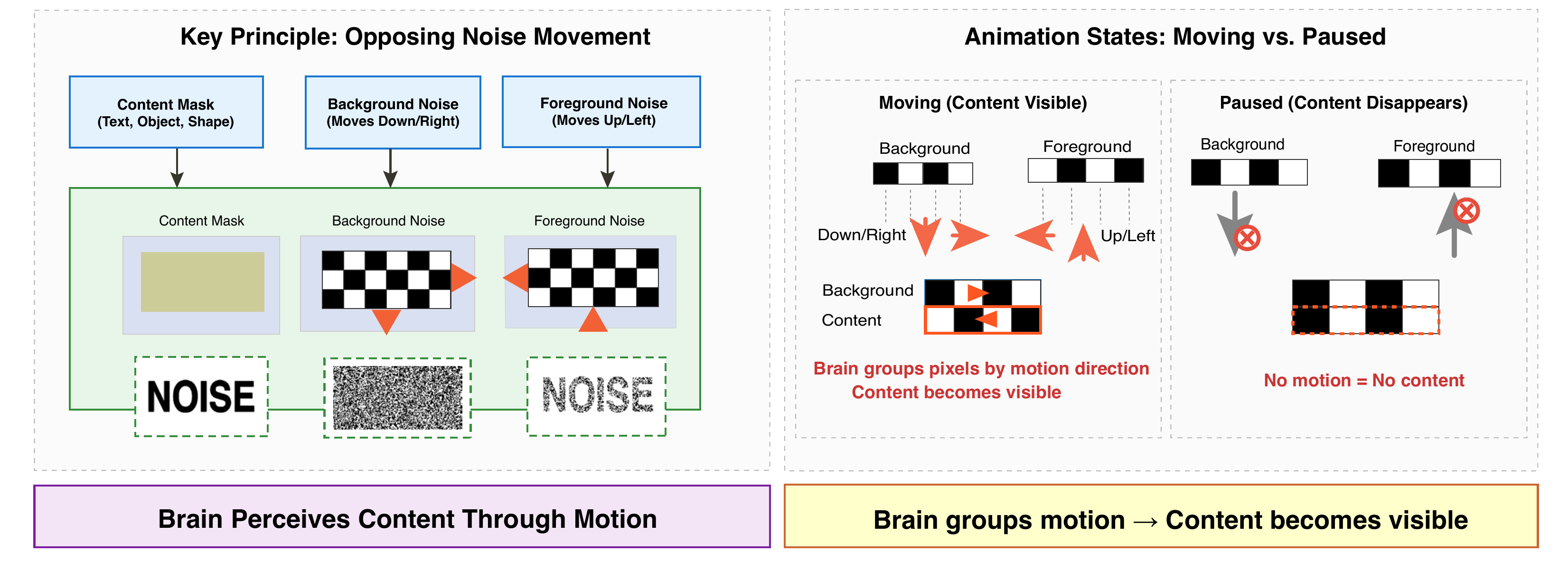}
    \caption{Illustration of the temporal encoding framework used in \texttt{SpookyBench}. \textbf{Left:} Core mechanism showing how content becomes visible through opposing motion patterns. A content mask defines regions where foreground noise (moving up/left) and background noise (moving down/right) are applied. When animated, the human visual system groups pixels with similar motion, causing the content to emerge. \textbf{Right:} Comparison between moving and paused states, demonstrating how content is only perceptible during animation and disappears when static, as individual frames contain only structured noise.}

    \label{fig:dataset_generation}
\end{figure*}

%%%%%%%%%%%%%%%%%%%%%%%%%%
% \vspace{-0.05in}
\subsection{Dataset Generation}
\begin{figure}
    \centering
    \includegraphics[width=0.90\linewidth]{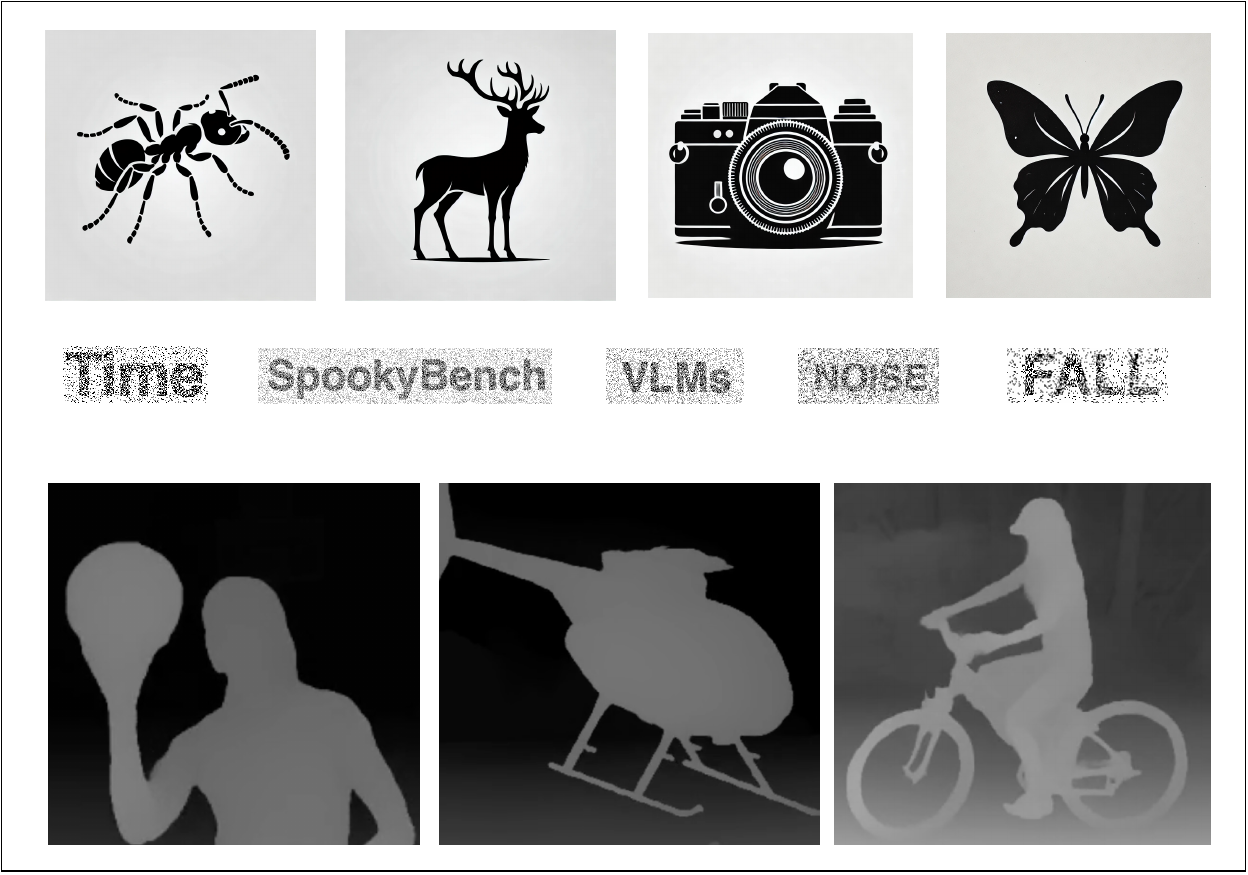}
    \caption{Noise generation process: (Top) masks applied for dynamic noise video generation, (Mid) word-specific mask, and (Bottom) depth map of video frame used for constructing noise-overlaid stimulus.}
    \label{fig:categories}
\end{figure}

Figure \ref{fig:dataset_generation} shows our proposed data generation framework. The dataset consists of specially designed videos that encode three types of content - words, images, and videos - using binary noise patterns with specific motion properties. In this approach, content is embedded within noise patterns such that individual frames appear as random noise, while the content becomes perceptible only when viewed as a temporal sequence. Our dataset encodes different types of content (Figure \ref{fig:categories}) through temporal noise animations in the following categories:
\textbf{1) Words:} Text rendered as masks in which the background noise and foreground noise move in opposite directions, making the text visible only through temporal movement.
% \item \textbf{Shapes:} Basic geometric patterns (rectangles, circles, polygons) encoded using the same opposing motion technique as text.
\textbf{2)Images:} Binary masks generated using SAM2~\citep{ravi2024sam} from single-object images generated using text-to-image model Flux~\citep{flux2024}, encoded using the same content mask animation approach as words.
\textbf{3) Dynamic Scenes:} Depth maps extracted from videos in single-object tracking datasets LaSOT~\citep{fan2019lasot} and OTB2015~\citep{otb2015yi} using Video Depth Anything~\citep{chen2025videodepth}. These are encoded using a technique in which pixels above a brightness threshold move while others remain static as shown in the algorithm \ref{alg:video_depth}.

\subsection{Temporal Encoding Framework}
\label{subsec:temporal-encoding-framework}

Our temporal encoding framework implements two distinct motion configurations as detailed in Algorithms~\ref{alg:mask_animation} and \ref{alg:video_depth}.
For words, and image masks (Algorithm~\ref{alg:mask_animation}), we employ opposing motion patterns between foreground and background. The content is first converted to a binary mask $M$ where $M(x,y) = 1$ represents foreground pixels and $M(x,y) = 0$ represents background. We generate two separate noise patterns $N_{bg}$ and $N_{fg}$ consisting of random binary values (0 or 255). During animation, foreground pixels sample from $N_{fg}$ with a positive offset that increases with time ($y+vt \bmod h$), while background pixels sample from $N_{bg}$ with a negative offset ($y-vt \bmod h$). 
% \begin{table}{r}{0.65\textwidth}
\begin{table}[t]
\centering
\resizebox{\linewidth}{!}{
\begin{tabular}{lccccc}
\toprule
\textbf{Category} & \textbf{Basic SNR (dB)} & \textbf{Perceptual SNR} & \textbf{Temporal Coherence} & \textbf{Motion Contrast} \\
\hline
Images & -46.95 $\pm$ 2.40 & -47.28 $\pm$ 2.28 & 8.00 $\pm$ 2.08 & 7.17 $\pm$ 5.00 \\
Dynamic Scenes & -48.95 $\pm$ 3.64 & -63.43 $\pm$ 5.74 & 21.91 $\pm$ 5.76 & -3.18 $\pm$ 10.17 \\
Text & -39.27 $\pm$ 1.58 & -49.18 $\pm$ 3.31 & 7.84 $\pm$ 0.65 & 8.26 $\pm$ 6.44 \\
\bottomrule
\end{tabular}
}
% \vspace{-0.1in}
\caption{Signal-to-Noise Ratio (SNR) metrics across SpookyBench categories.}
\label{tab:snr-metrics}
\end{table}
% \vspace{-0.1in}
This creates the perception of opposing motion within and outside the masked regions. For video depth maps (Algorithm~\ref{alg:video_depth}), we employ a threshold-based approach. Using depth maps $D$ extracted from videos, pixels with brightness values between lower and upper thresholds ($t_l \leq d \leq t_u$) are animated by sampling a noise pattern $N$ with a time-varying offset ($y+vt \bmod h$), while pixels outside this range remain static. This creates the illusion that brighter regions (typically foreground objects) 
are moving while darker regions (typically background) remain static. The noise patterns are generated using binary values (0 or 255) in square blocks of variable size. We used different speckle sizes ranging from $1\times1$ to $3\times3$ pixels to investigate the effect of noise granularity on perception. For each speckle size, we also varied the noise density - the probability that a block is white versus black - using values of 10\%, 30\%, 50\%, and 90\%. These noise patterns arranged in pixel blocks create optimal perceptual conditions for human viewers while remaining challenging for vision language models. To ensure seamless animation, the noise patterns are made tileable by copying edge pixels to the opposite boundaries. 
All videos maintain consistent technical specifications: $960\times540$ pixel resolution, with an average duration of 7.11 seconds (ranging from 1.0 to 35.0 seconds) and an average of 333.5 frames per video. Text videos have a consistent duration of around 4 seconds; however, videos of dynamic scenes are longer, ranging up to 35 seconds.
Figure~\ref{fig:dataset_generation} illustrates the structure of the data set and the encoding patterns in categories. 
 We used binary masks for the images using SAM2~\citep{ravi2024sam}. For videos, depth maps are extracted using Depth Anything V2~\citep{yang2025depth} and Video Depth Anything~\citep{chen2025videodepth} from the LaSOT~\citep{fan2019lasot} and OTB2015~\citep{otb2015yi} datasets.
\vspace{-0.05in}
\subsection{Data Statistics}
\label{sec:data-stats}
\vspace{-0.05in}
% \begin{wrapfigure}{r}{0.4\textwidth} % {placement}{width}
%     \centering
%     \includegraphics[width=\linewidth]{assets/spookybench_categories.pdf}
%     \caption{Distribution of the \textbf{SpookyBench} dataset across three video categories.}
%     \label{fig:spookybench-distribution}
% \end{figure}
SpookyBench comprises 451 videos in three distinct categories, each requiring purely temporal reasoning for content identification. 
The dataset is distributed as follows: Text (46.6\%, 210 videos), Object Images (40.8\%, 184 videos) and Dynamic Scenes (12.6\%, 57 videos). Although Dynamic Scenes comprise only 12.6\% of videos, they account for 42.5\% of total frames due to their longer duration, balancing the frame-level distribution across categories. This distribution ensures comprehensive coverage of different temporal perception challenges while maintaining a natural frequency distribution that reflects real-world scenarios. Additionally, more dataset can be generated indefinitely through the data generator on our project page, thus the dataset size is essentially unlimited. 
The ``Text'' category contains common English words rendered through temporal noise patterns, enabling evaluation of models' ability to identify linguistic content through purely temporal cues. The ``Object Images'' category presents single objects extracted from high-quality images using segmentation techniques~\citep{ravi2024sam}, encoded with the same temporal animation approach. It also contains a synthetic silhouette of simple objects generated using DALL-E 3~\citep{betker2023dalle3} and flux~\citep{flux2024}.  
% Figure \ref{fig:spookybench-distribution} shows the data distribution of SpookyBench. 
% \label{sec:stats}

% \begin{figure}[htbp]
%     \centering
%     \includegraphics[width=0.7\linewidth]{assets/spookybench_categories.pdf}
%     \caption{Distribution of the \textbf{SpookyBench} dataset across three video categories. Each category represents a different type of content encoded through temporal noise patterns: \textit{Text}, \textit{Object Images}, and \textit{Dynamic Scenes}.}
%     \label{fig:spookybench-distribution}
% \end{figure}
\subsubsection{Analysis of Temporal Metrics}
To ensure a rigorous quantification of the temporal information present in each video, we analyzed five key  \ref{tab:snr-metrics}. SNR metrics that capture different aspects of the complexity and perceptibility of temporal patterns in SpookyBench, as shown in Table.
These metrics provide insight into why temporal patterns might be visible to humans but challenging to detect by computational models.

% \vspace{-0.2in}
% \begin{table}[ht]
% \centering
% \resizebox{0.98\linewidth}{!}{
% \begin{tabular}{lccccc}
% \toprule
% \textbf{Category} & \textbf{Basic SNR (dB)} & \textbf{Perceptual SNR} & \textbf{Temporal Coherence} & \textbf{Motion Contrast} \\
% \hline
% Object Images & -46.53 $\pm$ 2.81 & -46.97 $\pm$ 2.67 & 8.16 $\pm$ 2.44 & 8.85 $\pm$ 5.87 \\
% Shapes & -49.27 $\pm$ 1.88 & -49.00 $\pm$ 2.02 & 7.10 $\pm$ 1.48 & -2.20 $\pm$ 3.27 \\
% Dynamic Scenes & -48.95 $\pm$ 3.64 & -63.43 $\pm$ 5.74 & 21.91 $\pm$ 5.76 & -3.18 $\pm$ 10.17 \\
% Text & -39.27 $\pm$ 1.58 & -49.18 $\pm$ 3.31 & 7.84 $\pm$ 0.65 & 8.26 $\pm$ 6.44 \\
% \bottomrule
% \end{tabular}
% }
% \caption{Signal-to-Noise Ratio (SNR) metrics across SpookyBench categories.}
% \label{tab:snr-metrics}
% \end{table}
\vspace{-0.05in}
\paragraph{Basic SNR} measures signal-to-noise ratio in decibels:
\begin{equation}
\text{SNR}_B = 10 \log_{10}\left(\frac{P_S}{P_N}\right)
\end{equation}
where $P_S = \mathbb{E}[\lVert \nabla \mathbf{F} \rVert^2]$ is motion boundary energy derived from spatial gradients of optical flow field $\mathbf{F}(x,y) = (F_x, F_y)$, and $P_N = \text{Var}(I_0)$ is variance of the static frame $I_0$.

\noindent\textbf{Perceptual SNR} incorporates frequency-dependent visual sensitivity:
\begin{equation}
\text{SNR}_P = 10 \log_{10}\left(\frac{\lVert \mathcal{H}(B) \odot W \rVert^2}{\lVert \mathcal{H}(N) \odot W \rVert^2}\right)
\end{equation}
where $B$ is the average motion boundary strength, $N$ is the static noise frame, $\mathcal{H}$ is the 2D Fourier transform, $\odot$ denotes element-wise multiplication, and $W(f) = f \cdot e^{-f/f_0}$ is the contrast sensitivity weighting function with peak $f_0 \approx 0.1$ cycles/pixel.

\noindent\textbf{Temporal Coherence SNR} quantifies motion consistency:
\begin{equation}
\text{SNR}_T = 10 \log_{10}\left(\frac{\text{Var}(C)}{\mathbb{E}[\text{Var}_{\text{local}}(C)]}\right)
\end{equation}
where $C = e^{-\text{Var}_{\theta}(\mathbf{F})} \cdot \mathbbm{1}(\lVert \mathbf{F} \rVert > \tau)$ is the directional coherence map, $\text{Var}_{\theta}$ computes circular variance of flow direction angles over time, $\mathbbm{1}$ is indicator function, $\tau$ is magnitude threshold, and $\text{Var}_{\text{local}}$ computes variance over small spatial neighborhoods.

\noindent\textbf{Motion Contrast SNR} measures foreground-background motion differentiation:
\begin{equation}
\text{SNR}_M = 10 \log_{10}\left(\frac{\lVert \boldsymbol{\mu}_M - \boldsymbol{\mu}_B \rVert^2}{\frac{1}{2}(\sigma^2_M + \sigma^2_B)}\right)
\end{equation}
where $\boldsymbol{\mu}_M = \mathbb{E}[\mathbf{F} \mid M]$ and $\boldsymbol{\mu}_B = \mathbb{E}[\mathbf{F} \mid \neg M]$ are mean flow vectors within mask region $M$ and background region $\neg M$ respectively, $\sigma^2_M = \mathbb{E}[\lVert \mathbf{F} - \boldsymbol{\mu}_M \rVert^2 \mid M]$ and $\sigma^2_B = \mathbb{E}[\lVert \mathbf{F} - \boldsymbol{\mu}_B \rVert^2 \mid \neg M]$ are corresponding motion variances. The mask $M$ is estimated from the motion boundaries.

The distribution of these metrics reveals why current vision models struggle with \textbf{\texttt{SpookyBench}}: they lack mechanisms to leverage temporal coherence (particularly high in Dynamic Scenes, 21.91 $\pm$ 5.76 dB) and motion contrast (negative for Dynamic Scenes, -2.20 and -3.18 dB), while text stimuli benefit from higher basic SNR (-39.27 $\pm$ 1.58 dB), explaining the observed performance gap.

\subsubsection{Binary SNR Threshold Effect in Detection}
\begin{figure}
    \centering
    \includegraphics[width=\linewidth]{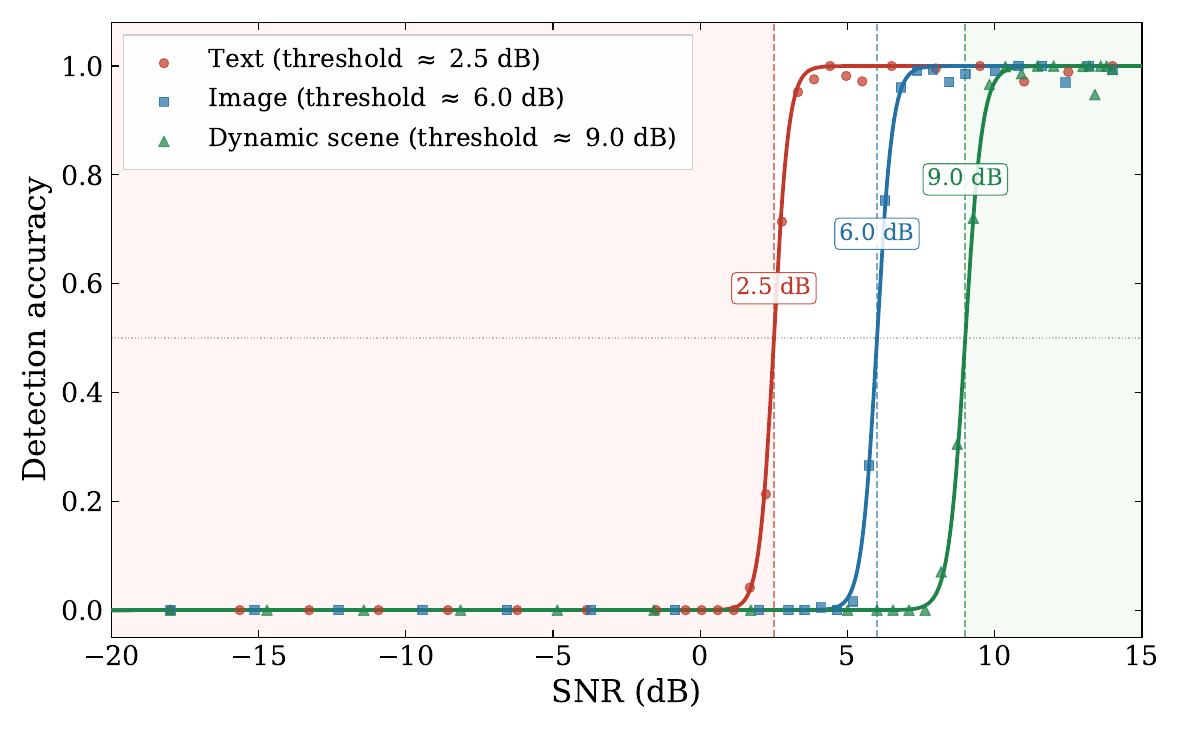}
    \vspace{-0.25in}
    \caption{Analysis of effects of SNR on detecting text, image and dynamic scenes with direct prompting and chain of thought prompting.}
    \label{fig:snr-vs-performance}
    \vspace{-0.2in}
\end{figure}

Our analysis shows a binary threshold phenomenon in detecting text within dynamic noise videos. The words shows negligible detection ($\sim$0\%) below 2.5dB SNR, but jumped to 85.7\% accuracy above this threshold, displaying an abrupt rather than gradual transition as show in figure \ref{fig:snr-vs-performance}. Similar threshold behavior appears for images (inflection at 6.0\,dB) and dynamic scenes (9.0\,dB). 
Prompts performed best (40\% accuracy), with Chain-of-Thought reasoning improving general identification tasks compared to direct prompting. This phenomenon parallels medical imaging diagnostics, where pathologies like microcalcifications in mammography become either entirely visible or invisible based on specific SNR thresholds~\citep{sahiner2012computer}. The implications are significant: unlike perceptual phenomena that degrade gradually with noise, text detection functions as a step function, creating vulnerabilities in safety-critical applications. Just as radiologists cannot diagnose what remains invisible, language models cannot identify text below certain noise thresholds, leading to false certainties and potential catastrophic performance drops with minimal noise increases. This characteristic creates particular concerns for autonomous vehicles reading road signs or medical systems interpreting labels, while also exposing systems to adversarial attacks where slight SNR manipulations could render text completely undetectable.
% \noindent\textbf{Temporal Coherence SNR} quantifies motion consistency:
% \begin{equation}
% \text{SNR}_T = 10 \log_{10}\left(\frac{\text{Var}(C)}{\mathbb{E}[\text{Var}_{\text{local}}(C)]}\right)
% \end{equation}
% where $C = e^{-\text{Var}_{\theta}(\mathbf{F})} \cdot \mathbbm{1}(\lVert \mathbf{F} \rVert > \tau)$ is the directional coherence map, $\text{Var}_{\theta}$ computes circular variance of flow direction angles over time, $\mathbbm{1}$ is indicator function, $\tau$ is magnitude threshold, and $\text{Var}_{\text{local}}$ computes variance over small spatial neighborhoods.
\begin{figure*}[t]
    \begin{minipage}[t]{0.48\textwidth}
        \begin{algorithm}[H] % Use [H] to keep the algorithm within the minipage
        \caption{Content Mask Animation}
        \label{alg:mask_animation}
        \begin{algorithmic}[1]
        \State \textbf{Input:} Content mask $M$, velocity $v$
        \State \textbf{Output:} Animated frame $F_t$

        \State Generate noise patterns $N_{bg}$, $N_{fg}$
        \For{each pixel $(x,y)$}
            \Statex \Comment{Check pixel's mask status}
            \If{$M(x,y)$}
                \State $F_t(x,y) \gets N_{fg}(x,y+vt \bmod h)$
                \Statex \Comment{Foreground}
            \Else
                \State $F_t(x,y) \gets N_{bg}(x,y-vt \bmod h)$
                \Statex \Comment{Background}
            \EndIf
        \EndFor
        \end{algorithmic}
        \end{algorithm}
    \end{minipage}
    \hfill % This acts as a flexible space between the two minipages
    \begin{minipage}[t]{0.48\textwidth}
        \begin{algorithm}[H] % Use [H] to keep the algorithm within the minipage
        \caption{Video Depth Map Animation}
        \label{alg:video_depth}
        \begin{algorithmic}[1]
        \State \textbf{Input:} Depth map $D$, thresholds $(t_l, t_u)$, velocity $v$
        \State \textbf{Output:} Animated frame $F_t$

        \State Generate noise pattern $N$
        \For{each pixel $(x,y)$}
            \State $d \gets$ brightness from $D(x,y)$
            \If{$t_l \leq d \leq t_u$}
                \State $F_t(x,y) \gets N(x,y+vt \bmod h)$
                \Statex \Comment{Moving noise}
            \Else
                \State $F_t(x,y) \gets N(x,y)$ \Statex \Comment{Static noise}
            \EndIf
        \EndFor
        \end{algorithmic}
        \end{algorithm}
    \end{minipage}
\end{figure*}
\vspace{-0.05cm}
\section{Experiments} \label{sec:experiment}
\vspace{-0.05in}
\subsection{Experimental Setup}

\noindent{\bf Models.} We evaluate \textbf{\texttt{SpookyBench}} on both open source models (Video-LLaVA~\citep{zhang2024llavavideo}, LLaVA-NeXT-Video~\citep{li2024llavanextinterleave}, TimeChat~\citep{ren2024timechat}, InternVL2~\citep{chen2024intern2vl}, Qwen2-VL~\citep{wang2024qwen2vl}, 
Qwen2.5-VL~\citep{bai2025qwen2.5vl} etc.) and closed source models (GPT-4o~\citep{hurst2024gpt4o}, Gemini 2.0 Flash~\citep{deepmind2023gemini2.0flash}, and Gemini 1.5 Pro~\citep{team2024gemini1.5pro}.
We design different prompts for each category. All the prompts are included in the Appendix. All prompts instruct models to respond with only 1-5 words identifying the content. We input sequences of multiple video frames simultaneously for models that do not directly support video input.

% \noindent{\bf Setup.} We evaluate model performance using exact match accuracy between model responses and our labels. For the Text categories, each video has a single correct label $y_i$. For {\em Object Images} and {\em Dynamic Scenes} categories, we define a set of acceptable labels $Y_i = \{y_{i1}, y_{i2}, \ldots, y_{in}\}$ to account for semantic ambiguity. For example, a video showing ``a man playing basketball'' accepts responses such as ``playing basketball'', ``man'', ``human'', or ``woman playing basketball'' as correct.
% Formally, for each video $i$, given a model response $r_i$ and corresponding label or set of labels $L_i$ (where $L_i = y_i$ for Text or $L_i = Y_i$ for objects and dynamic scenes), we calculate the accuracy as: $\text{Accuracy} = \frac{1}{N} \sum_{i=1}^{N} \mathbbm{1}(r_i \in L_i)$, 
% where $\mathbbm{1}$ is the indicator function that equals 1 if $r_i \in L_i$ and 0 otherwise, and $N$ is the total number of videos in the evaluation set. We also reviewed all the model's response manually. Despite this flexible evaluation protocol that accepts multiple valid responses for certain categories, none of the models tested produced responses that matched any of the acceptable options.
\noindent{\bf Setup.} We evaluate model performance using exact match accuracy. For Text, each video has a single correct label $y_i$. For {\em Object Images} and {\em Dynamic Scenes}, we define a set of acceptable labels $Y_i = \{y_{i1}, y_{i2}, \ldots, y_{in}\}$ to account for semantic ambiguity (e.g., a video showing ``a man playing basketball'' accepts ``playing basketball'', ``man'', ``human'', or ``woman playing basketball''). Accuracy is computed as $\text{Accuracy} = \frac{1}{N} \sum_{i=1}^{N} \mathbbm{1}(r_i \in L_i)$, where $L_i = y_i$ for Text and $L_i = Y_i$ for the other categories. We also verified all responses manually and with LLM-as-a-judge evaluation, a standard practice in VLM evaluation pipelines~\citep{duan2024vlmevalkit}; both yield identical 0\% performance (details in Appendix). Despite this flexible protocol, no model produced responses matching any acceptable option.

\vspace{-0.05in}
\subsection{Human Evaluation} \label{subsec:human-evaluation}
To evaluate human performance against our benchmark, we designed and conducted a controlled experiment involving human participants. We recruited a total of six human participants for this study, each independently evaluating all videos. Participants were instructed to view each video carefully and subsequently record their responses on an anonymized website in the following structured form:
1) \textbf{Perceptibility Rating (1-5):} Participants rated how perceptible the presented word, shape, or object was, ranging from 1 (very difficult to perceive) to 5 (very clearly perceptible). This measure provided insights into the clarity and ease of visual grouping.
%%%%%%%%%%%%%%%%%%%%%%%%%%%
% \begin{table}{r}{0.60\textwidth}
\begin{table}[t]
\centering
% \vspace{-0.1in}
\resizebox{\linewidth}{!}{
\begin{tabular}{l|cc|cc|cc}
\toprule
\rowcolor{lightgray}
\multirow{2}{*}{\textbf{Annotator}} & \multicolumn{2}{c|}{\textbf{Text}} & \multicolumn{2}{c|}{\textbf{Images}} & \multicolumn{2}{c}{\textbf{Dynamic Scenes}} \\
\cline{2-7}
 & \textbf{Acc(\%)} & \textbf{Perc(1-5)} & \textbf{Acc(\%)} & \textbf{Perc(1-5)} & \textbf{Acc(\%)} & \textbf{Perc(1-5)} \\
\midrule
Annotator 1 & 99.5 & 4.7 & 99.5 & 4.7 & 96.5 & 4.3 \\
Annotator 2 & 98.6 & 4.8 & 98.4 & 4.9 & 91.2 & 4.0 \\
Annotator 3 & 99.5 & 4.9 & 97.2 & 4.5 & 94.7 & 4.4 \\
Annotator 4 & 97.6 & 4.6 & 96.7 & 4.5 & 91.2 & 4.0 \\
Annotator 5 & 100.0 & 4.8 & 99.5 & 4.7 & 99.0 & 4.7 \\
Annotator 6 & 98.0 & 4.7 & 97.8 & 4.5 & 93.0 & 4.2 \\
\midrule
\rowcolor{lightgray}
\textbf{Mean} & \textbf{98.9$\pm$0.7} & \textbf{4.8$\pm$0.0} & \textbf{98.2$\pm$1.1} & \textbf{4.7$\pm$0.1} & \textbf{94.3$\pm$3.1} & \textbf{4.3$\pm$0.1} \\
\bottomrule
\end{tabular}
}
% \vspace{-0.1in}
\caption{Human evaluation results showing accuracy and perceptibility ratings across different visual categories in \textbf{\texttt{SpookyBench}}.}
\label{tab:human_evaluation}
\vspace{-0.10in}
\end{table}
%%%%%%%%%%%%%%%%%%%%%%%%%%%
2) \textbf{Words/Objects Identification:} Participants typed out exactly what they identified in the video. This response directly tested the accuracy of their visual perception.
We collect and evaluate participant responses using exact match criteria based on our predefined labels. Similar to the evaluation accuracy of the video language models for the categories of Object Images and Dynamic Scenes, we accepted multiple correct responses to avoid ambiguity. Table \ref{tab:human_evaluation} shows the average precision and the perception rating of different annotators for different categories. The results show high human performance across all categories: participants correctly identified Words with 98\% accuracy, while Object Images had 92\% accuracy. We also observe a very high perceptibility rating (4.8 for texts and 4.3 and 4.0 for Object images and Dynamic scenes, respectively) across all three categories. This shows that the human brain can easily extract coherent information in videos, which seems to be very difficult for video language models.
\vspace{-0.05in}
% \subsection{Impact of Frame Rates On Human and Model Accuracy}
\subsection{Impact of Frame Rates}
\vspace{-0.05in}
To examine whether temporal sampling affects performance, we evaluate both humans and VLMs across frame rates from 1 to 30 FPS.
\begin{table}[t]
% \begin{table}{r}{0.53\textwidth}
\centering
\resizebox{\linewidth}{!}{
\begin{tabular}{lrrrrr}
\toprule
\textbf{Category} & \textbf{1 FPS} & \textbf{5 FPS} & \textbf{10 FPS} & \textbf{20 FPS} & \textbf{30 FPS} \\
\midrule
Images & 0.0 & 12.5 & 80.0 & 95.8 & 97.5 \\
Words & 0.0 & 10.8 & 35.8 & 95.8 & 95.8 \\
Videos & 0.0 & 15.0 & 62.5 & 93.3 & 93.3\\
\midrule
\textbf{Average} & 0.0 & 12.8 & 59.4 & 95.0 & 95.6 \\
\bottomrule
\end{tabular}
}
% \vspace{-0.1in}
\caption{Human accuracy (\%) across different content categories at varying frame rates. Results are averaged across 3 participants on 120 videos (40 per category).}
\label{tab:human_fps}
% \end{table}
\end{table}
Three human participants tested 120 randomly sampled videos (40 per category) at 1, 5, 10, 20, and 30 FPS, while four VLMs (Qwen2-VL-7B, Qwen2.5-VL-7B, Qwen2.5-VL-3B, and GPT-4o) were evaluate using identical temporal downsampling. As shown in Tables~\ref{tab:human_fps} and~\ref{tab:model_fps}, human accuracy remains above 95\% at 20-30 FPS, degrades to 59.4\% at 10 FPS, and drops to 0\% at 1 FPS. In contrast, all VLMs achieved 0\% accuracy across all frame rates. This demonstrates that temporal sampling frequency does not explain the performance gap between humans and current video-language models, indicating that VLMs lack the architectural mechanisms to process information conveyed through temporal patterns regardless of temporal resolution.
\vspace{-0.05in}
\subsection{Impact of Finetuning}
% \subsection{Impact of Finetuning on Model Accuracy}
\vspace{-0.05in}
\begin{figure}
    \centering
    \includegraphics[width=0.85\linewidth]{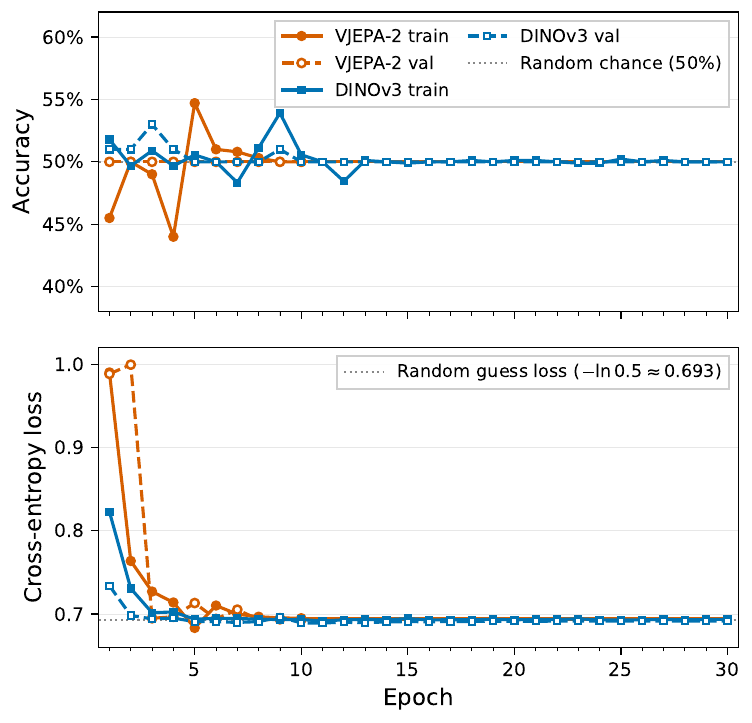}
    \vspace{-0.15in}
    \caption{Training and validation loss over all batch steps for a random binary classifier tasked with detecting foreground noise in video clips. The loss oscillates around 0.7, consistent with random prediction on a balanced dataset.}
    \label{fig:videocls_train_curve}
\end{figure}

To investigate whether the performance gap stems from out-of-distribution data rather than architectural limitations, we finetuned two state-of-the-art video-language models on SpookyBench: InternVL2.5-8B and Qwen2-VL-7B. Both models were trained on 400 SpookyBench videos for up to 30 epochs using LlamaFactory~\citep{zheng2024llamafactory}. Despite this targeted training on the exact task and data distribution, both models maintained 0\% accuracy on the test set. This result demonstrates that the failure to decode temporal patterns is not attributable to domain mismatch or insufficient exposure to the task, but rather indicates a fundamental architectural inability to process information conveyed purely through motion without relying on spatial content. We refer the reader to the supplementary material for detailed results.

\noindent \textbf{Can SOTA Visual Models Tell There's an Object? } We finetuned two video classification models, VJEPA-2\citep{assran2025vjepa2selfsupervisedvideo} and DINOv3\cite{simeoni2025dinov3}, on the task of predicting foreground noise presence. After 30 epochs, training loss saturated near chance-level accuracy (50\%) on the validation set (Figure~\ref{fig:videocls_train_curve}), indicating that neither model could learn a discriminative representation from the frame-level features. The inability to overfit confirms that temporal patterns in this data are inaccessible to architectures operating on individual frames.
\vspace{-0.05in}

\subsection{Impact of Motion Boundary Augmentation}

\label{subsec:motion-boundary-augmentation}

\vspace{-0.05in}

To further validate our hypothesis that VLM failure stems from an inability to extract temporal patterns rather than task impossibility or OOD, we pre-computed motion boundaries using classical optical flow (Farneback method~\citep{farneback2003two}) and overlaid them onto the noisy frames before feeding them to models. \textit{This converts implicit temporal information into explicit spatial cues}.
As shown in Table~\ref{tab:motion_masked_results}, Qwen2-VL-7B jumps from 0\% to 51.54\% overall (50.95\% on text, 70.51\% on images), and GPT-4o reaches 59.10\%. The substantial performance gain when temporal information is converted to spatial representations provides computational evidence that: (1) the benchmark tasks are solvable when appropriate preprocessing is applied, (2) current VLM architectures can process motion information if presented spatially, and (3) the fundamental limitation lies in the absence of temporal integration mechanisms that perform frame differencing and motion extraction, operations that classical computer vision methods accomplish routinely.
% \begin{table}[t]
% \centering
% \resizebox{0.7\linewidth}{!}{
% \begin{tabular}{lccc}
% \toprule
% \textbf{Category} & \textbf{Baseline} & \textbf{Motion-Augmented} & \textbf{Gain} \\
% \midrule
% Words & 0\% & 50.95\% & +50.95\% \\
% Images & 0\% & 70.51\% & +70.51\% \\
% Videos & 0\% & 1.75\% & +1.75\% \\
% \midrule
% \textbf{Overall} & 0\% & 51.54\% & +51.54\% \\
% \bottomrule
% \end{tabular}
% }
% \vspace{-0.1in}
% \caption{Performance comparison of Qwen2-VL-7B with and without motion boundary augmentation. Pre-computing and highlighting motion boundaries (in red) on frames dramatically improves accuracy, demonstrating that temporal information is computationally extractable but inaccessible to current VLM architectures.}
% \label{tab:motion_masked_results}
% \vspace{-0.15in}
% \end{table}

\begin{table}[t]
\centering
\resizebox{0.95\linewidth}{!}{
\begin{tabular}{lcccc}
\toprule
\textbf{Model} & \textbf{Words} & \textbf{Images} & \textbf{Videos} & \textbf{Overall} \\
\midrule
Qwen2-VL-7B (Baseline) & 0.0\% & 0.0\% & 0.0\% & 0.0\% \\
Qwen2-VL-7B (Augmented) & 50.95\% & 70.51\% & 1.75\% & 51.54\% \\
\midrule
GPT-4o (Baseline) & 0.0\% & 0.0\% & 0.0\% & 0.0\% \\
GPT-4o (Augmented) & 56.19\% & 83.33\% & 3.51\% & 59.10\% \\
\bottomrule
\end{tabular}
}
% \vspace{-0.1in}
\caption{Performance comparison with and without motion boundary augmentation. Pre-computing and highlighting motion boundaries dramatically improves accuracy, demonstrating that temporal information is computationally extractable but inaccessible to current VLM architectures.}
\label{tab:motion_masked_results}
\end{table}
The Videos category shows minimal improvement (1.75\%), consistent with our optical flow analysis in the supplementary material: complex articulated motion in dynamic scenes produces fragmented boundaries that remain hard to interpret even when spatially highlighted.

\vspace{-0.05in}
\section{Results and Discussion}
\vspace{-0.05in}
Table \ref{tab:model_performance} presents the accuracy scores on the \textbf{\texttt{SpookyBench}} benchmark. Human participants achieved 98\% accuracy under all test conditions. 
In contrast, all Video-VLMs scored 0\% regardless of the type, size, or origin of the model. This pattern was held across all three task categories in our benchmark: 
temporal symbol recognition, temporal sequence understanding, and temporal pattern reasoning.
We test two different prompting strategies to determine if performance limitations could be overcome through interface modifications. First, we used direct prompts with basic instructions asking the models to identify content in the videos. Next, we implemented chain-of-thought prompts with explicit guidance to focus on temporal patterns rather than individual frames. As shown in Table \ref{tab:model_performance}, none of these approaches yielded improvements. All models maintained 0\% accuracy across all prompting conditions, suggesting that the limitation is inherent in the model architectures rather than a matter of optimization or prompt design.
Examination of model output revealed consistent failure modes when processing \textbf{\texttt{SpookyBench}} videos.

\begin{table}[t]
\centering
\resizebox{\linewidth}{!}{
\begin{tabular}{lcccc}
\toprule
\textbf{Model} & \textbf{Qwen2-VL-7B} & \textbf{Qwen2.5-VL-7B} & \textbf{Qwen2.5-VL-3B} & \textbf{GPT-4o} \\
\midrule
\textbf{Accuracy (\%)} & 0.0 & 0.0 & 0.0 & 0.0 \\
\bottomrule
\end{tabular}
}
% \vspace{-0.1in}
\caption{VLM accuracy (\%) averaged across all tested frame rates (1-30 FPS).}
\label{tab:model_fps}
\end{table}
% \vspace{-0.12in}
Across all models tested, we observed attempts to extract information from individual frames rather than temporal patterns. When explicitly prompted to consider temporal changes, the models acknowledged the instruction but still failed to identify the patterns. {\bf For instance, Qwen-series models consistently predict ``clock'' or ``coffee cup'' regardless of input, indicating pattern collapse rather than meaningful temporal processing}. Fine-tuned models produced outputs that mimicked training examples without correctly identifying test patterns. In particular, specialized temporal models like TimeChat~\citep{ren2024timechat}, which were specifically designed for fine-grained temporal understanding, failed at the same rate as general-purpose models. This suggests that the limitation extends beyond general Video-VLMs to models explicitly optimized for temporal tasks.
% %%%%%%%%%%%%%%%%%%%%
% \begin{table}[t]
% \centering
% \resizebox{0.85\linewidth}{!}{
% \begin{tabular}{lccc}
% \toprule
% \textbf{Model} & \textbf{100 videos} & \textbf{400 videos} & \textbf{1,100 videos} \\
% \midrule
% VJEPA-2 & 52.3\% & 56.1\% & 52.8\% \\
% DINOv3 & 52.7\% & 52.9\% & 53.2\% \\
% Qwen3-VL-8B & 0.0\% & 0.0\% & 0.0\% \\
% \bottomrule
% \end{tabular}
% }
% \vspace{-0.1in}
% \caption{Training accuracy after finetuning on SpookyBench. All models converge to near-random performance ($\sim$50\%) or 0\%, regardless of dataset size.}
% \label{tab:finetuning_scale}
% \end{table}
% %%%%%%%%%%%%%%%%%%%%%

% \noindent{\bf Architectural Implications for Vision Models.}
\subsection{Architectural Implications for Vision Models.}
Distinctive signal profiles in \textbf{\texttt{SpookyBench}} demonstrate a fundamental gap between human and machine perception of temporal information. 
Current vision models struggle with \textbf{\texttt{SpookyBench}} stimuli primarily because they: (1) lack robust temporal integration mechanisms that could leverage high temporal coherence, (2) process information primarily through spatial rather than temporal channels, and (3) fail to perform motion-based figure-ground segregation effectively.
%%%%%%%%%%%%%%%%%%%%%%%%
% \begin{wrapfigure}{r}{0.6\textwidth}
%%%%%%%%%%%%%%%%%%%%%%%%
The consistently high temporal coherence values in Dynamic Scenes, coupled with their poor static-frame metrics, suggest that successful models must implement recurrent processing or attention mechanisms that operate across extended temporal windows rather than focusing on frame-level feature extraction. The negative motion contrast observed in Dynamic Scenes further indicates that models require more sophisticated motion segregation capabilities to match human perceptual abilities in dynamic visual environments. These findings highlight the need for architectural innovations that specifically address temporal processing limitations. Future models should incorporate dedicated temporal coherence pathways, motion contrast analysis, and longer temporal integration windows to bridge the perception gap demonstrated by \textbf{\texttt{SpookyBench}}.
\vspace{-0.05in}
\section{Conclusion}
\vspace{-0.05in}
In this paper, we introduced \textbf{\texttt{SpookyBench}}, a novel benchmark designed to evaluate the temporal reasoning capabilities of video-language models by isolating temporal understanding from spatial comprehension. Our experiments revealed a striking performance gap: while humans effortlessly achieve 98\% accuracy on tasks requiring pure temporal pattern recognition, all tested models, including state-of-the-art open and closed-source systems, fail completely with 0\% accuracy. This consistent failure across different model architectures, scales, and prompting strategies highlights a fundamental limitation in current video understanding approaches, which typically process spatial features first and then establish temporal connections, rather than integrating spatio-temporal information simultaneously. 
Importantly, our additional analyses with foreground noise detection and providing explicit motion cues to the model further validate that the deficit arises not from task impossibility but from architectural inability to internally extract temporal structure. The benchmark effectively exposes the \textit{time blindness} of current architectures that remain hidden in conventional evaluation settings where spatial features can provide shortcuts to correct answers. We hope that {\bf \texttt{SpookyBench}} will inspire the development of next-generation temporal-connected models.

{
    \small
    \bibliographystyle{ieeenat_fullname}
    \bibliography{main}
}

\clearpage

\end{document}

\onecolumn  % Force one-column even if cvpr.sty overrides documentclass

\maketitlesupplementary
\appendix

% \begin{center}
% % \textbf{\Large Overview}
% \end{center}

\setcounter{tocdepth}{2}
\tableofcontents

\clearpage

\section{Data Statistics}
\label{sec:stats}
\begin{figure}[htbp]
    \centering
    \includegraphics[width=0.4\linewidth]{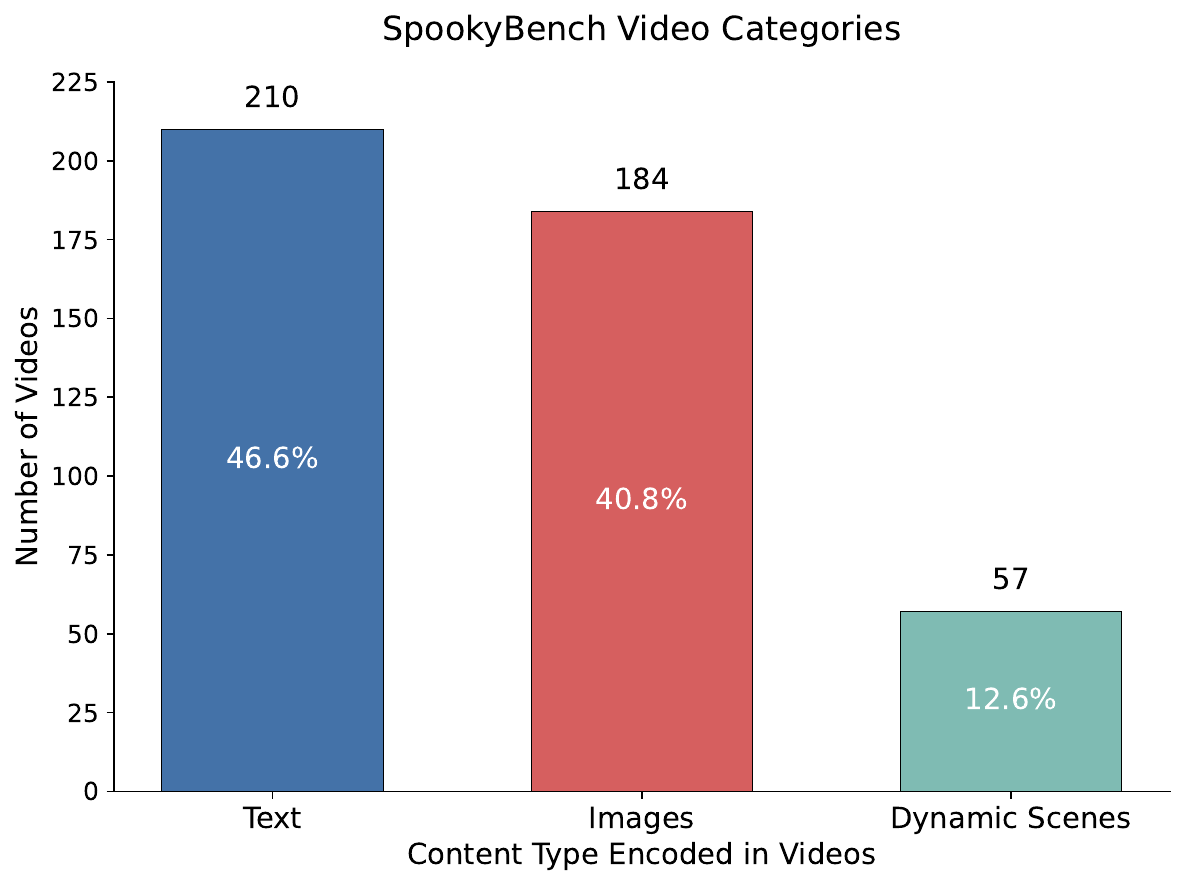}
    \caption{Distribution of the \textbf{SpookyBench} dataset across three video categories. Each category represents a different type of content encoded through temporal noise patterns: \textit{Text}, \textit{Object Images}, and \textit{Dynamic Scenes}.}
    \label{fig:spookybench-distribution}
\end{figure}

Figure \ref{fig:spookybench-distribution} shows the data distribution of SpookyBench.

\section{Prompt Design for Evaluation}
\label{sec:prompts}

Prompt design significantly affects the performance of vision-language models~\citep{jin2021good, gu2023systematic}. We performed careful prompt engineering to ensure fair and comprehensive evaluation, developing a systematic prompting methodology that builds on established principles while introducing elements specific to temporal pattern recognition.

\subsection{Prompt Design Principles}

We designed our prompts based on three key principles:

\begin{enumerate}
    \item \textbf{Specificity}: Each prompt explicitly states that the content is encoded through temporal patterns to direct attention to motion-based cues rather than static frame analysis.
    
    \item \textbf{Category targeting}: We created specialized prompts for each content category (text, objects, dynamic scenes) to account for the different perceptual mechanisms involved in each.
    
    \item \textbf{Constrained response format}: All prompts request brief, specific answers (1-3 words) to ensure objective evaluation and minimize the influence of language generation capabilities.
\end{enumerate}

\subsection{Direct vs. Chain-of-Thought Prompting}

We implemented two distinct prompting strategies shown below. Figure~\ref{fig:category-direct-prompts-detailed} and \ref{fig:category-cot-prompts-detailed} present our category-specific prompts for both strategies.

\begin{itemize}
    \item \textbf{Direct prompts} test immediate pattern recognition without explicit guidance, similar to how humans naturally perceive temporal patterns without conscious step-by-step processing.
    
    \item \textbf{Chain-of-Thought (CoT) prompts} provide explicit steps to guide attention and processing, testing whether models could benefit from structured reasoning about temporal patterns.
\end{itemize}

\begin{figure}[t]
    \centering
    % Text Category
    \begin{tcolorbox}[
        colback=blue!5!white,
        colframe=blue!50!white,
        title=\textbf{Text-Specific Direct Prompt},
        fonttitle=\bfseries\color{white},
        width=0.7\columnwidth
    ]
    This video contains text encoded through temporal patterns. What specific word or phrase do you see? The text is only visible through the temporal changes in the video. Please respond with just the text you identify.
    \end{tcolorbox}
    
    \vspace{0.3cm}
    
    % Object Images Category
    \begin{tcolorbox}[
        colback=blue!5!white,
        colframe=blue!50!white,
        title=\textbf{Object-Specific Direct Prompt},
        fonttitle=\bfseries\color{white},
        width=0.7\columnwidth
    ]
    This video contains a common object encoded through temporal patterns. Individual frames may appear as noise, but an object is visible through the temporal changes. What object do you see? Please respond with just the object name.
    \end{tcolorbox}
    
    \vspace{0.3cm}
    
    % Dynamic Scenes Category
    \begin{tcolorbox}[
        colback=blue!5!white,
        colframe=blue!50!white,
        title=\textbf{Dynamic Scenes-Specific Direct Prompt},
        fonttitle=\bfseries\color{white},
        width=0.7\columnwidth
    ]
    This video contains movement or action encoded through temporal patterns. The content is only visible through temporal changes, not in individual frames. What is being shown? Please respond with just 1-3 words describing what you see.
    \end{tcolorbox}
    
    \caption{Category-specific direct prompts for the SpookyBench benchmark. These prompts test immediate pattern recognition without step-by-step guidance.}
    \label{fig:category-direct-prompts-detailed}
\end{figure}

\begin{figure}[t]
    \centering
    % Text Category
    \begin{tcolorbox}[
        colback=green!5!white,
        colframe=green!50!white,
        title=\textbf{Text-Specific CoT Prompt},
        fonttitle=\bfseries\color{black},
        width=0.7\columnwidth
    ]
    This video encodes text through temporal patterns. Let's think step by step to identify it:
    \begin{enumerate}
        \item Observe how noise pixels move differently across regions of the frame
        \item Look for areas where opposing motion patterns create visible boundaries forming letters
        \item Focus on the overall word or phrase that emerges from these motion boundaries
        \item Read the specific text content revealed by the temporal dynamics
    \end{enumerate}
    
    Please respond with just the text you identify.
    \end{tcolorbox}
    
    \vspace{0.3cm}
    
    % Object Images Category
    \begin{tcolorbox}[
        colback=green!5!white,
        colframe=green!50!white,
        title=\textbf{Object-Specific CoT Prompt},
        fonttitle=\bfseries\color{black},
        width=0.7\columnwidth
    ]
    This video encodes an object through temporal patterns. Let's think step by step to identify it:
    \begin{enumerate}
        \item Notice that individual frames appear as random noise
        \item Track how groups of pixels move coherently over time in different directions
        \item Look for areas where motion patterns reveal object contours and silhouette
        \item Focus on the overall form that emerges and determine what specific object is represented
    \end{enumerate}
    
    Please respond with just the object name.
    \end{tcolorbox}
    
    \vspace{0.3cm}
    
    % Dynamic Scenes Category
    \begin{tcolorbox}[
        colback=green!5!white,
        colframe=green!50!white,
        title=\textbf{Dynamic Scenes-Specific CoT Prompt},
        fonttitle=\bfseries\color{black},
        width=0.7\columnwidth
    ]
    This video encodes movement through temporal patterns. Let's think step by step to identify it:
    \begin{enumerate}
        \item Observe that each frame looks like noise, but regions move differently over time
        \item Look for areas where temporal changes reveal motion of a foreground object
        \item Focus on the action or activity that emerges from the separation of moving and static regions
        \item Identify the specific movement or object in motion
    \end{enumerate}
    
    Please respond with just 1-3 words describing what you see.
    \end{tcolorbox}
    
    \caption{Category-specific chain-of-thought prompts for the SpookyBench benchmark. These prompts provide explicit step-by-step guidance to test structured temporal reasoning.}
    \label{fig:category-cot-prompts-detailed}
\end{figure}

\subsection{Prompt Effectiveness Analysis}

Neither prompt strategy improved model performance on SpookyBench. All tested models achieved 0\% accuracy regardless of prompt type, indicating a fundamental architectural limitation rather than a prompt engineering issue. The complete ineffectiveness of even carefully engineered prompts across all tested models further supports our claim that current video-language models lack the mechanisms needed for processing purely temporal patterns.

\section{LLM-as-a-Judge Evaluation}
\label{sec:llm-judge}

To verify that the observed 0\% accuracy is not an artifact of our exact-match evaluation protocol, we additionally employ an LLM-as-a-judge approach, a standard evaluation methodology widely adopted in VLM benchmarks~\citep{duan2024vlmevalkit}. Specifically, we prompt an LLM to assess whether each model's output is consistent with the ground-truth label, rather than requiring a verbatim match. This supplementary evaluation addresses the possibility that models may exhibit partial understanding of temporal content yet express it in a form that diverges from the predefined label space.

\subsection{Evaluation Pipeline}

Figure~\ref{fig:llm-judge-pipeline} illustrates our LLM-as-a-judge evaluation pipeline. For each video, we provide the judge model (GPT-4o) with: (1) the ground-truth label(s), (2) the model's response, and (3) instructions to determine whether the response correctly identifies the content, allowing for paraphrases, synonyms, and partial matches.

\begin{figure}[t]
\centering
\resizebox{0.6\linewidth}{!}{
\begin{tikzpicture}[
    node distance=1.2cm and 2.0cm,
    block/.style={rectangle, draw, fill=blue!10, text width=10em, text centered, rounded corners, minimum height=3em, font=\small},
    input/.style={rectangle, draw, fill=green!10, text width=9em, text centered, rounded corners, minimum height=3em, font=\small},
    output/.style={rectangle, draw, fill=orange!10, text width=10em, text centered, rounded corners, minimum height=3em, font=\small},
    arrow/.style={-{Stealth[length=3mm]}, thick}
]

% Input nodes
\node[input] (video) {SpookyBench Video};
\node[input, right=2.5cm of video] (gt) {Ground-Truth Label(s)};

% VLM node
\node[block, below=1.0cm of video] (vlm) {Video-Language Model};

% Response node
\node[block, below=1.0cm of vlm] (response) {Model Response};

% Judge node
\node[block, below=1.0cm of response, text width=14em, xshift=2.5cm] (judge) {LLM Judge (GPT-4o) \\ \footnotesize ``Does the response correctly \\ identify the content?''};

% Output node
\node[output, below=1.0cm of judge] (verdict) {Verdict: Correct / Incorrect};

% Arrows
\draw[arrow] (video) -- (vlm);
\draw[arrow] (vlm) -- (response);
\draw[arrow] (response) -- (judge);
\draw[arrow] (gt) |- (judge);
\draw[arrow] (judge) -- (verdict);

\end{tikzpicture}
}
\caption{LLM-as-a-Judge evaluation pipeline. The judge receives both the ground-truth label and the model's response, and determines whether the response correctly identifies the temporal content, allowing for synonyms and paraphrases.}
\label{fig:llm-judge-pipeline}
\end{figure}

\subsection{Results}

The LLM-as-a-judge evaluation yields identical 0\% accuracy across all models and categories. This confirms that the failure is not due to strict string matching: models do not produce responses that are even semantically close to the correct answers. Instead, they hallucinate unrelated content (see Appendix~\ref{sec:qualitative} for examples), confirming that the evaluation protocol is robust.

\section{Binary Classifier Experiment Details}
\label{sec:binary-cls}

To test whether state-of-the-art visual models can detect even the \textit{presence} of foreground objects in SpookyBench videos (without identifying them), we trained binary classifiers using VJEPA-2~\citep{assran2025vjepa2selfsupervisedvideo} and DINOv3~\citep{simeoni2025dinov3}.

\subsection{Dataset Construction}

We constructed a balanced binary classification dataset as follows:
\begin{itemize}
    \item \textbf{Positive class}: SpookyBench videos containing encoded foreground content (text, objects, or dynamic scenes).
    \item \textbf{Negative class}: Videos generated with the same noise generation pipeline but without any foreground mask applied, i.e., uniform random noise with no embedded temporal pattern.
\end{itemize}
The dataset was split into 80\% training and 20\% validation sets, maintaining class balance in both splits.

\subsection{Experiment Details}

To evaluate whether pretrained video and image representation models can discriminate foreground noise presence from raw video, we finetune two recent self-supervised models on our binary classification task: VJEPA-2 (ViT-L/16, 64 frames per clip)~\citep{assran2025vjepa2selfsupervisedvideo} and DINOv3 (ConvNeXt-S, pretrained on LV1689M)~\citep{simeoni2025dinov3}. For both DINOv3 and VJEPA-2, we load their respective models with a randomly initialized attention pooler and classifier head, and finetuned the full model on 64-frame clips at 256$\times$256 resolution. Both models were trained for 100 epochs using cosine-annealed learning rates (DINOv3: $10^{-3}$ with Adam; VJEPA-2: $10^{-4}$ with AdamW, weight decay $0.01$) on a stratified 90/10 train/validation split of the 1{,}000-video dataset (500 per class). Training was performed on a single NVIDIA RTX PRO 6000 GPU per model (batch sizes of 64 and 4, respectively). Neither model learned a discriminative representation. DINOv3 training loss converged to $\sim$0.689 with validation accuracy fluctuating between 48--52\%, barely above the 50\% chance baseline. VJEPA-2 exhibited even stronger signs of failure: validation accuracy remained at exactly 50\% for the majority of training, with the loss plateauing near $\ln 2 \approx 0.693$ --- the theoretical maximum-entropy baseline for balanced binary classification.

These results confirm that the temporal signal encoding foreground noise presence in our stimuli is inaccessible to architectures that operate on per-frame spatial features, even when those features are aggregated temporally (DINOv3 mean pooling) or processed by a dedicated spatiotemporal encoder (VJEPA-2). The foreground and background noise patterns are visually indistinguishable in any single frame; discrimination requires detecting coherent differential motion across frames, a signal that neither frozen per-frame representations nor end-to-end video encoders could extract from the raw pixel input under these training conditions.

\subsection{Results}

Both models converge to approximately 50\% validation accuracy (random chance on a balanced dataset), as shown in Figure 4 of the main paper. VJEPA-2 reached 52.8\% and DINOv3 reached 53.2\% after 30 epochs, with training loss oscillating rather than decreasing. This confirms that the features extracted by these models do not encode the temporal motion patterns present in SpookyBench videos, even at the coarse level of detecting whether a foreground object exists. We also fine-tune Qwen3-VL-8B on the same binary classification task, which similarly achieved 0\% meaningful accuracy, further confirming that VLM architectures cannot extract these temporal patterns even with supervised signal. Table~\ref{tab:binary-cls-training-acc} reports the training accuracy after 80 epochs across varying dataset sizes. VJEPA-2 and DINOv3 hover near chance level regardless of the number of training videos, while Qwen3-VL-8B fails entirely at 0\%.

\begin{table}[h]
\centering
\begin{tabular}{lccc}
\toprule
\textbf{Model} & \textbf{100 videos} & \textbf{400 videos} & \textbf{1,100 videos} \\
\midrule
VJEPA-2      & 52.3\% & 56.1\% & 52.8\% \\
DINOv3       & 52.7\% & 52.9\% & 53.2\% \\
\bottomrule
\end{tabular}
\caption{Training accuracy after 80 epochs for binary classification (foreground presence detection) across different dataset sizes. All models remain near or below chance level.}
\label{tab:binary-cls-training-acc}
\end{table}

\section{Overfitting Experiment with Varying Dataset Sizes}
\label{sec:overfit}

A natural question is whether the 0\% accuracy of fine-tuned VLMs is simply due to insufficient training data. To address this, we conducted finetuning experiments with varying dataset sizes.

\subsection{Setup}

We fine-tuned InternVL2.5-8B and Qwen2-VL-7B using LlamaFactory~\citep{zheng2024llamafactory} on three dataset configurations:

\begin{table}[h]
\centering
\begin{tabular}{lccc}
\toprule
\textbf{Configuration} & \textbf{Training Videos} & \textbf{Epochs} & \textbf{Test Accuracy} \\
\midrule
Small & 100 & 30 & 0\% \\
Medium & 400 & 30 & 0\% \\
Large & 1,100 & 30 & 0\% \\
\bottomrule
\end{tabular}
\caption{Fine-tuning results across different dataset sizes. All configurations yield 0\% test accuracy.}
\label{tab:overfit-sizes}
\end{table}

\subsection{Additional Data Details}

For the large configuration (1,100 videos), we generated additional videos beyond the original 451-video SpookyBench dataset using our data generation pipeline. The additional videos follow the same distribution across categories and use the same noise generation parameters (speckle sizes, noise densities, velocity values) as the original dataset. The test set remained fixed across all configurations to ensure comparability.

\subsection{Analysis}

The consistent 0\% accuracy across all dataset sizes demonstrates that the failure cannot be attributed to insufficient training data. Even with 1,100 training videos and 30 epochs (providing ample opportunity for overfitting), the models cannot learn to extract temporal patterns from the data. This rules out a simple out-of-distribution explanation and points to an architectural inability to process the temporal information encoded in SpookyBench videos.

\section{Impact of FPS}
\label{sec:fps_impact}

To test whether frame rate affects the ability to perceive temporally encoded information, we evaluate both human participants and VLMs across frame rates from 1 to 30 FPS. We test three human participants on 120 randomly sampled videos (40 per category) at frame rates of 1, 5, 10, 20, and 30 FPS. For VLMs, we evaluated Qwen2-VL-7B, Qwen2.5-VL-7B, Qwen2.5-VL-3B, and GPT-4o. Human accuracy remains above 95\% at 20--30 FPS, degrades to 59.4\% at 10 FPS, and drops to 0\% at 1 FPS (see Table 3 in the main paper). In contrast, all VLMs achieved 0\% accuracy at every frame rate tested (Table 6 in the main paper). Human performance degrades gracefully as frame rate decreases, consistent with the known temporal resolution limits of motion perception. VLMs show no sensitivity to frame rate at all, confirming that they are not engaging with temporal information in any meaningful way, regardless of how much temporal data is provided. This rules out temporal undersampling as an explanation for the performance gap.

\section{Temporal Motion Coherence Analysis}
\label{sec:temporal-analysis}

Temporal coherence and motion boundaries provide powerful cues that enable the human visual system to extract shapes from noisy stimuli. We present a comprehensive analysis of these phenomena using our SpookyBench dataset.

\subsection{Motion-Based Perception in Noisy Environments}

\begin{figure}[t]
    \centering
    \includegraphics[width=0.4\columnwidth]{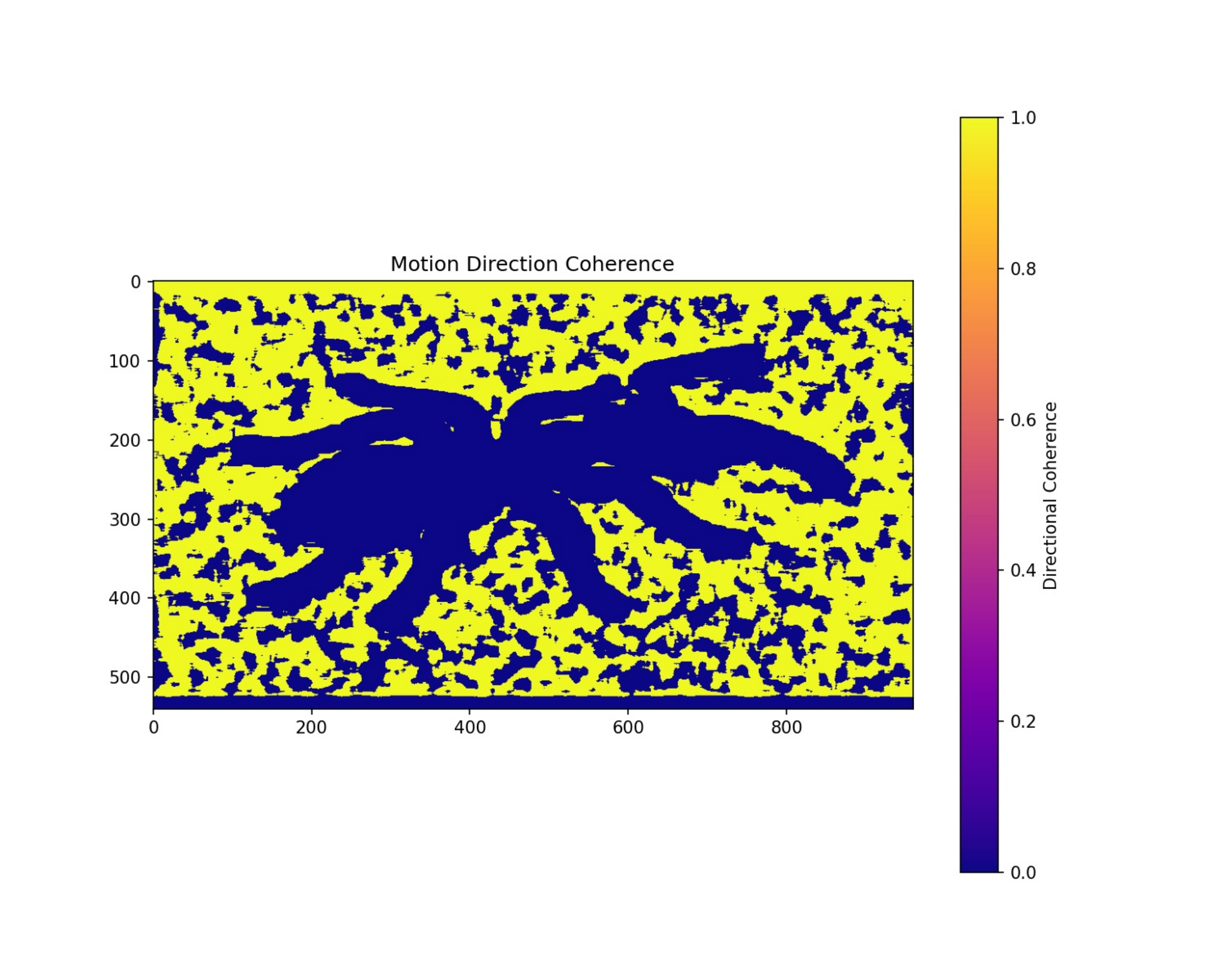}
    \caption{Motion Direction Coherence visualization for the ant silhouette video. Yellow regions (coherence value 1.0) indicate consistent motion direction across frames; blue regions (coherence value 0.0) represent the silhouette itself.}
    \label{fig:motion-analysis}
\end{figure}

Even when individual frames have low SNR, temporal integration of motion information allows robust shape perception. Figure~\ref{fig:motion-analysis} shows the motion direction coherence map of an ant silhouette, revealing how consistent motion patterns across frames enable object identification despite significant noise. Figure~\ref{fig:boundary-analysis} shows both the average motion boundary strength and its overlay on a noisy frame. Motion boundaries emerge clearly despite extreme noise levels in individual frames (measured at -49.07 dB basic SNR).

\begin{figure}[t]
    \centering
    \begin{subfigure}[b]{0.35\columnwidth}
        \includegraphics[width=\textwidth]{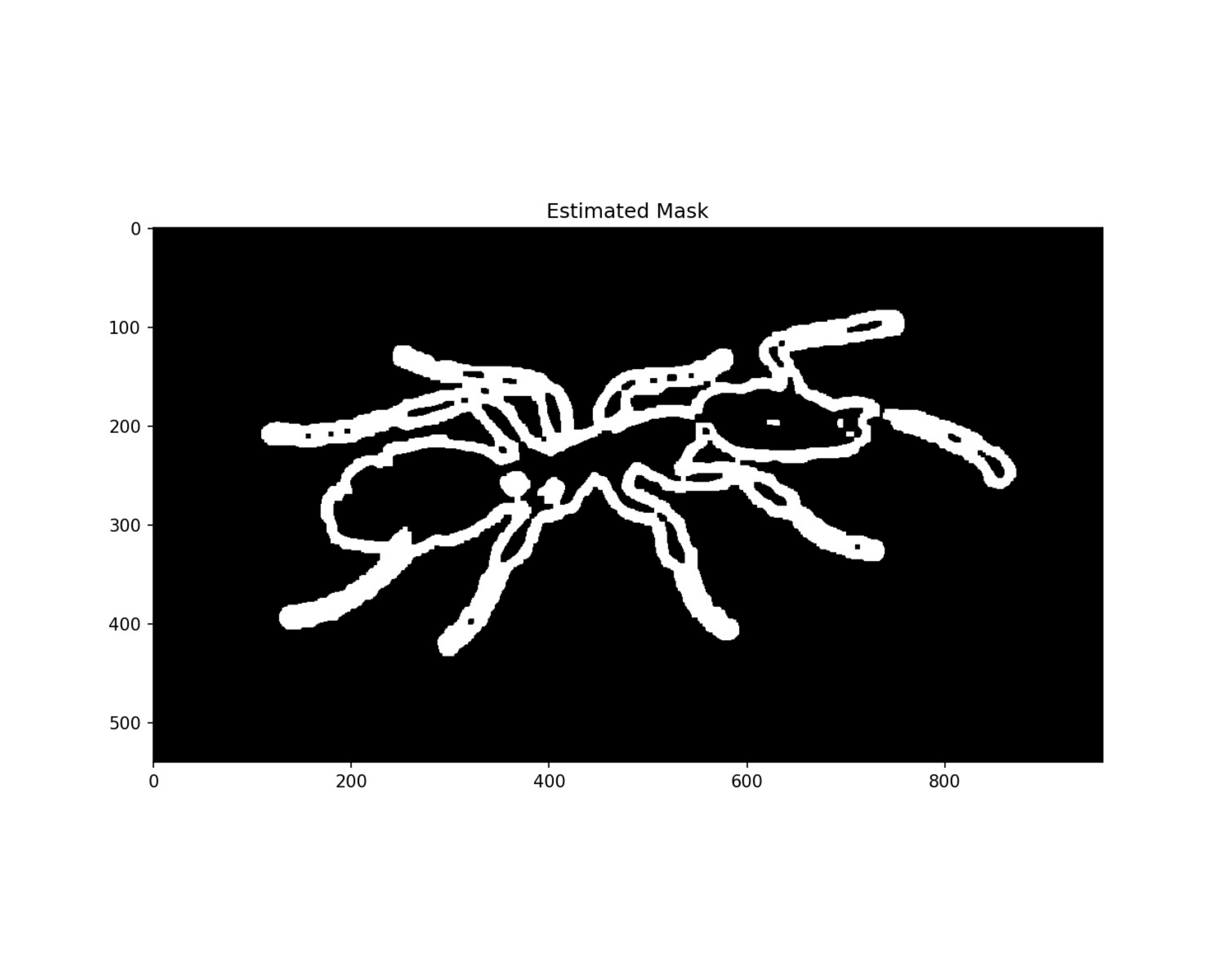}
        \caption{Estimated object mask from temporal motion coherence.}
        \label{fig:estimated-mask}
    \end{subfigure}
    \hfill
    \begin{subfigure}[b]{0.35\columnwidth}
        \includegraphics[width=\textwidth]{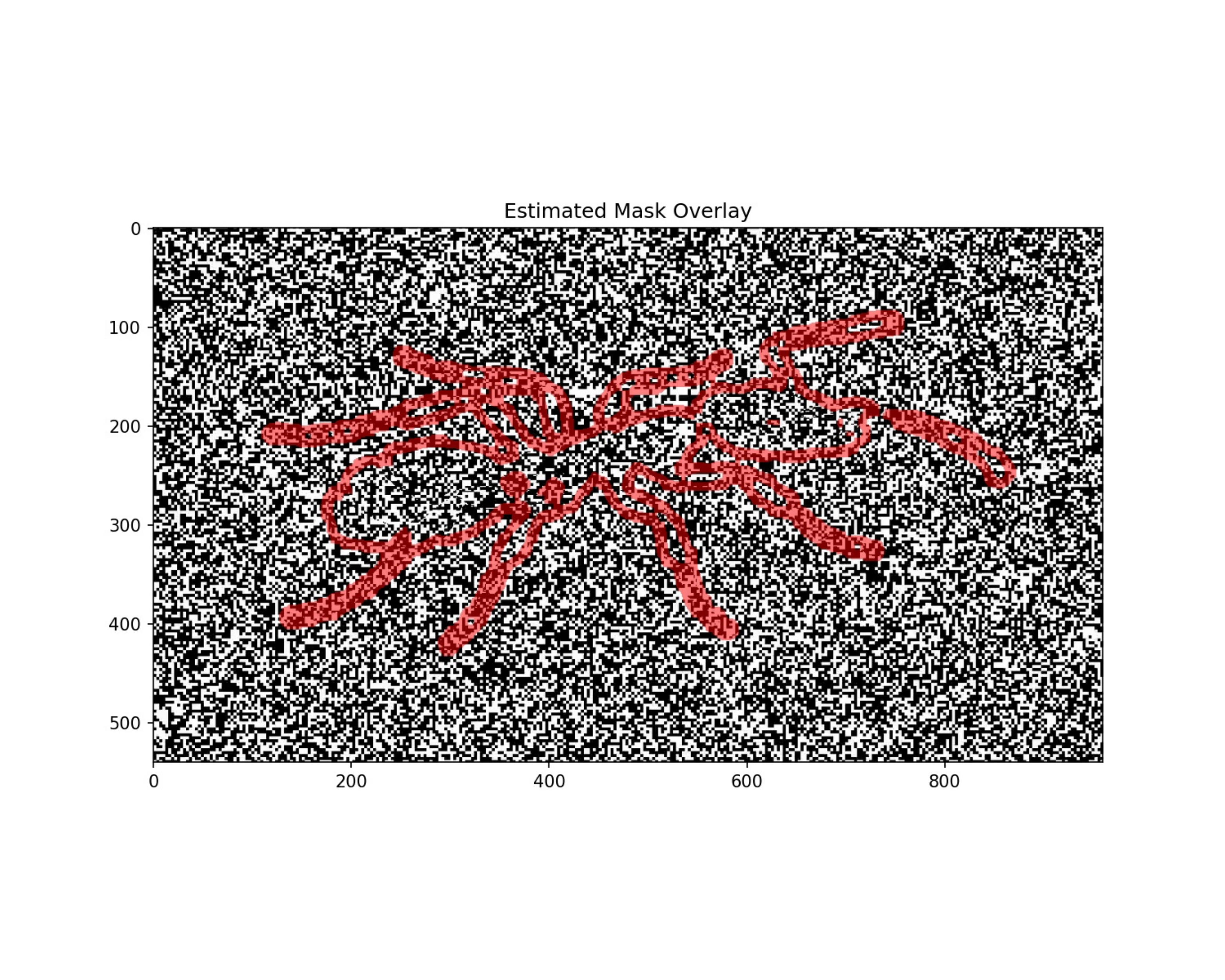}
        \caption{Estimated mask overlay (red) on a single noise frame.}
        \label{fig:mask-overlay}
    \end{subfigure}
    \caption{Shape extraction through temporal integration, demonstrating how object shape can be recovered from noisy video sequences.}
    \label{fig:shape-extraction}
\end{figure}

\begin{figure}[t]
    \centering
    \begin{subfigure}[b]{0.35\columnwidth}
        \includegraphics[width=\textwidth]{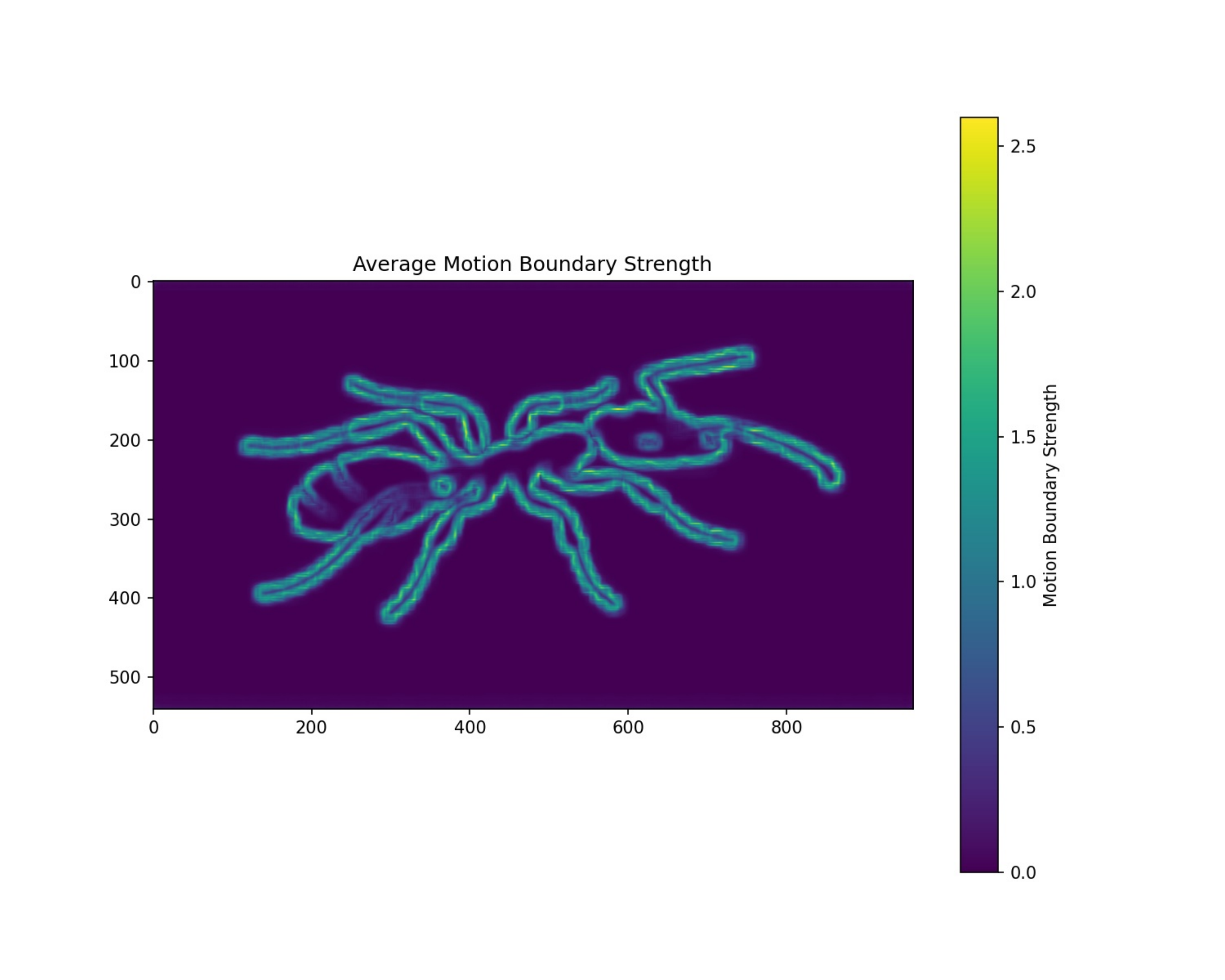}
        \caption{Average motion boundary strength across frames.}
        \label{fig:boundary-strength}
    \end{subfigure}
    \hfill
    \begin{subfigure}[b]{0.35\columnwidth}
        \includegraphics[width=\textwidth]{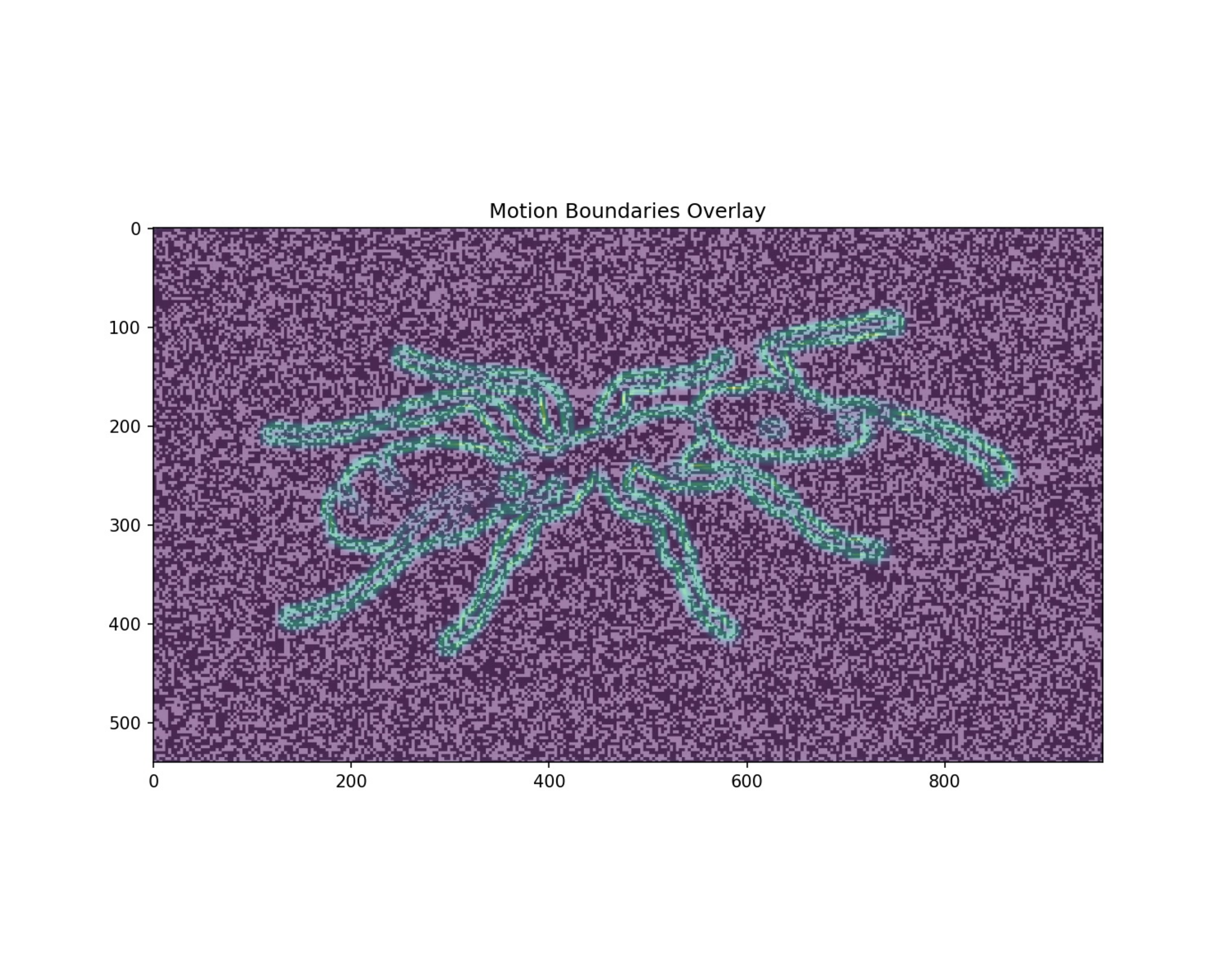}
        \caption{Motion boundaries (teal) overlaid on a single noise frame.}
        \label{fig:boundary-overlay}
    \end{subfigure}
    \caption{Motion boundary analysis showing how temporal integration extracts object boundaries despite extremely noisy individual frames.}
    \label{fig:boundary-analysis}
\end{figure}

\subsection{Signal-to-Noise Ratio Analysis}

Table~\ref{tab:snr-analysis} presents detailed SNR results for the ant silhouette video. While basic and perceptual SNR are extremely low (-49.07 dB and -55.02 dB), temporal coherence SNR and motion contrast SNR are positive (7.18 dB and 14.24 dB), showing that temporal information provides signal enhancement that supports human perception. The contrast between negative frame-based SNR values and positive temporal SNR metrics directly supports our hypothesis: temporal integration is essential for perceiving content in SpookyBench stimuli, and current architectures lack this capability.

\begin{table}[t]
\centering
\begin{tabular}{lc}
\toprule
\textbf{SNR Metric} & \textbf{Value (dB)} \\
\midrule
Basic SNR & -49.07 \\
Perceptual SNR & -55.02 \\
Temporal Coherence SNR & 7.18 \\
Motion Contrast SNR & 14.24 \\
Combined SNR & -20.61 \\
\bottomrule
\end{tabular}
\caption{Signal-to-Noise Ratio Analysis for Ant Silhouette Video.}
\label{tab:snr-analysis}
\end{table}

\section{Qualitative Examples of Model Responses}
\label{sec:qualitative}

We present representative model responses to illustrate the failure modes observed across VLMs on SpookyBench. These examples show that models do not produce partially correct or semantically related answers; instead, they hallucinate entirely unrelated content.

\subsection{Direct Prompting Responses}
 
Figure~\ref{fig:direct-prompt-responses} presents four representative model responses to the text-specific direct prompt. These responses illustrate distinct failure modes, including timestamp hallucination, prompt echoing, outright refusal, and confident hallucination of unrelated content.

\begin{figure}[t]
    \centering

\begin{tcolorbox}[
    colback=gray!5!white,
    colframe=blue!40!white,
    title=\textbf{Prompt},
    fonttitle=\bfseries\color{black},
    width=0.7\columnwidth
]
\textit{``This video contains text encoded through temporal patterns. What specific word or phrase do you see? The text is only visible through the temporal changes in the video. Please respond with just the text you identify.''}
\end{tcolorbox}

\vspace{0.3cm}

\begin{tcolorbox}[colback=red!5!white, colframe=red!40!white, title=\textbf{Response 1 (Timestamp Hallucination)}, fonttitle=\bfseries\color{black}, width=0.7\columnwidth]
``The text in the video is: 4.3, 4.5, 4.7, 4.8 seconds.''
\end{tcolorbox}
\vspace{0.2cm}

\begin{tcolorbox}[colback=red!5!white, colframe=red!40!white, title=\textbf{Response 2 (Prompt Echoing)}, fonttitle=\bfseries\color{black}, width=0.7\columnwidth]
``I see the phrase `This video contains text encoded through temporal patterns.'\,''
\end{tcolorbox}
\vspace{0.2cm}

\begin{tcolorbox}[colback=red!5!white, colframe=red!40!white, title=\textbf{Response 3 (Refusal / No Detection)}, fonttitle=\bfseries\color{black}, width=0.7\columnwidth]
``I apologize, but I am unable to see any text or patterns in the provided video. It appears to be a random sequence of pixels with no discernible pattern or text.''
\end{tcolorbox}
\vspace{0.2cm}

\begin{tcolorbox}[colback=red!5!white, colframe=red!40!white, title=\textbf{Response 4 (Confident Hallucination)}, fonttitle=\bfseries\color{black}, width=0.7\columnwidth]
``The word `GOOD' is visible in the video.''
\end{tcolorbox}

\caption{Direct prompting responses on SpookyBench. The prompt (top, blue) is shown together with four representative model responses (red), illustrating distinct failure modes.}
\label{fig:direct-prompt-responses}
\end{figure}

\subsection{Chain-of-Thought Prompting Responses}
 
Figure~\ref{fig:cot-prompt-responses} shows model responses when given explicit step-by-step reasoning guidance. Despite the structured CoT prompt, models generate elaborate but entirely fabricated scene descriptions that bear no relation to the actual video content.

\begin{figure}[t]
\centering

\begin{tcolorbox}[
    colback=gray!5!white,
    colframe=green!50!white,
    title=\textbf{Prompt},
    fonttitle=\bfseries\color{black},
    width=0.7\columnwidth
]
\textit{``This video encodes text through temporal patterns. Let's think step by step to identify it: 1. Observe how noise pixels move differently across regions of the frame. 2. Look for areas where opposing motion patterns create visible boundaries forming letters. 3. Focus on the overall word or phrase that emerges from these motion boundaries. 4. Read the specific text content revealed by the temporal dynamics. Please respond with just the text you identify.''}
\end{tcolorbox}

\vspace{0.3cm}

\begin{tcolorbox}[colback=red!5!white, colframe=red!50!white, title=\textbf{CoT Response 1 (Scene Fabrication)}, fonttitle=\bfseries\color{black}, width=0.7\columnwidth]
``In the video, a black and white photo of a cat is shown first. Next, the scene shifts to a black and white texture, followed by a black and white photo of a building. Then, a black and white photo of a cat is shown again. The scene changes to a black and white photo of a building and a black and white photo of a cat again. Finally, a black and white texture of a television screen appears. The video continues to show a black and white photo of a cat. Throughout the video, the dense captions describe the appearance of various objects in each scene, including trees, buildings, people, and various textures. The video is black and white, and the scenes appear to be sequenced in a linear fashion.''
\end{tcolorbox}
\vspace{0.2cm}

\begin{tcolorbox}[colback=red!5!white, colframe=red!50!white, title=\textbf{CoT Response 2 (Texture Description)}, fonttitle=\bfseries\color{black}, width=0.7\columnwidth]
``First, the video shows a texture of an old sheet of paper in black and white. Then, it shows a series of lines and textures on a black and white background. Next, the video shows a black and white texture of old paper. After that, it shows a texture of a piece of paper on a black and white background. Then, it shows a black and white background of textured material. Finally, the video shows a black and white textured pattern. Throughout the video, there are also various objects, such as a dog, a chair, and a handwritten note, but they are not the main focus of the video.''
\end{tcolorbox}

\caption{Chain-of-thought prompting responses on SpookyBench. The CoT prompt (top, green) is shown together with two representative model responses (red), illustrating elaborate but entirely fabricated descriptions.}
\label{fig:cot-prompt-responses}
\end{figure}

\subsection{Analysis of Failure Modes}

Across all tested models, we observe four distinct failure patterns:

\begin{enumerate}
    \item \textbf{Timestamp hallucination}: Models output numerical timestamps instead of content, suggesting they attempt to process temporal metadata rather than temporal visual patterns.
    
    \item \textbf{Prompt echoing}: Models repeat parts of the prompt as their answer, indicating no actual visual processing occurred.
    
    \item \textbf{Confident hallucination}: Models produce specific but incorrect answers with high confidence. For instance, Qwen-series models consistently predict ``clock'' or ``coffee cup'' regardless of input, indicating pattern collapse to high-frequency training concepts.
    
    \item \textbf{Scene fabrication}: When given CoT prompts, models generate elaborate descriptions of imaginary scenes (cats, buildings, paper textures) that have no correspondence to the actual video content, suggesting they are generating plausible-sounding narratives from their language prior rather than processing visual input.
\end{enumerate}

None of these failure modes produce responses that are even partially related to the actual content encoded in the videos, which is consistent with the 0\% accuracy observed under both exact-match and LLM-as-a-judge evaluation.

\section{Additional Images}
\label{sec:add_images}
\begin{figure*}[t]
    \centering
    \begin{tabular}{@{}c@{\hskip 4pt}c@{\hskip 4pt}c@{\hskip 4pt}c@{}}
        \includegraphics[width=0.22\textwidth]{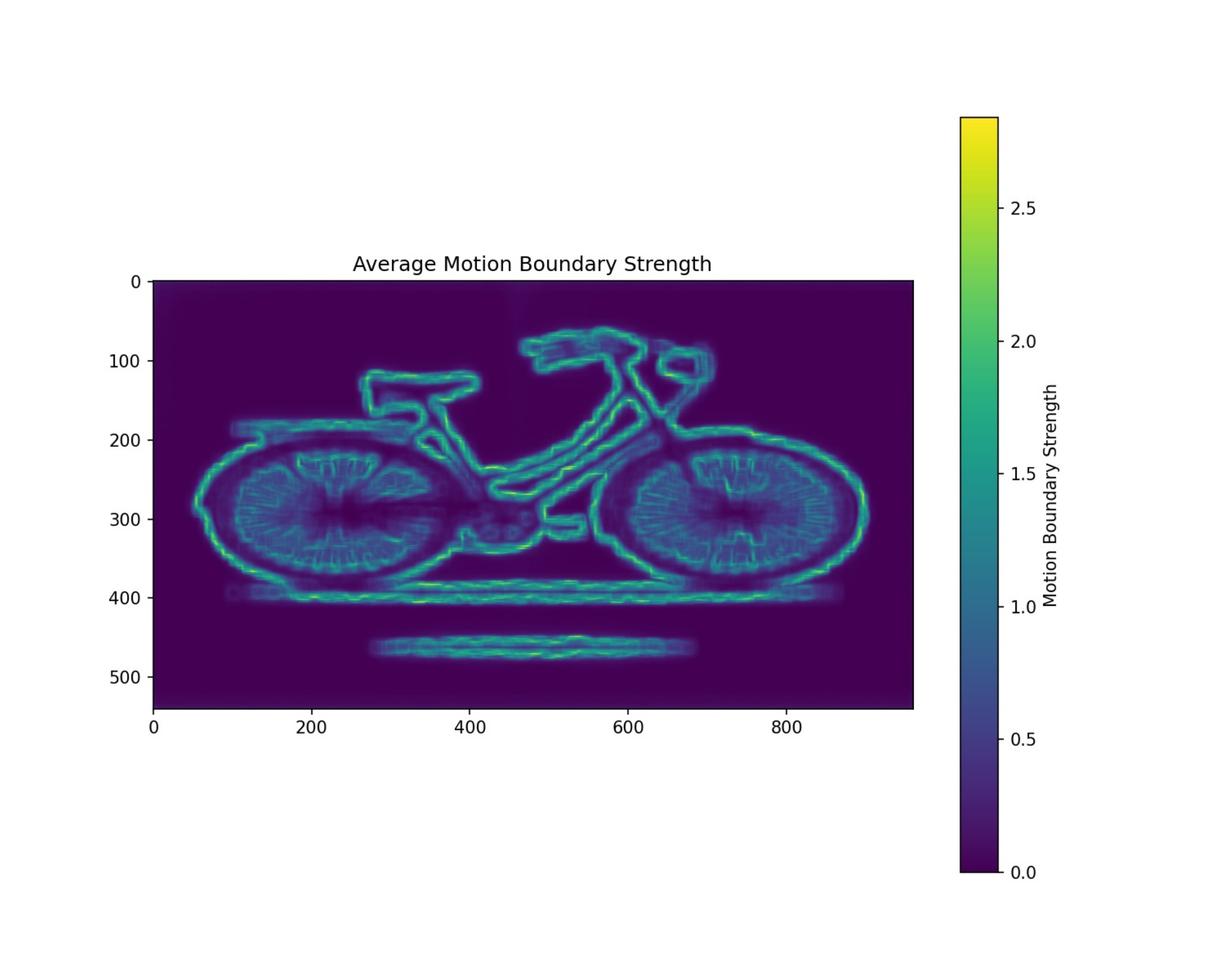} &
        \includegraphics[width=0.22\textwidth]{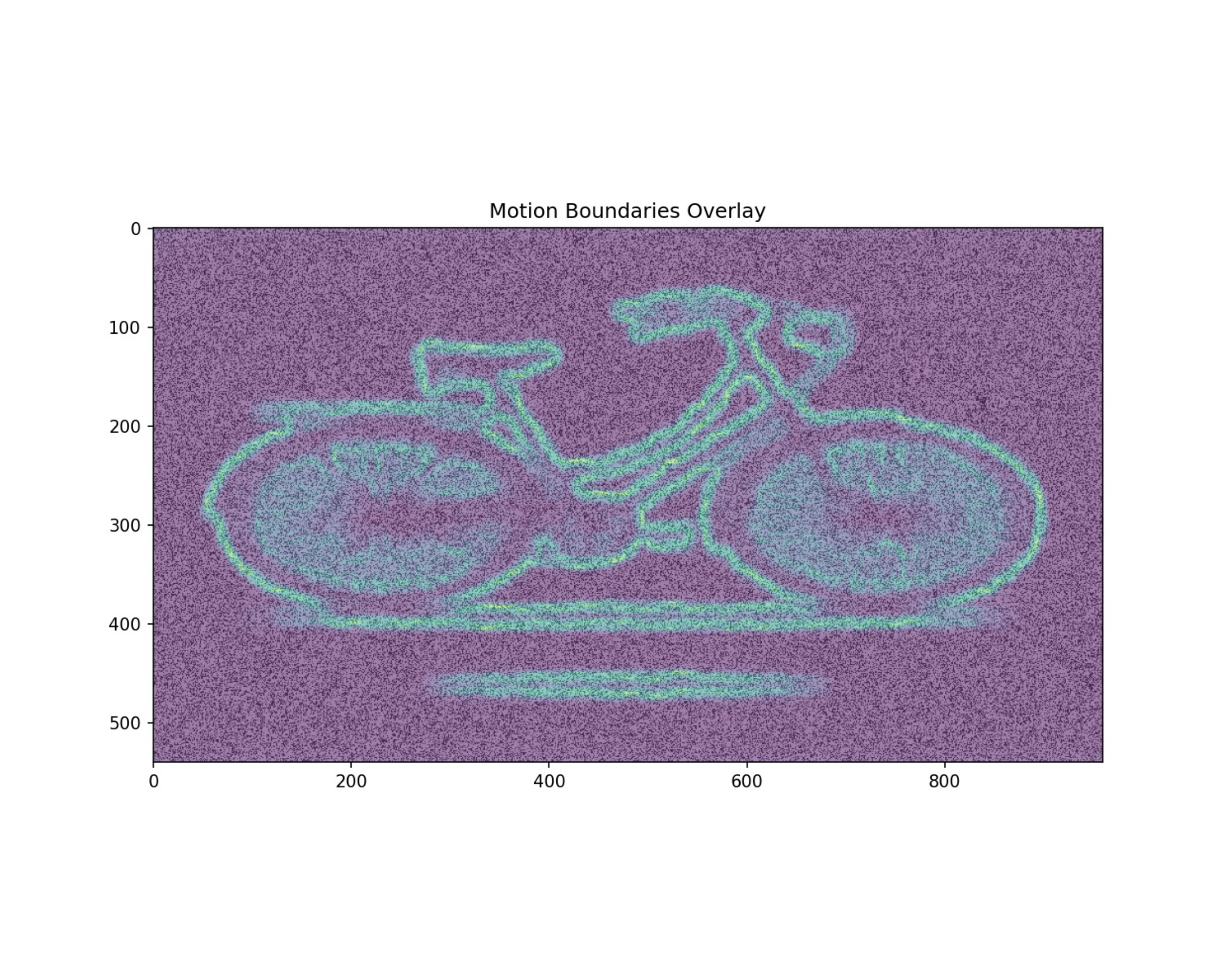} &
        \includegraphics[width=0.22\textwidth]{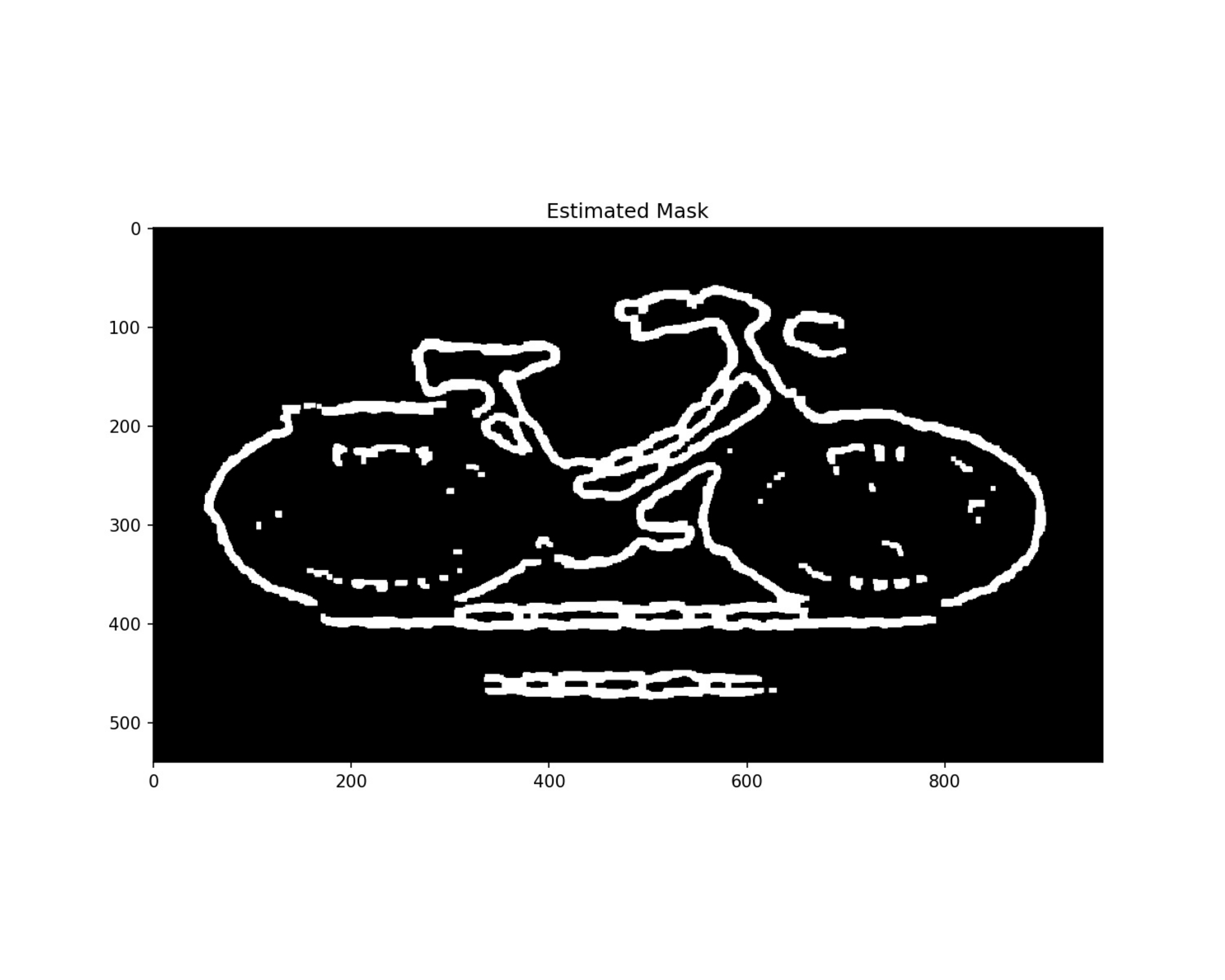} &
        \includegraphics[width=0.22\textwidth]{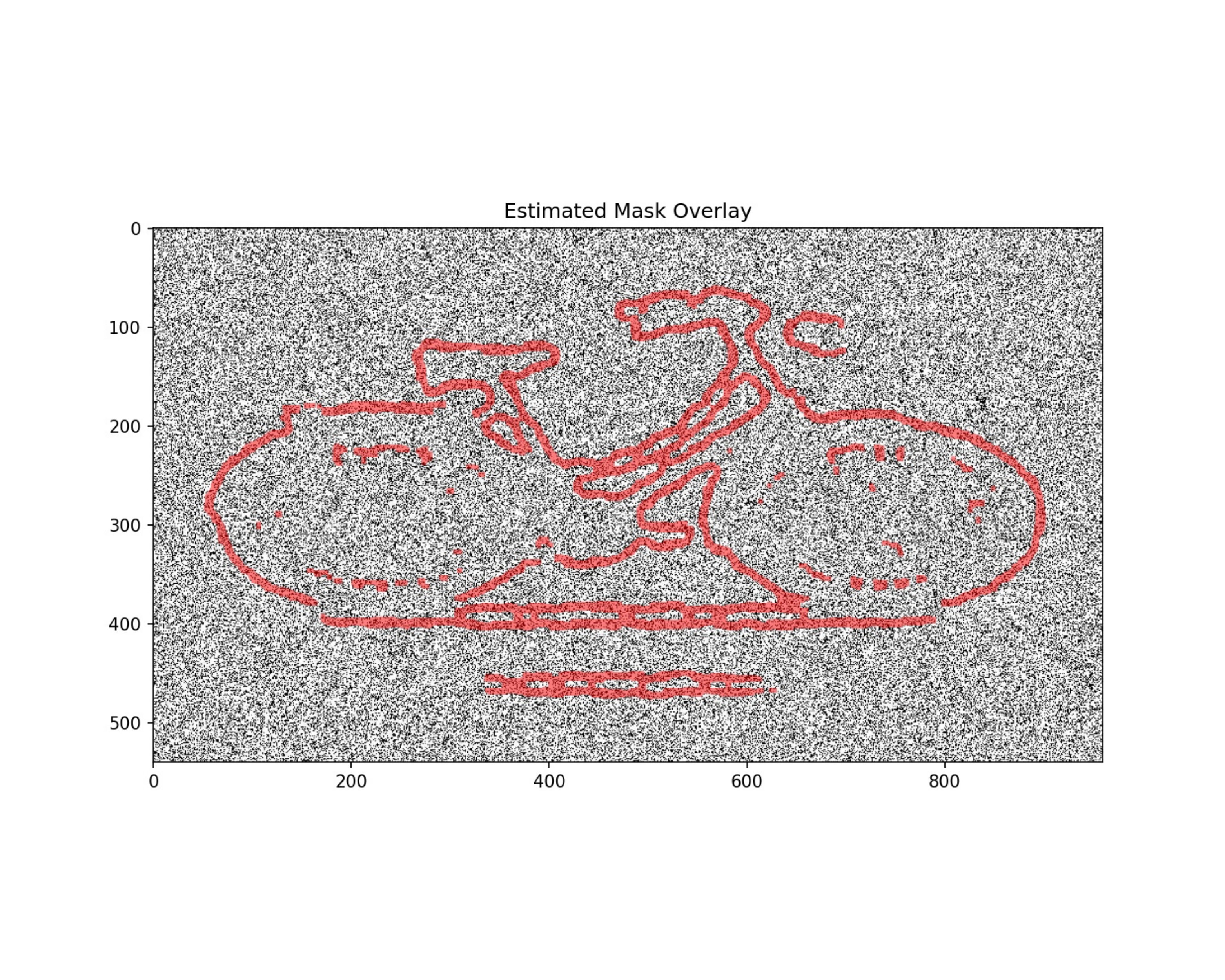} \\
        
        \includegraphics[width=0.22\textwidth]{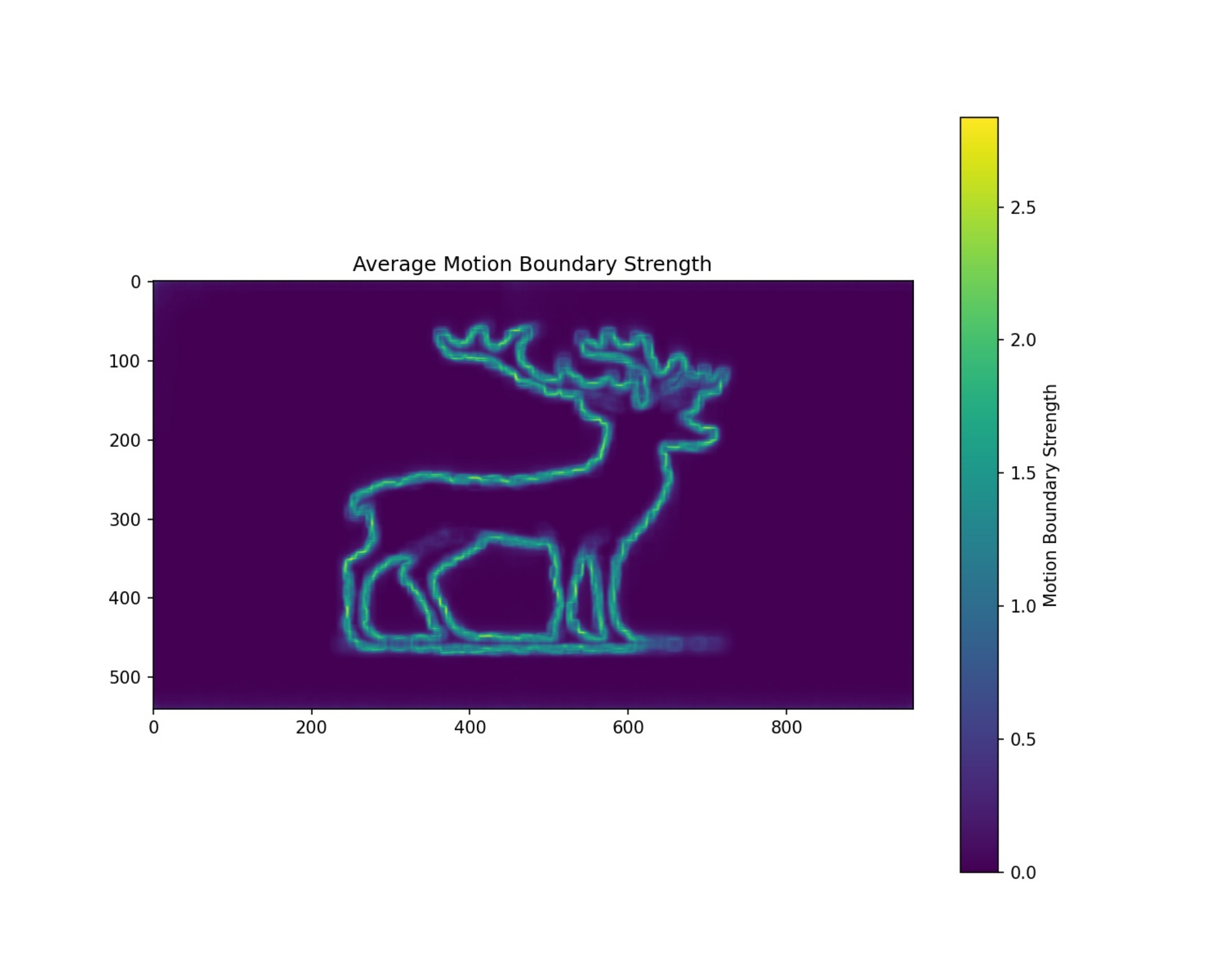} &
        \includegraphics[width=0.22\textwidth]{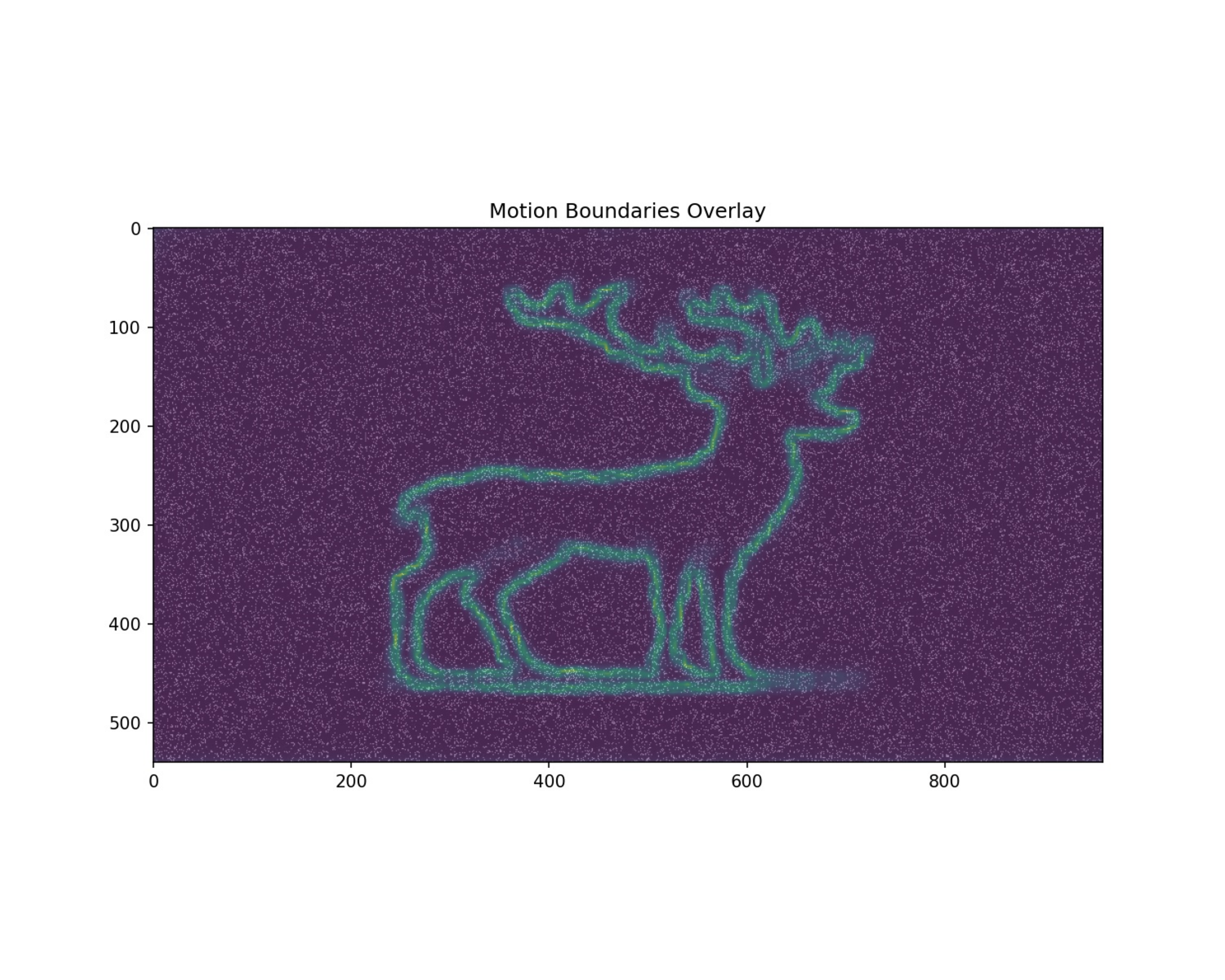} &
        \includegraphics[width=0.22\textwidth]{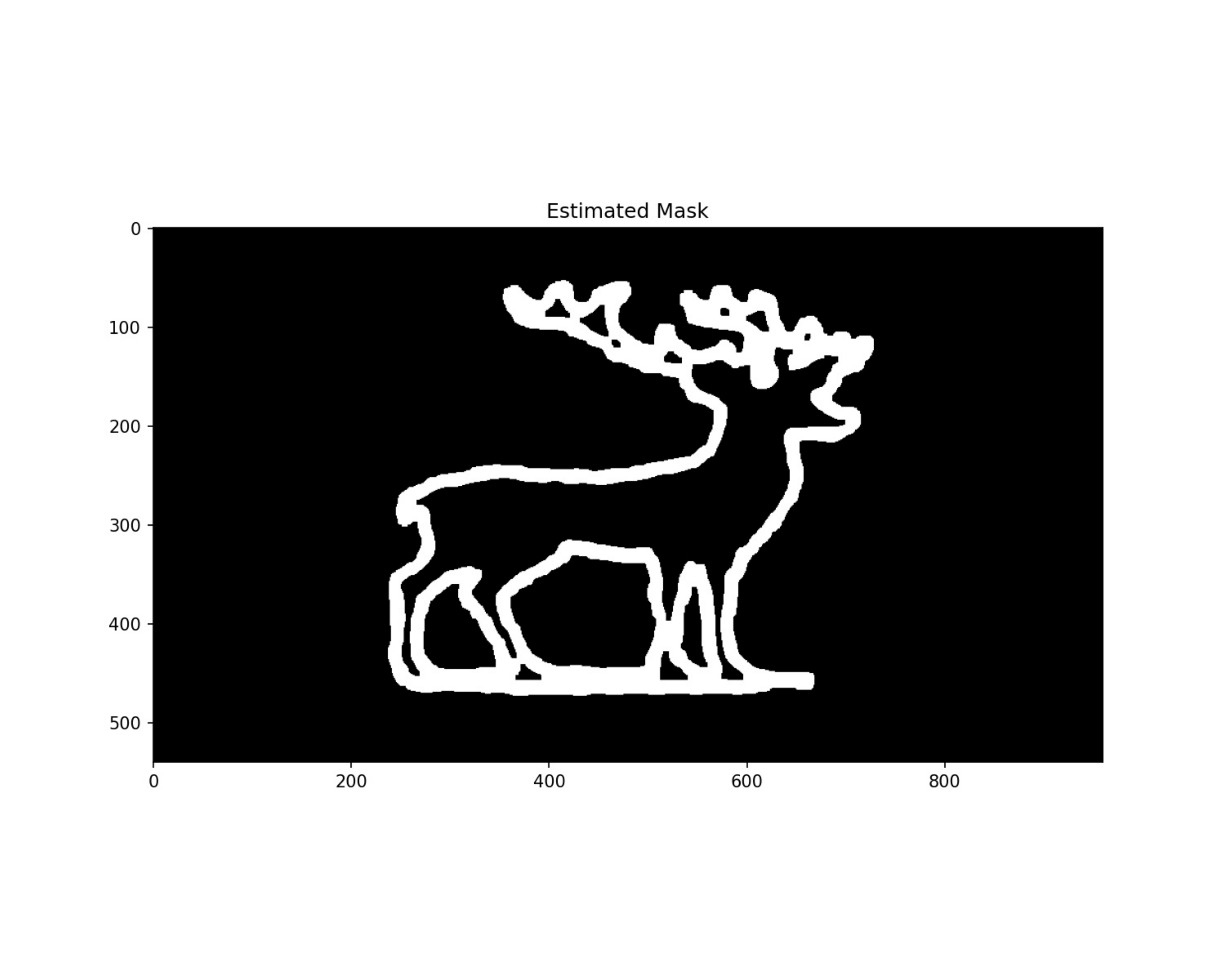} &
        \includegraphics[width=0.22\textwidth]{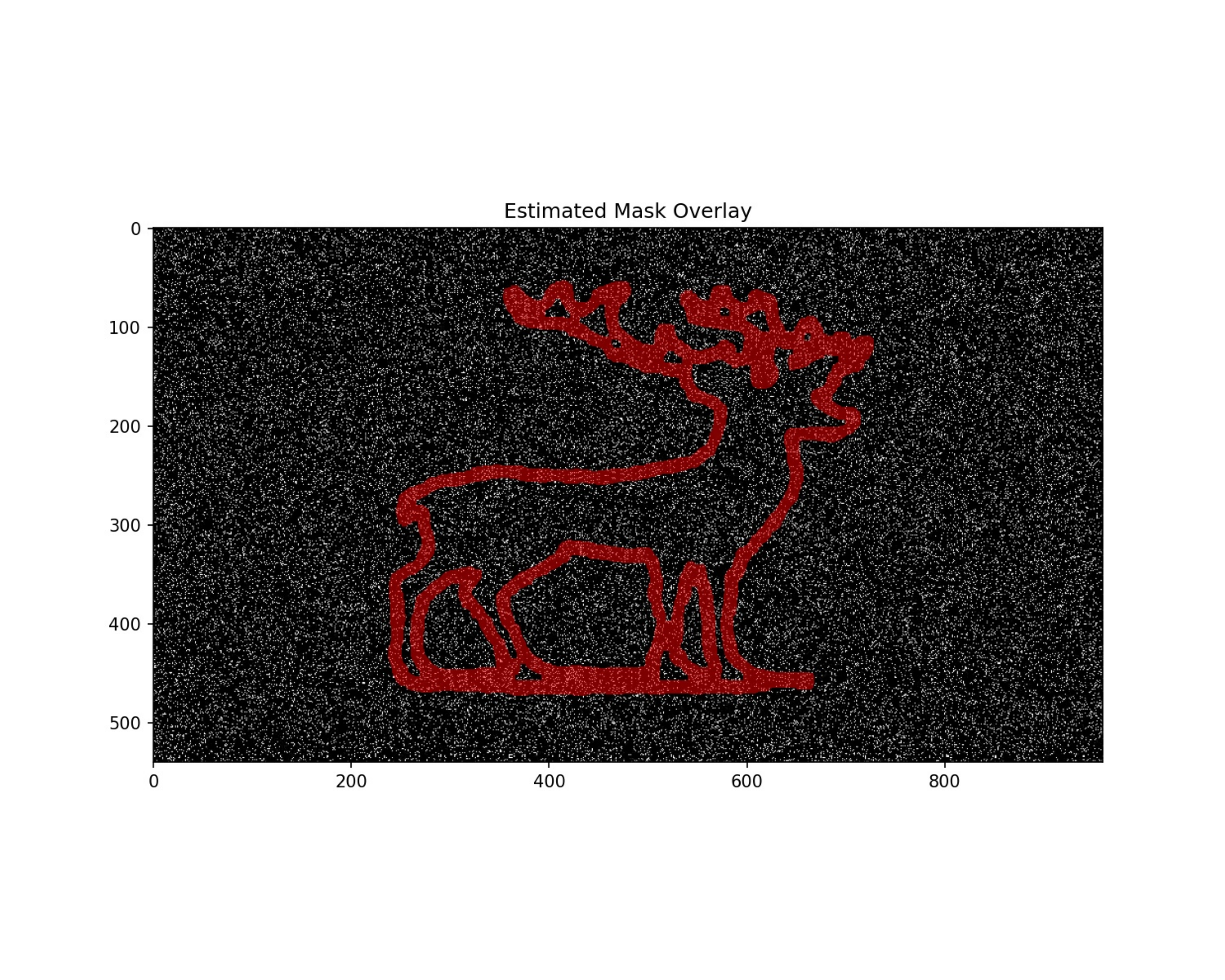} \\
        
        \includegraphics[width=0.22\textwidth]{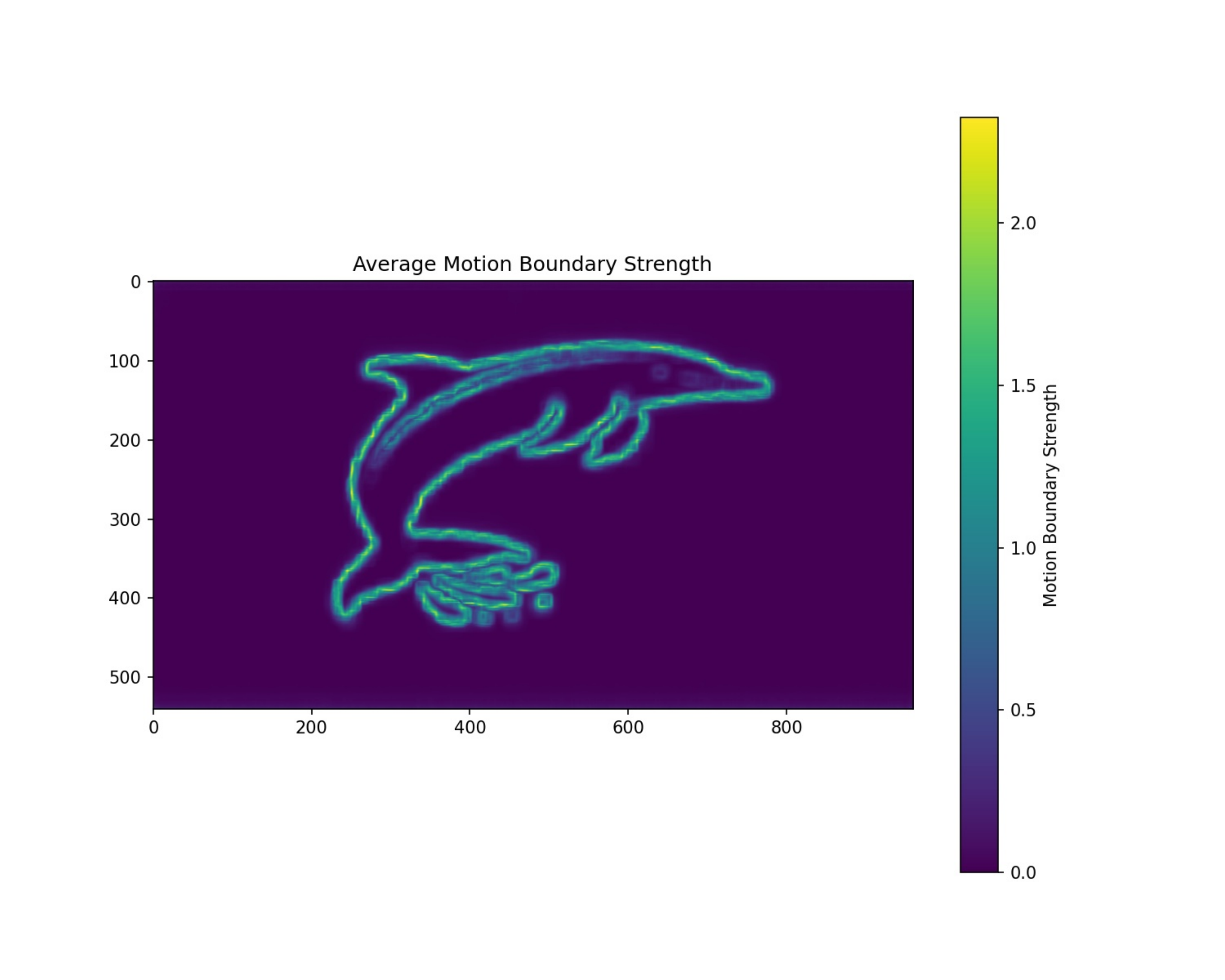} &
        \includegraphics[width=0.22\textwidth]{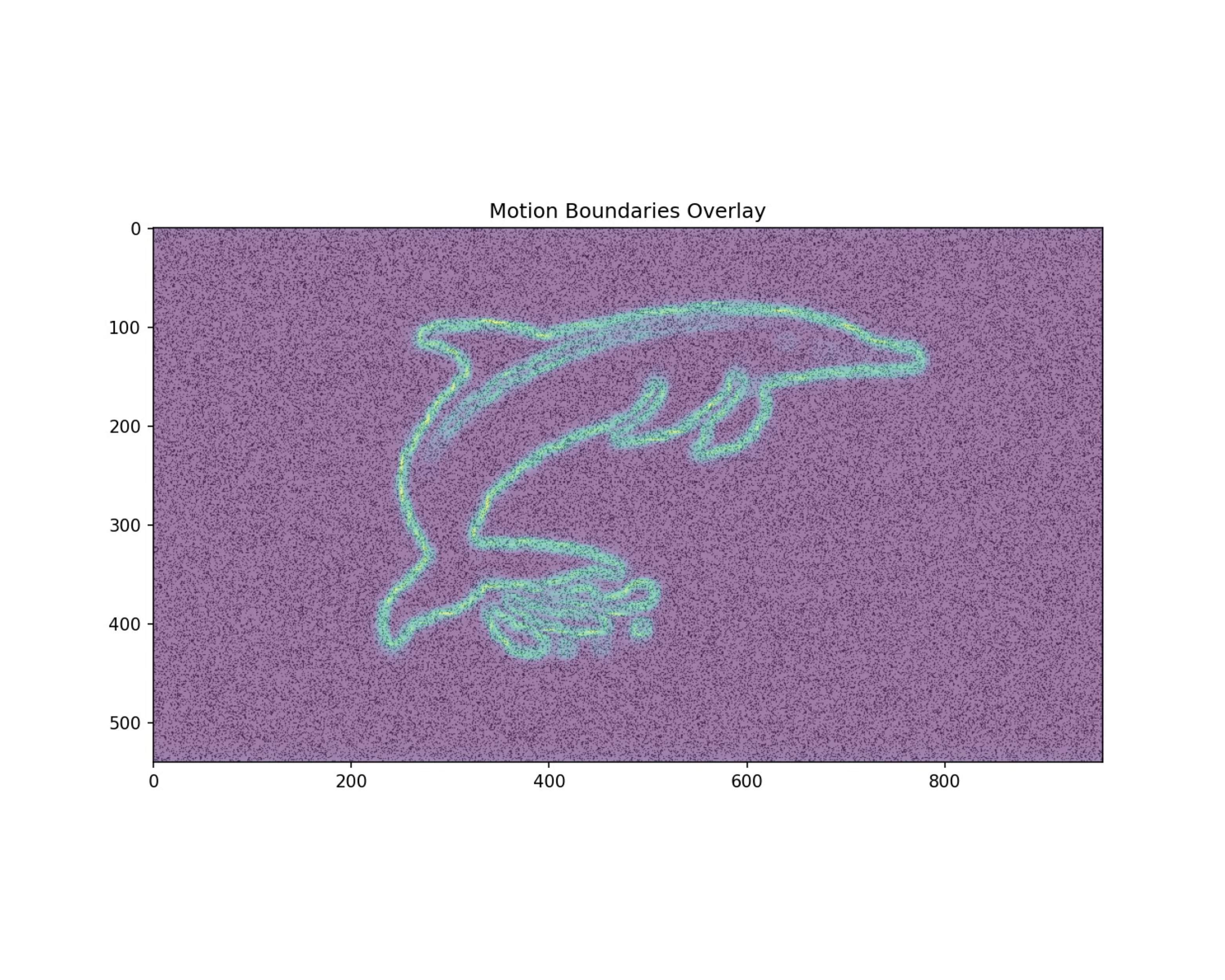} &
        \includegraphics[width=0.22\textwidth]{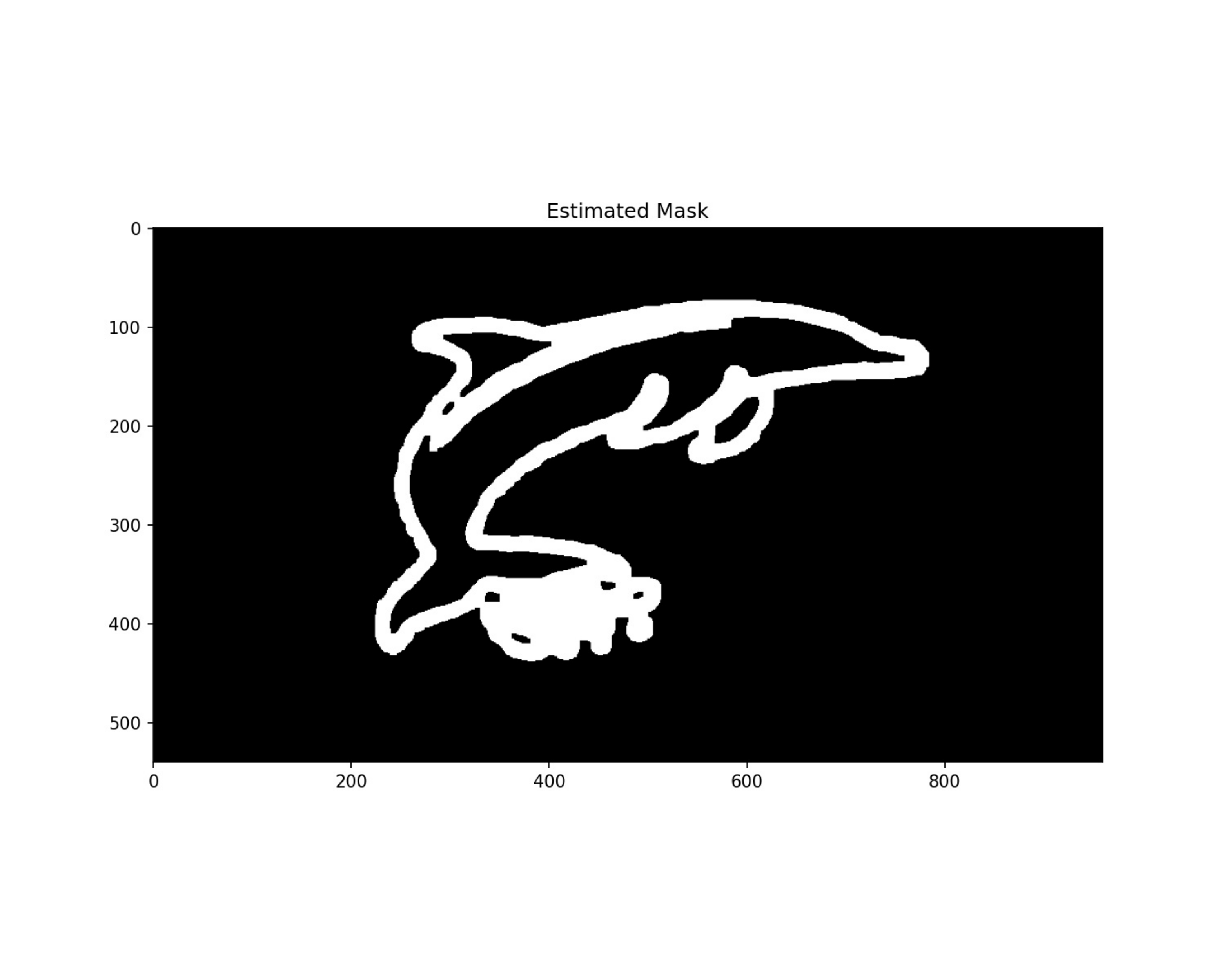} &
        \includegraphics[width=0.22\textwidth]{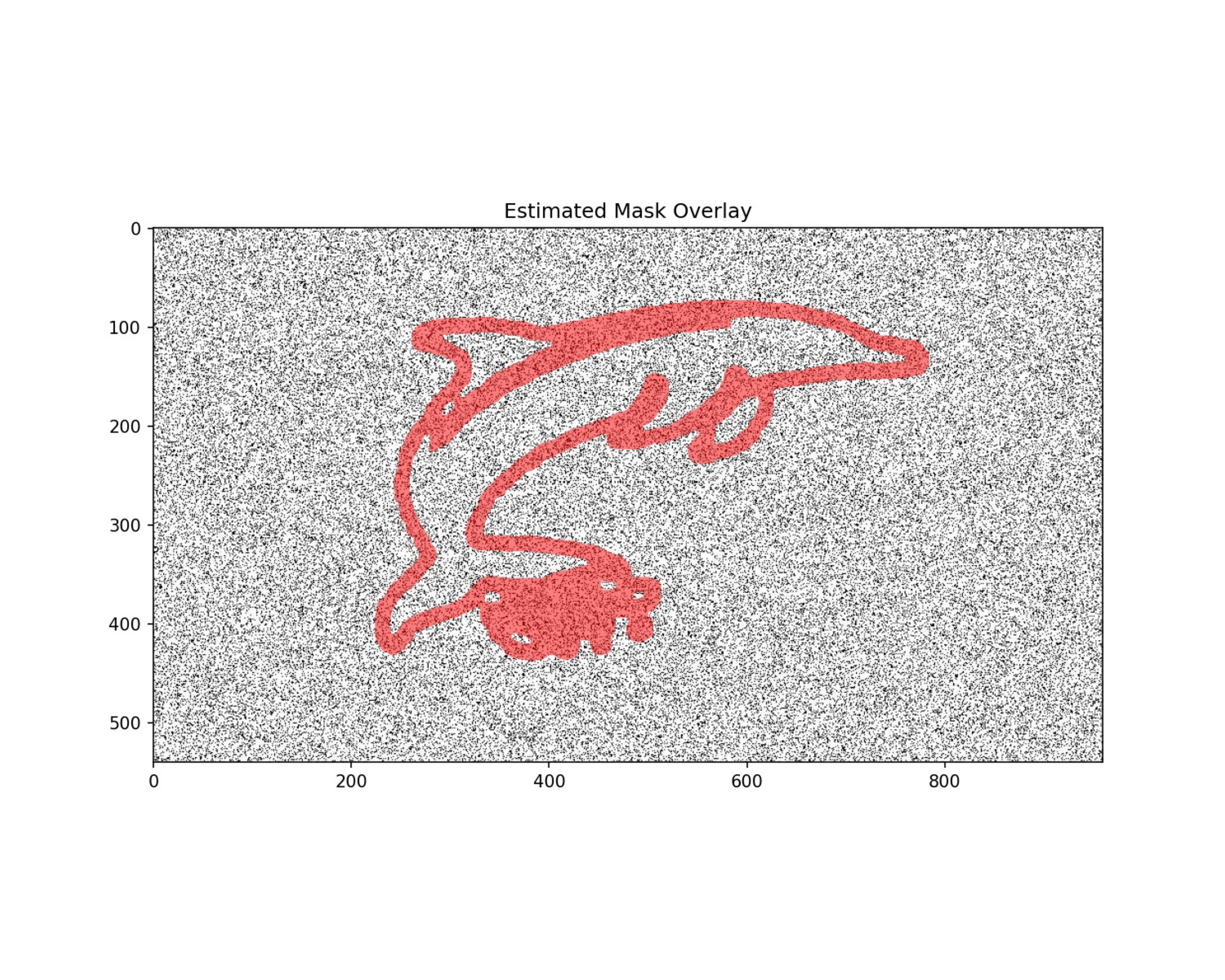} \\
        
        \includegraphics[width=0.22\textwidth]{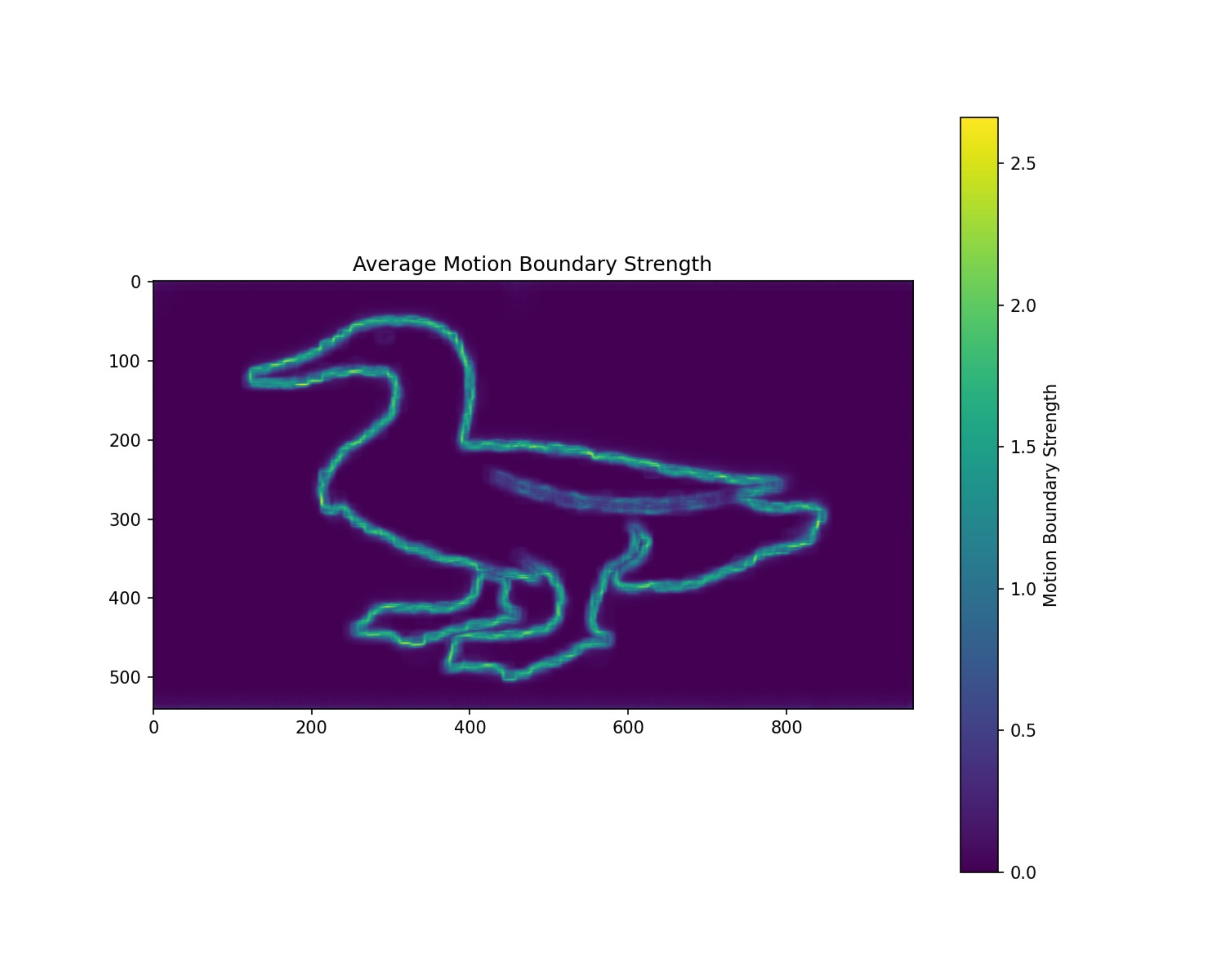} &
        \includegraphics[width=0.22\textwidth]{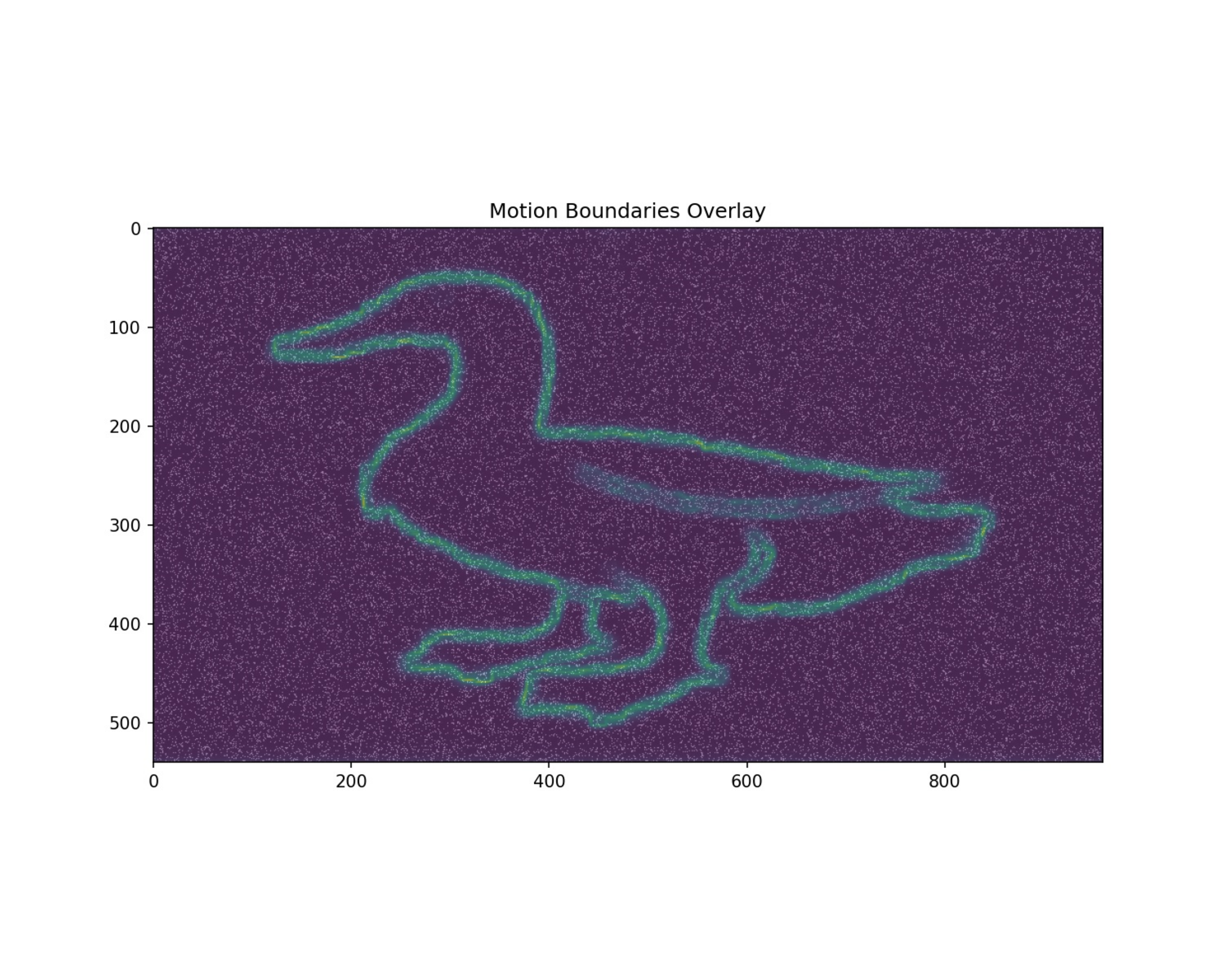} &
        \includegraphics[width=0.22\textwidth]{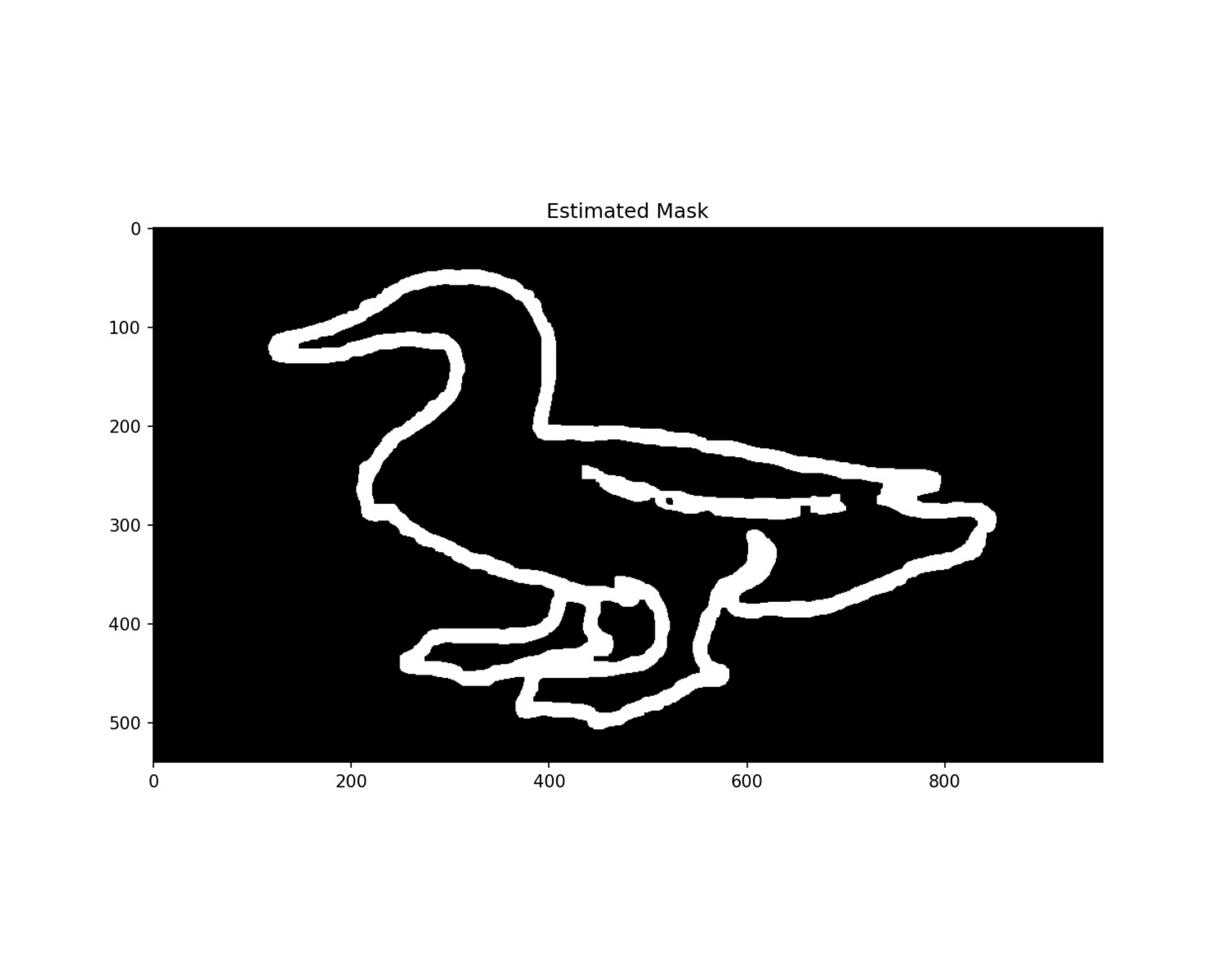} &
        \includegraphics[width=0.22\textwidth]{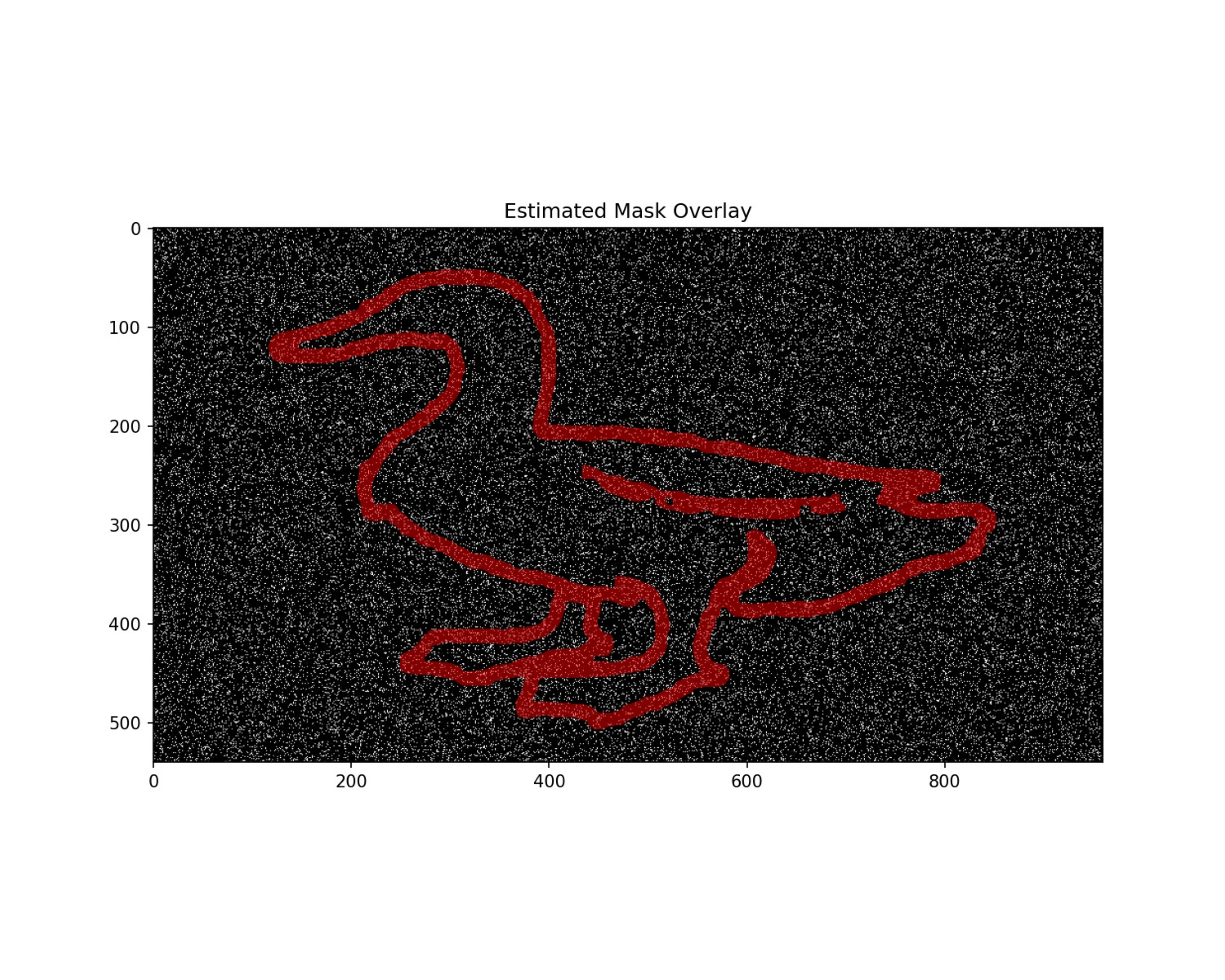} \\
    \end{tabular}
    \caption{Temporal motion coherence analysis for Images category (Part 1). Each row shows motion boundaries, boundary overlay, estimated mask, and mask overlay for: Cycle, Deer, Dolphin, and Duck (top to bottom).}
    \label{fig:images-temporal-analysis-1}
\end{figure*}

\begin{figure*}[t]
    \centering
    \begin{tabular}{@{}c@{\hskip 4pt}c@{\hskip 4pt}c@{\hskip 4pt}c@{}}
        \includegraphics[width=0.22\textwidth]{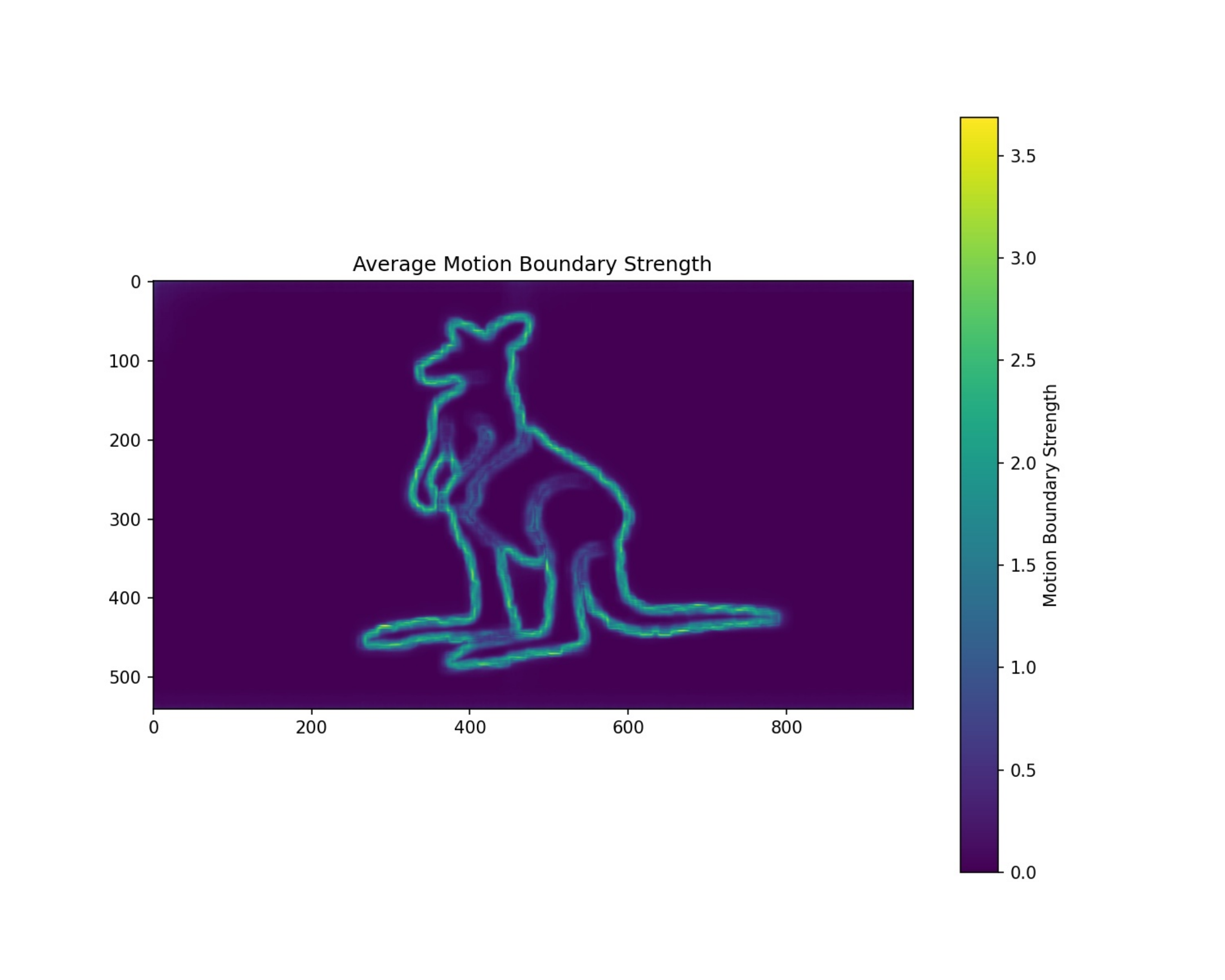} &
        \includegraphics[width=0.22\textwidth]{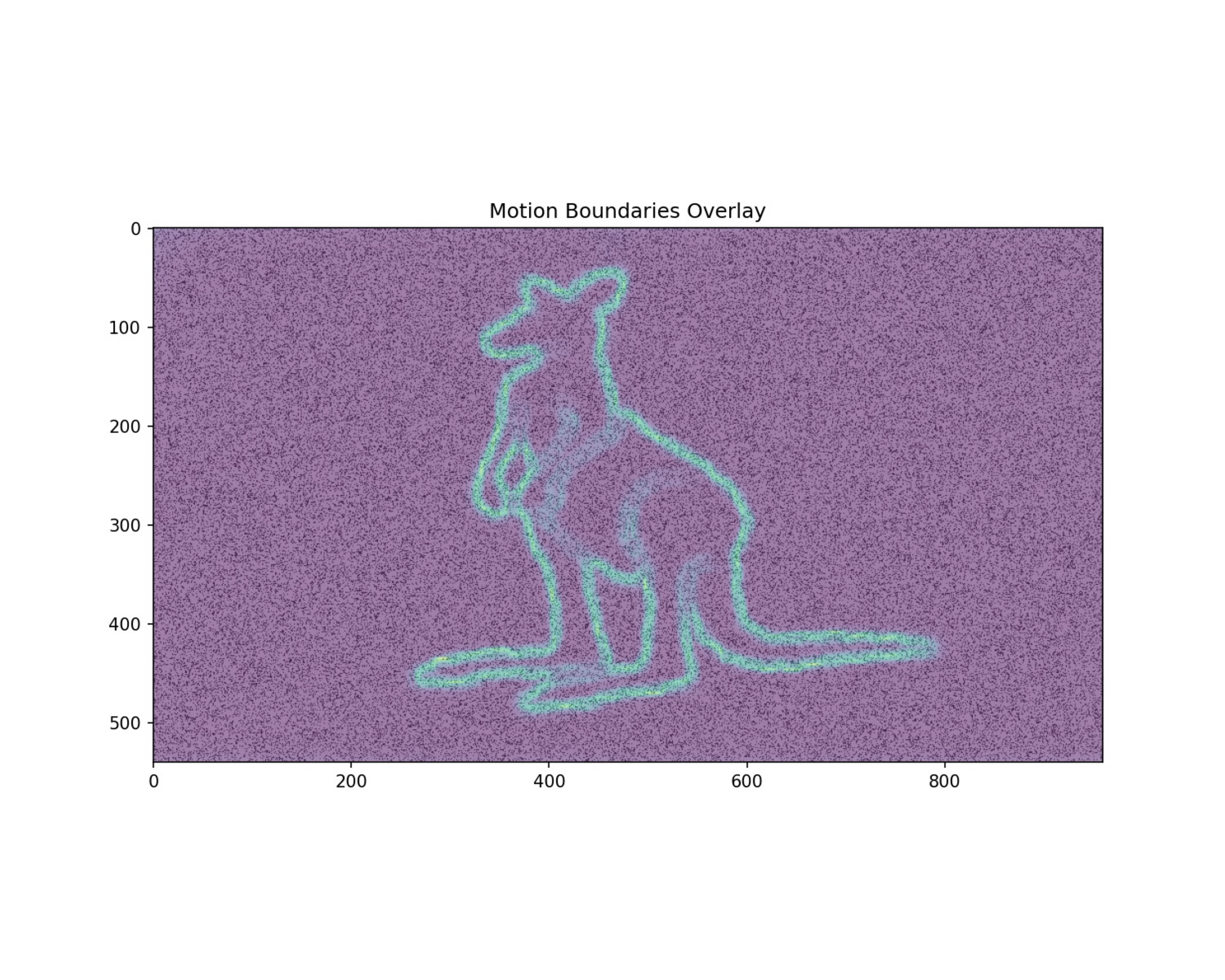} &
        \includegraphics[width=0.22\textwidth]{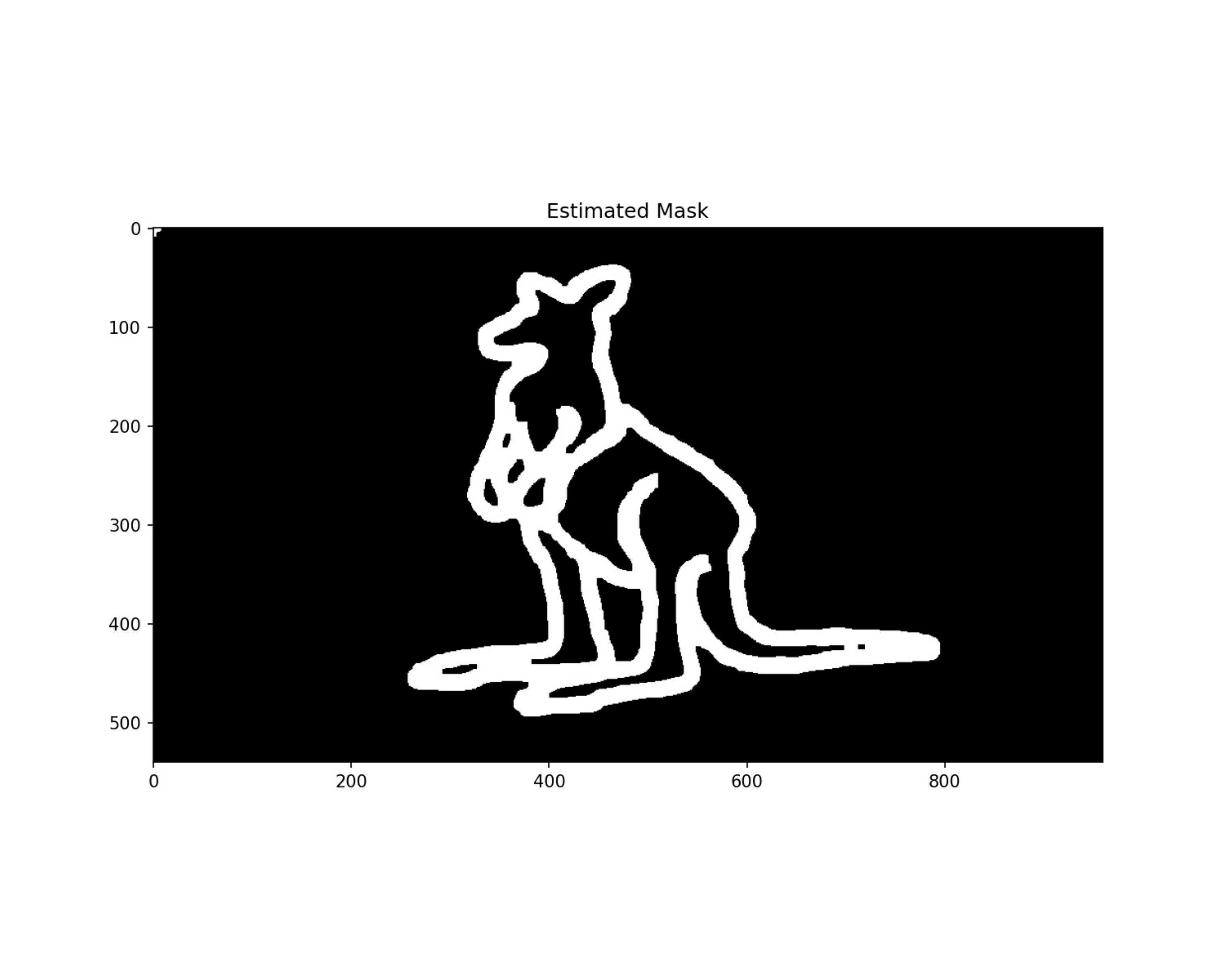} &
        \includegraphics[width=0.22\textwidth]{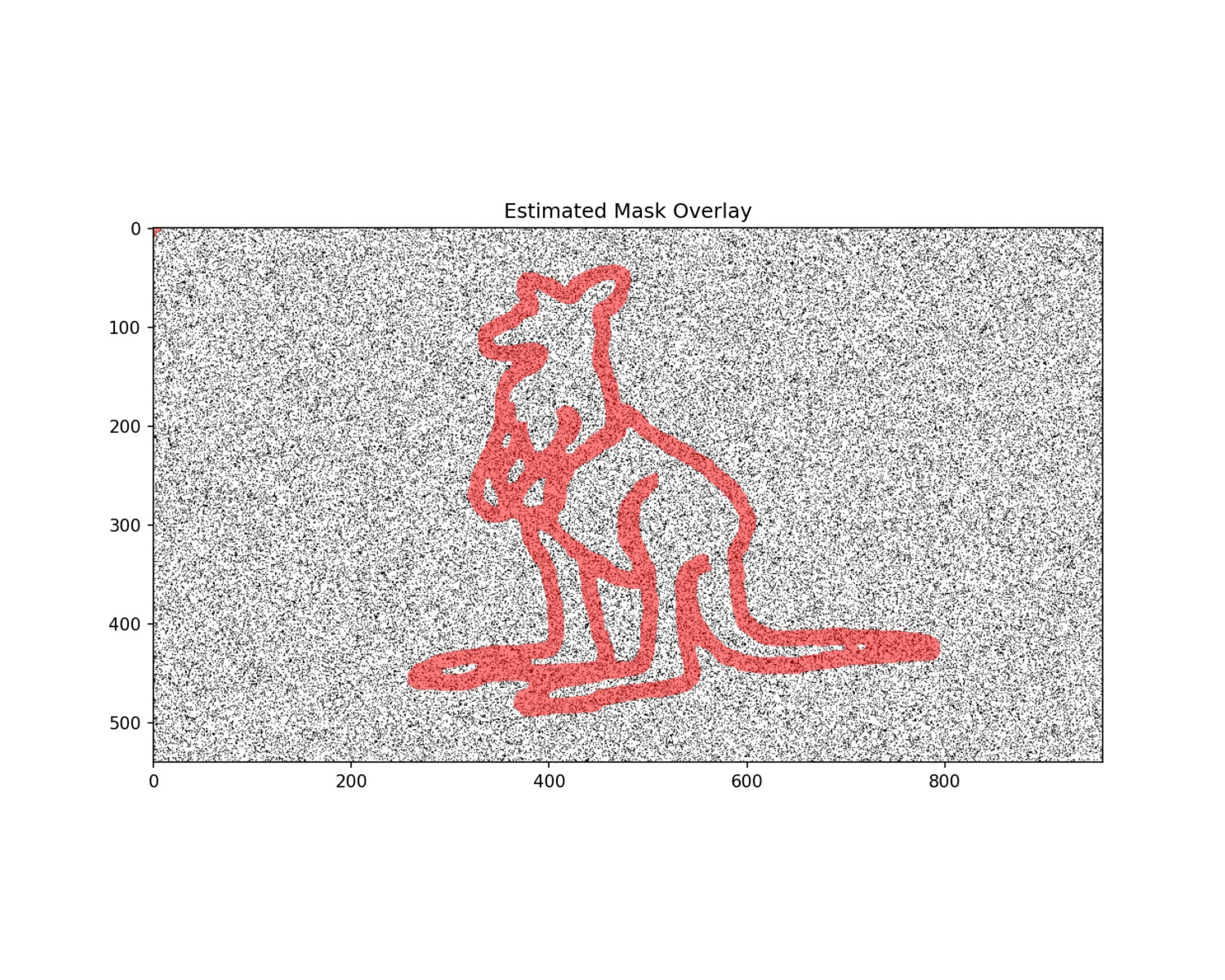} \\
        
        \includegraphics[width=0.22\textwidth]{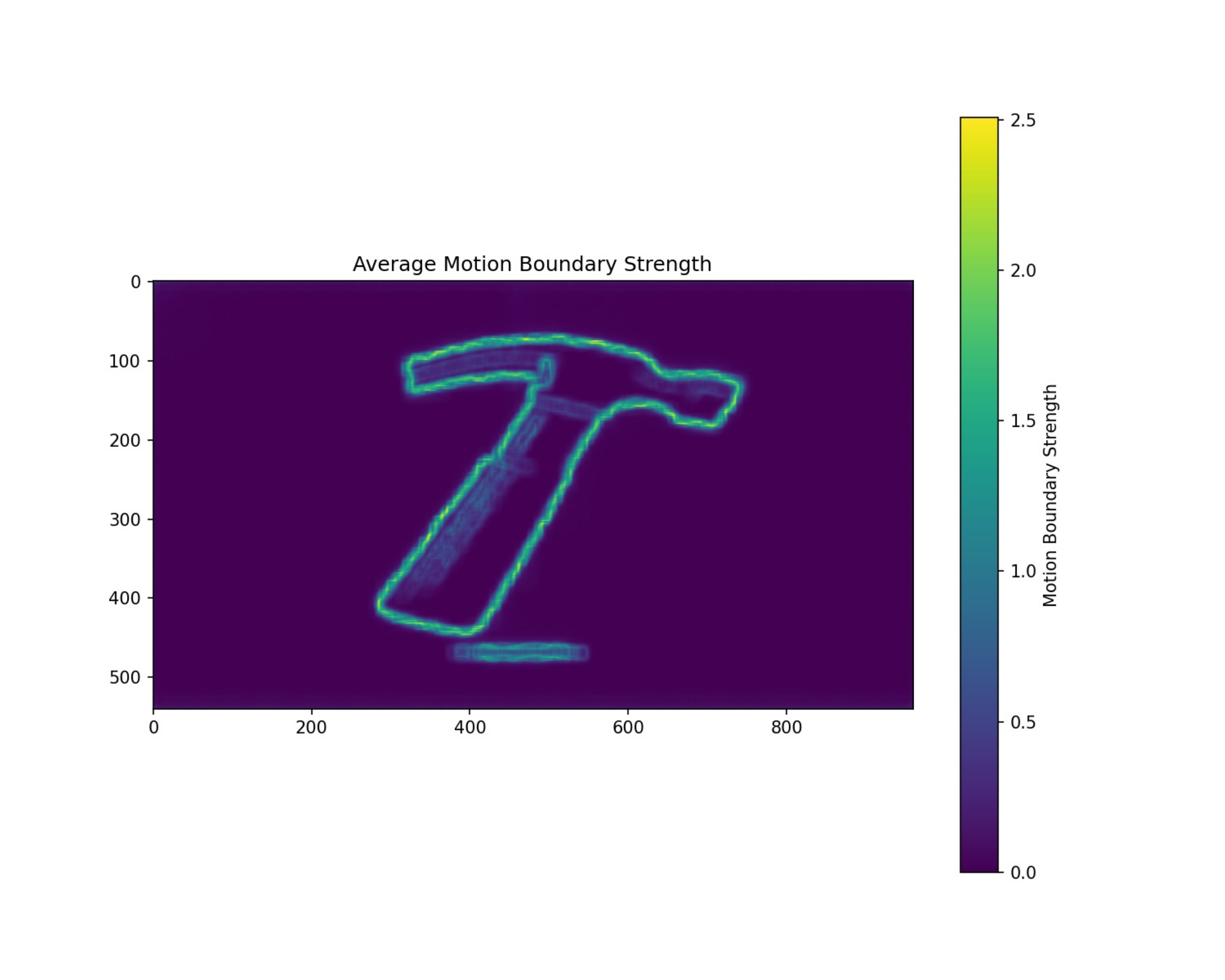} &
        \includegraphics[width=0.22\textwidth]{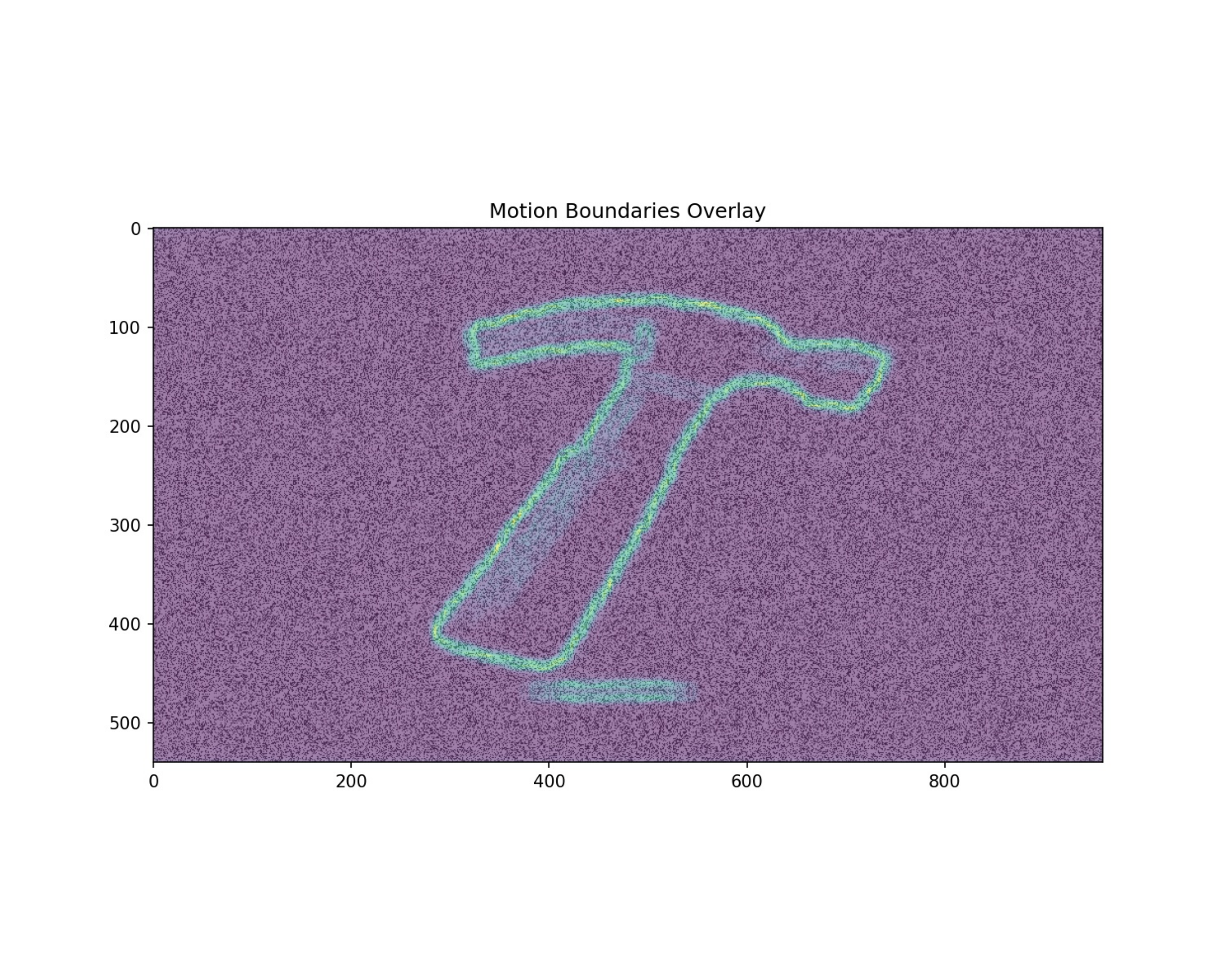} &
        \includegraphics[width=0.22\textwidth]{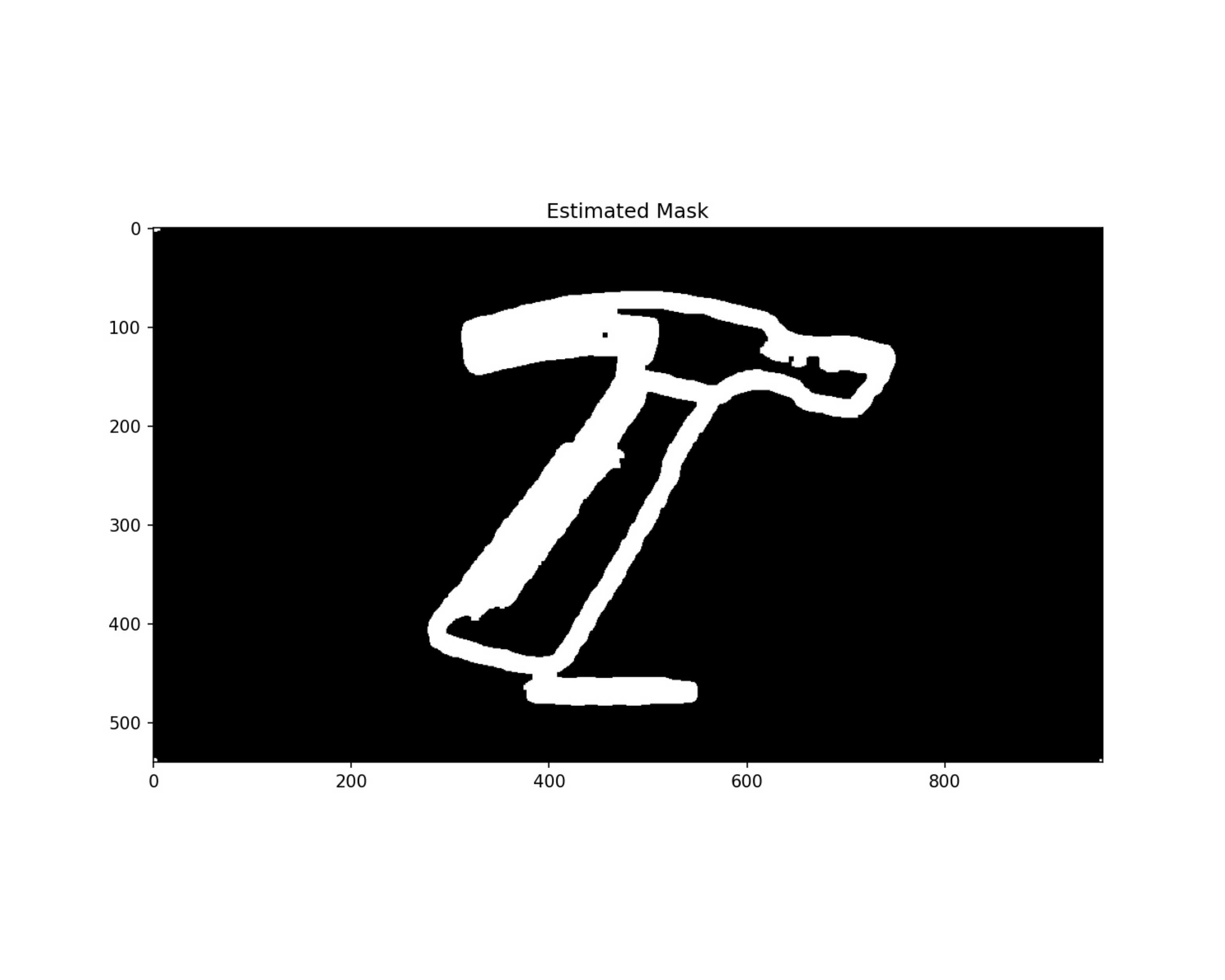} &
        \includegraphics[width=0.22\textwidth]{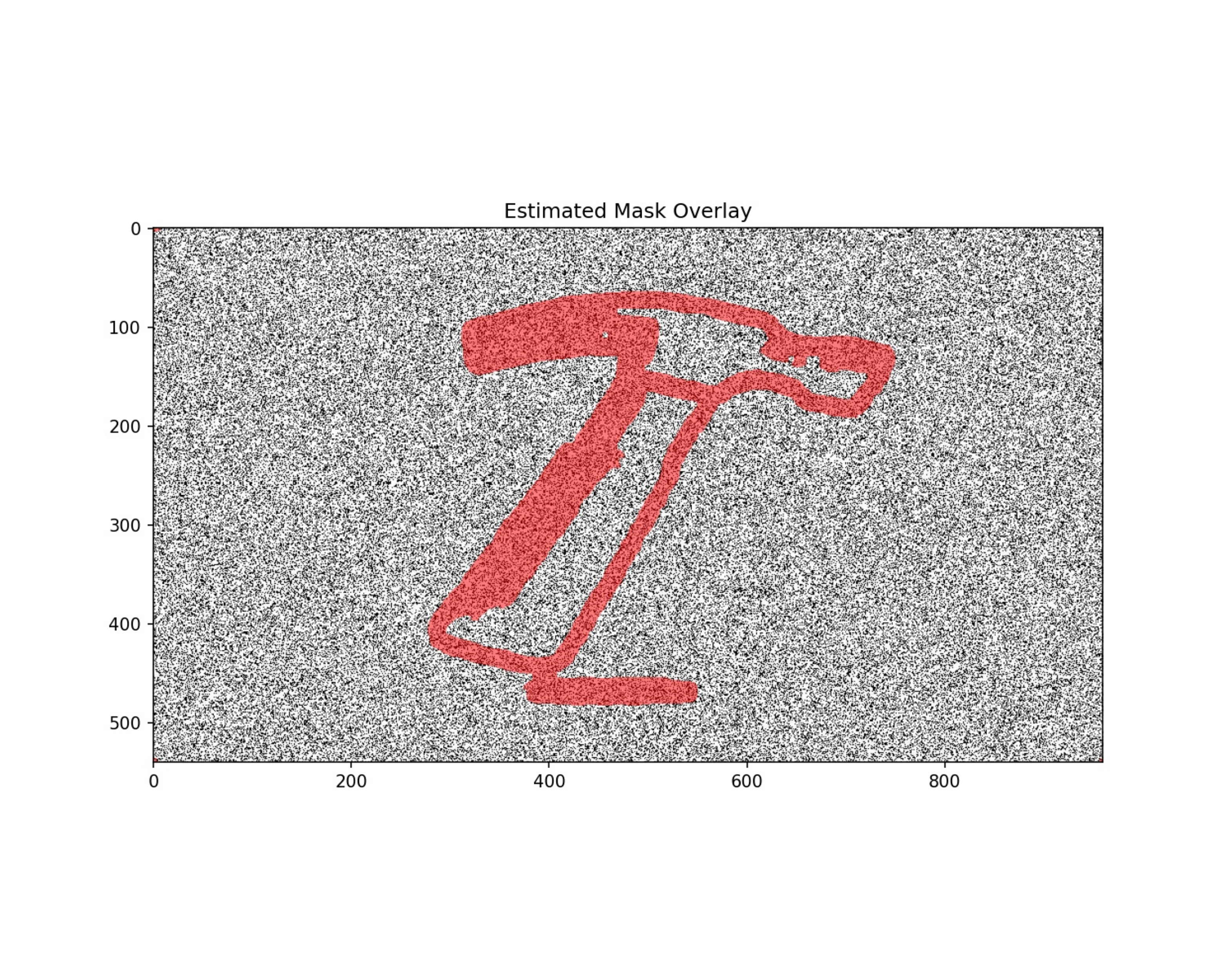} \\
        
        \includegraphics[width=0.22\textwidth]{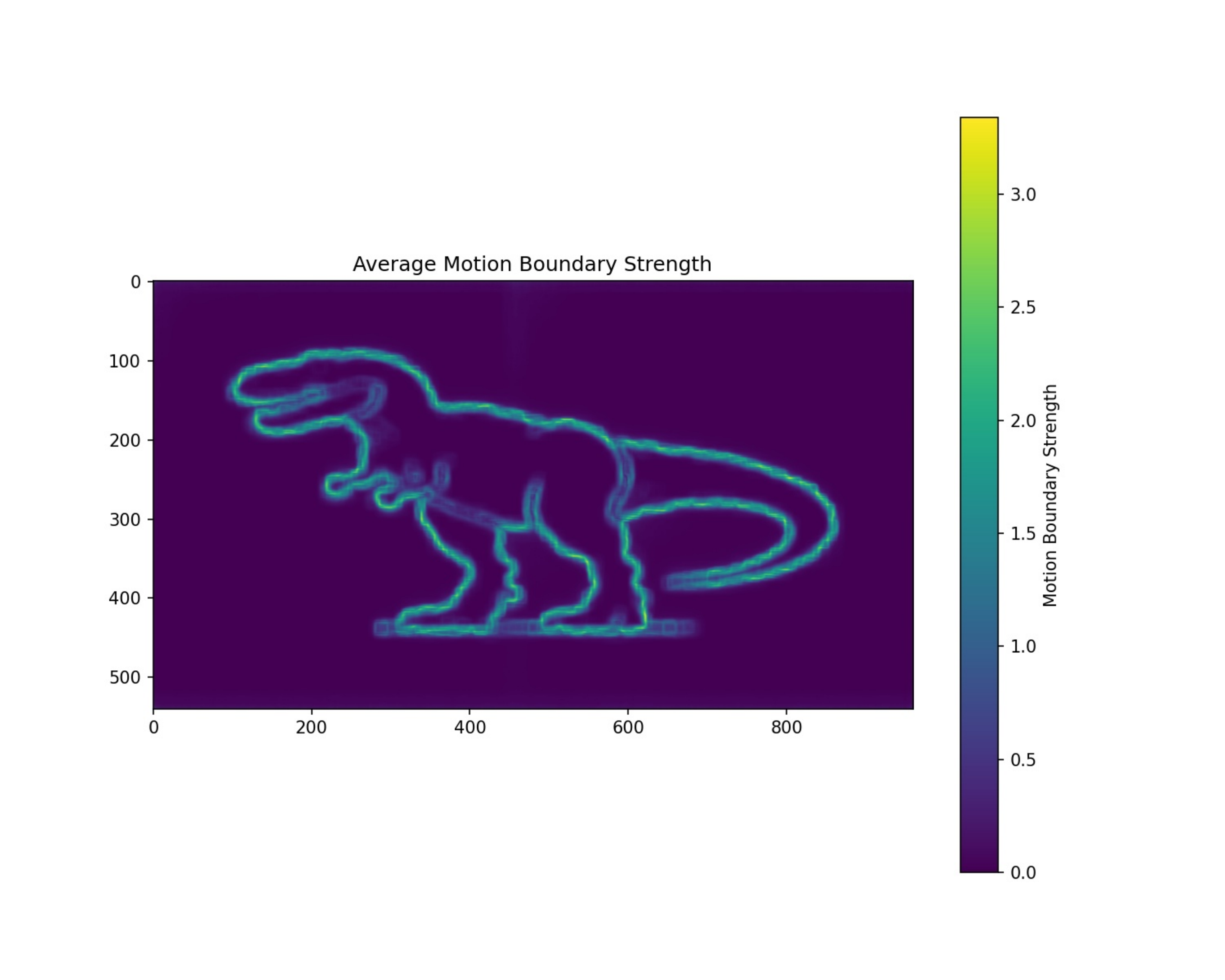} &
        \includegraphics[width=0.22\textwidth]{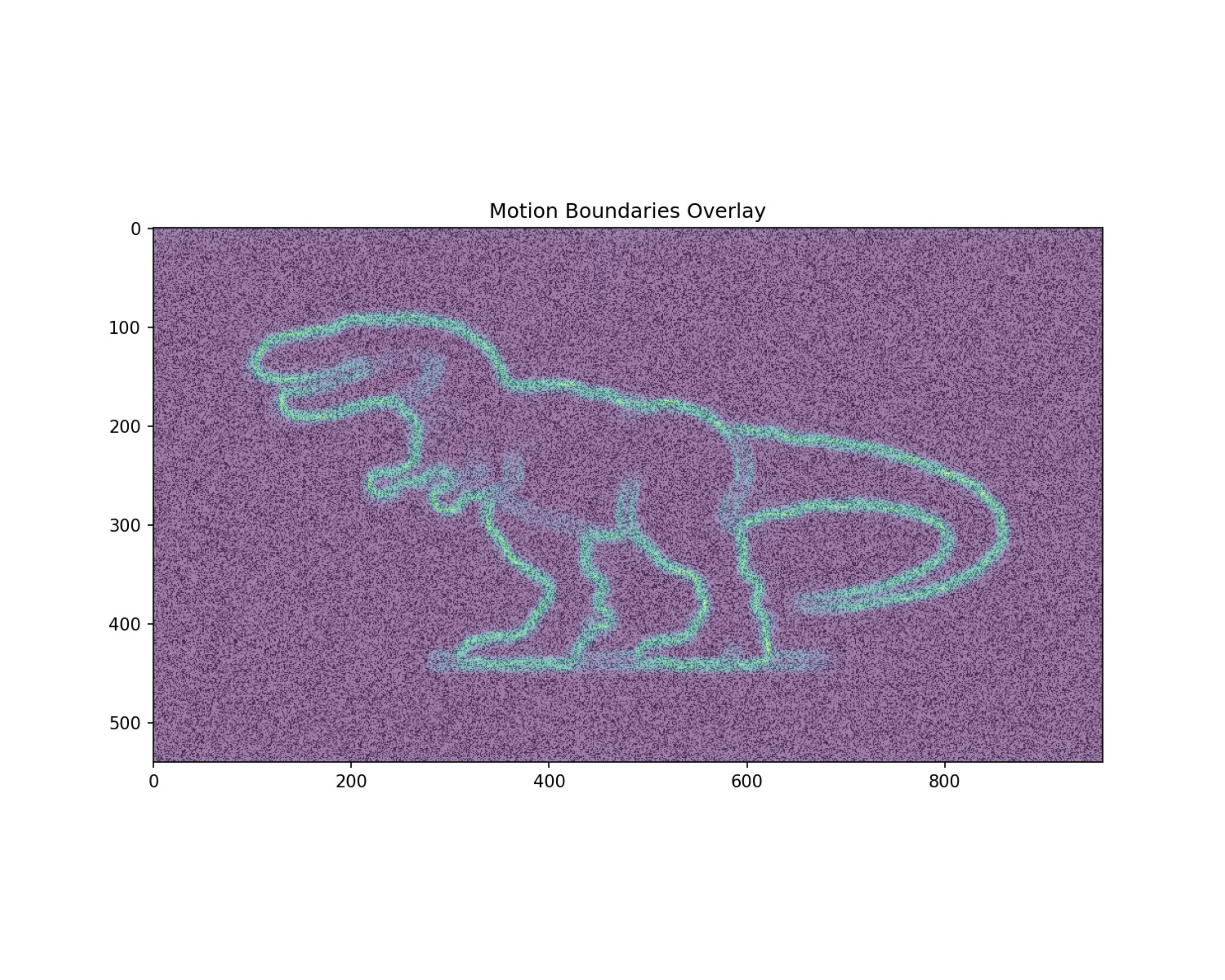} &
        \includegraphics[width=0.22\textwidth]{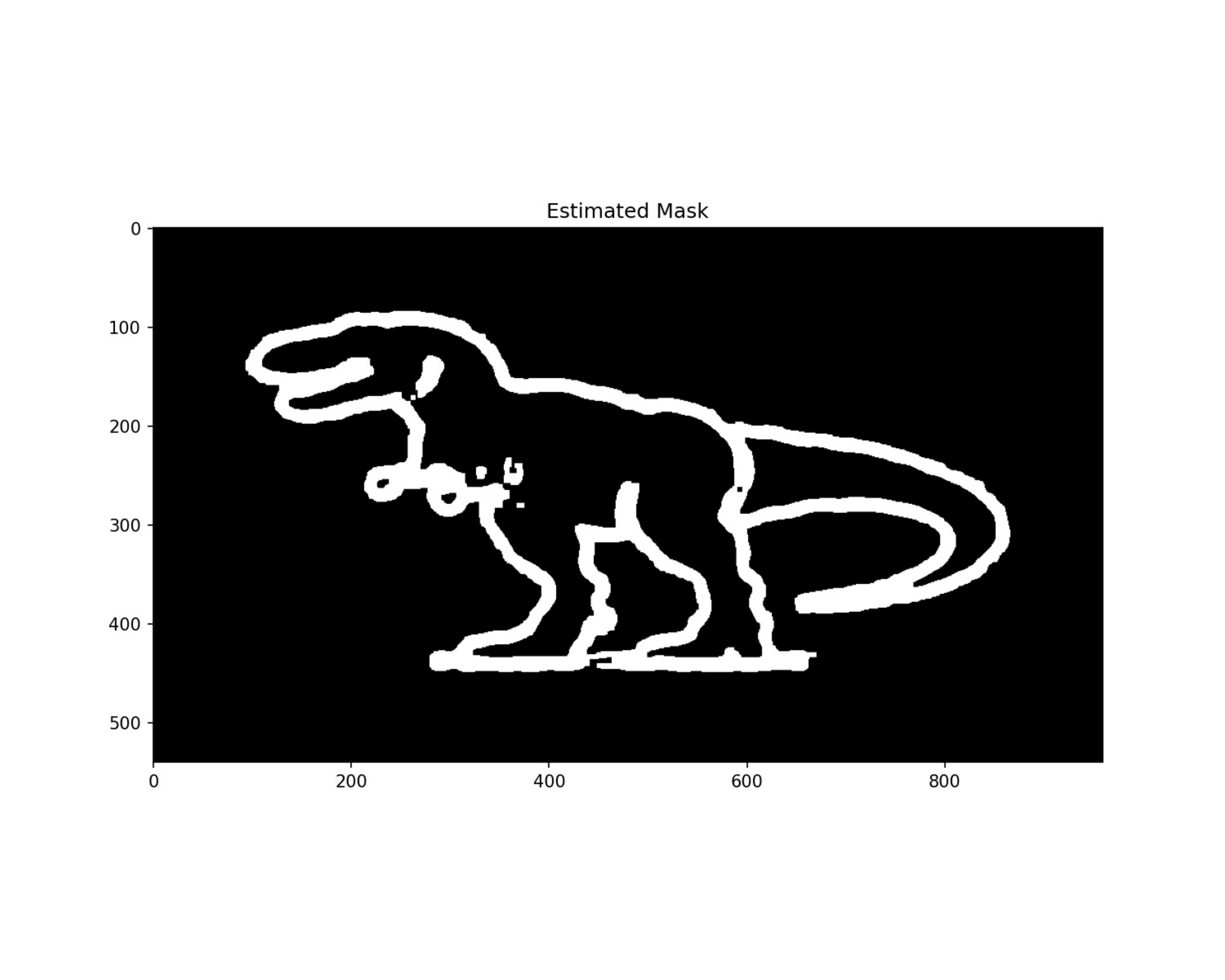} &
        \includegraphics[width=0.22\textwidth]{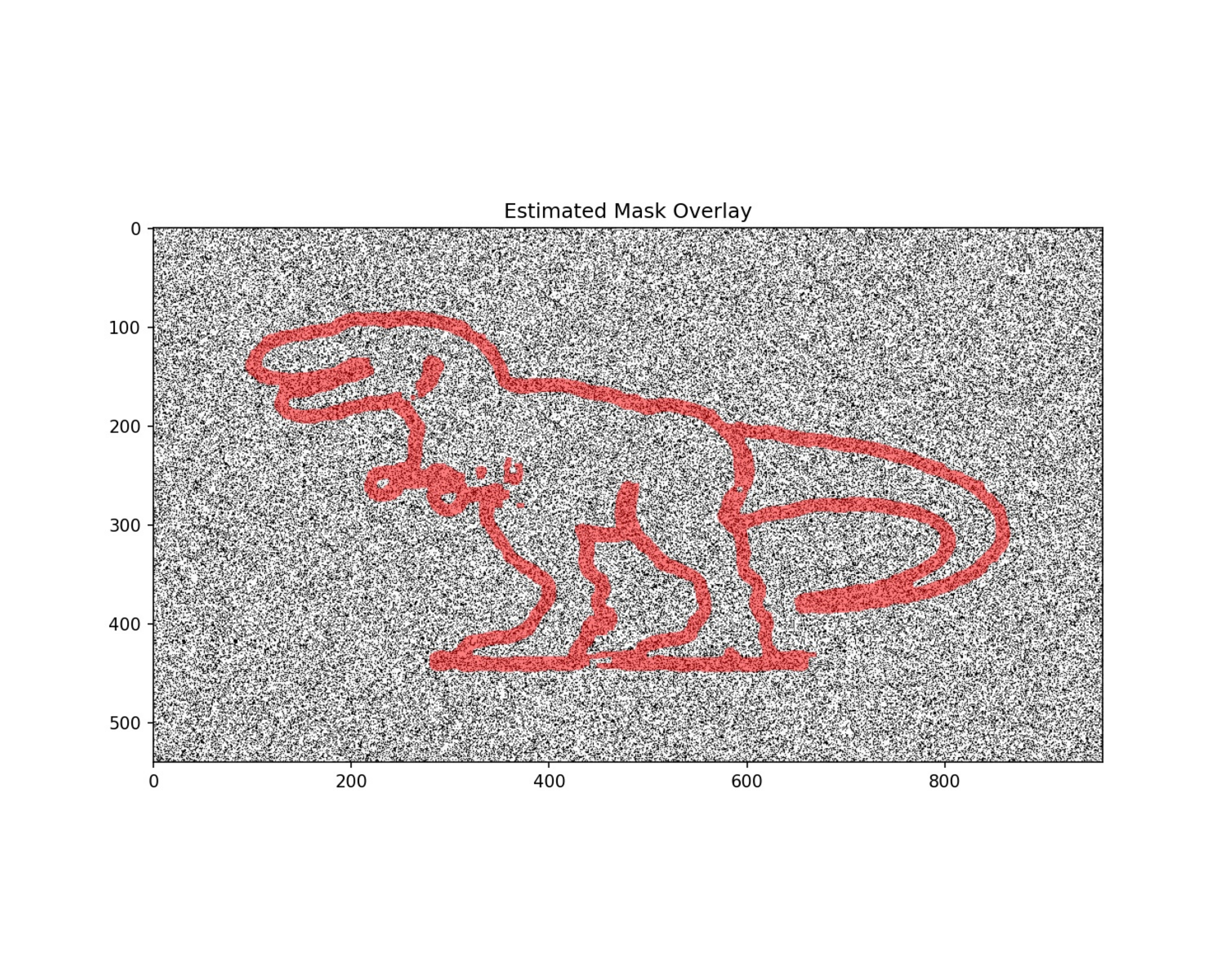} \\
        
        \includegraphics[width=0.22\textwidth]{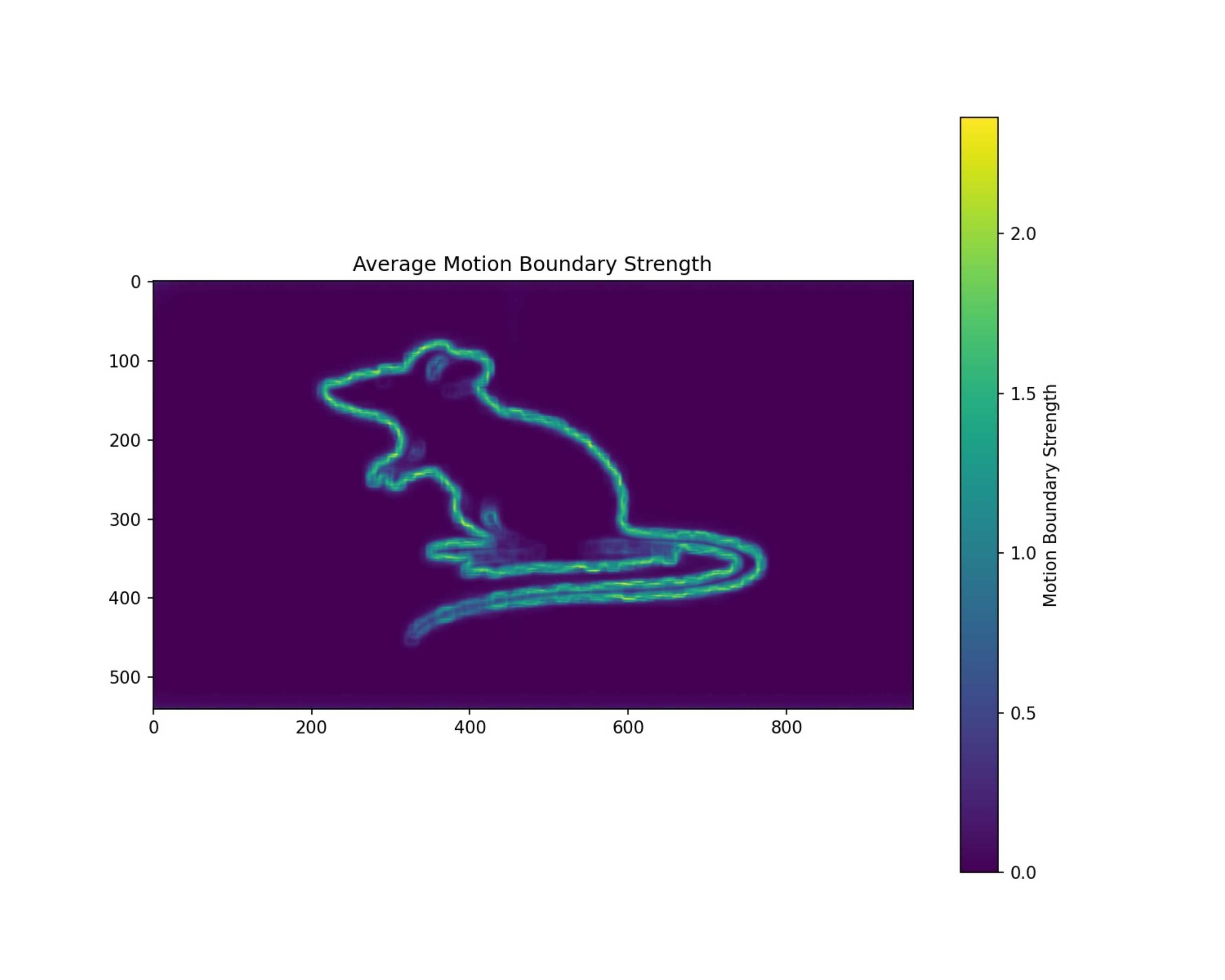} &
        \includegraphics[width=0.22\textwidth]{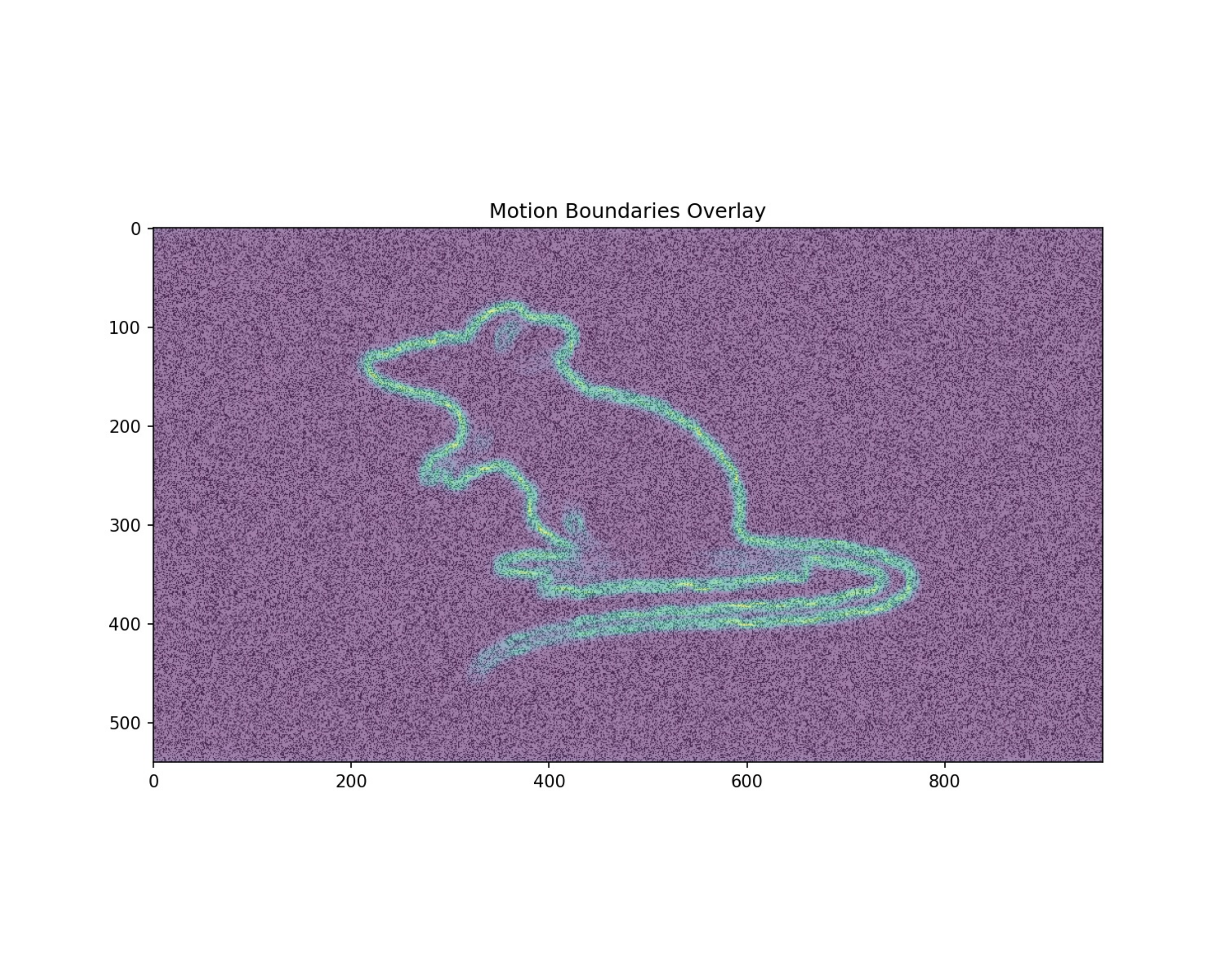} &
        \includegraphics[width=0.22\textwidth]{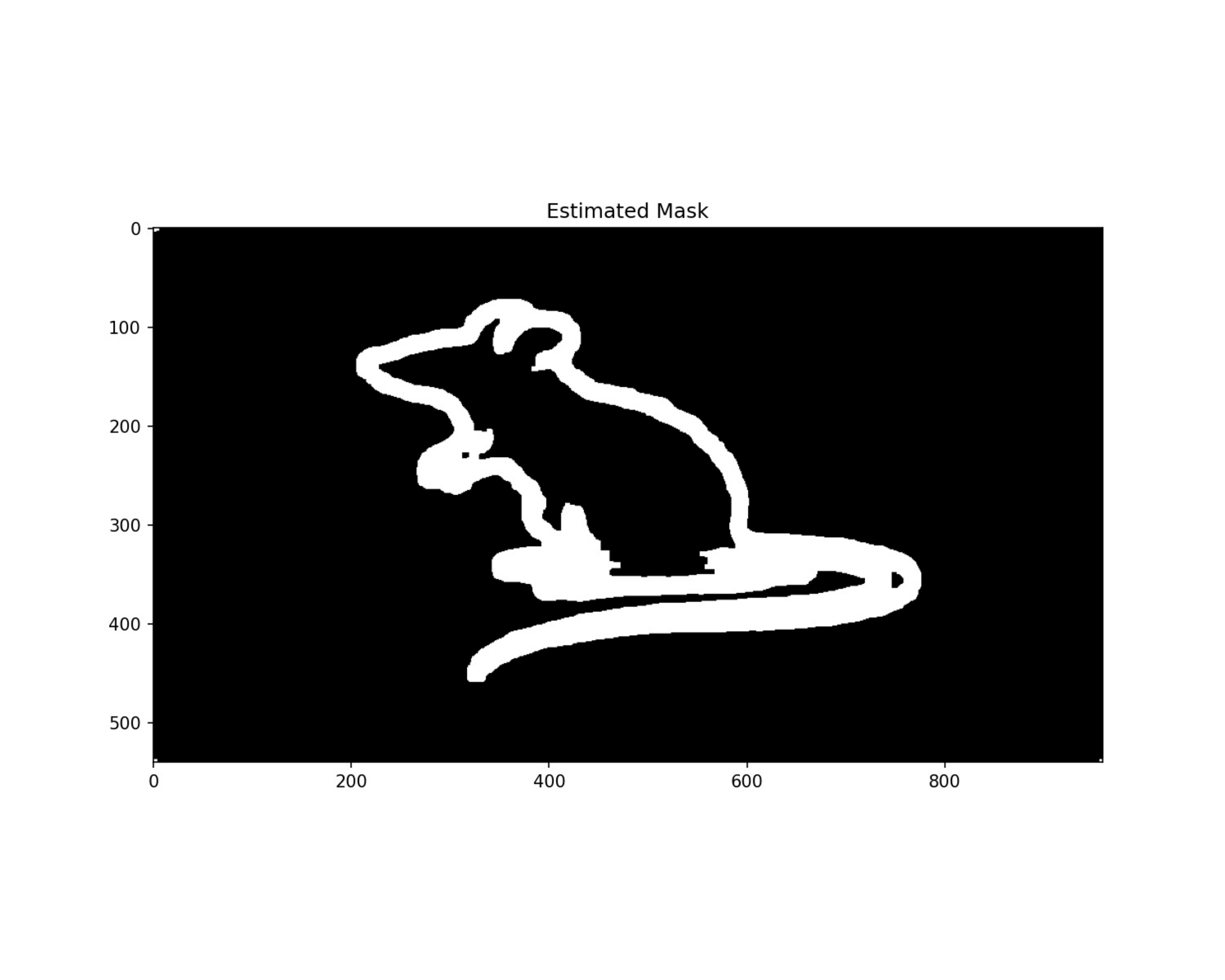} &
        \includegraphics[width=0.22\textwidth]{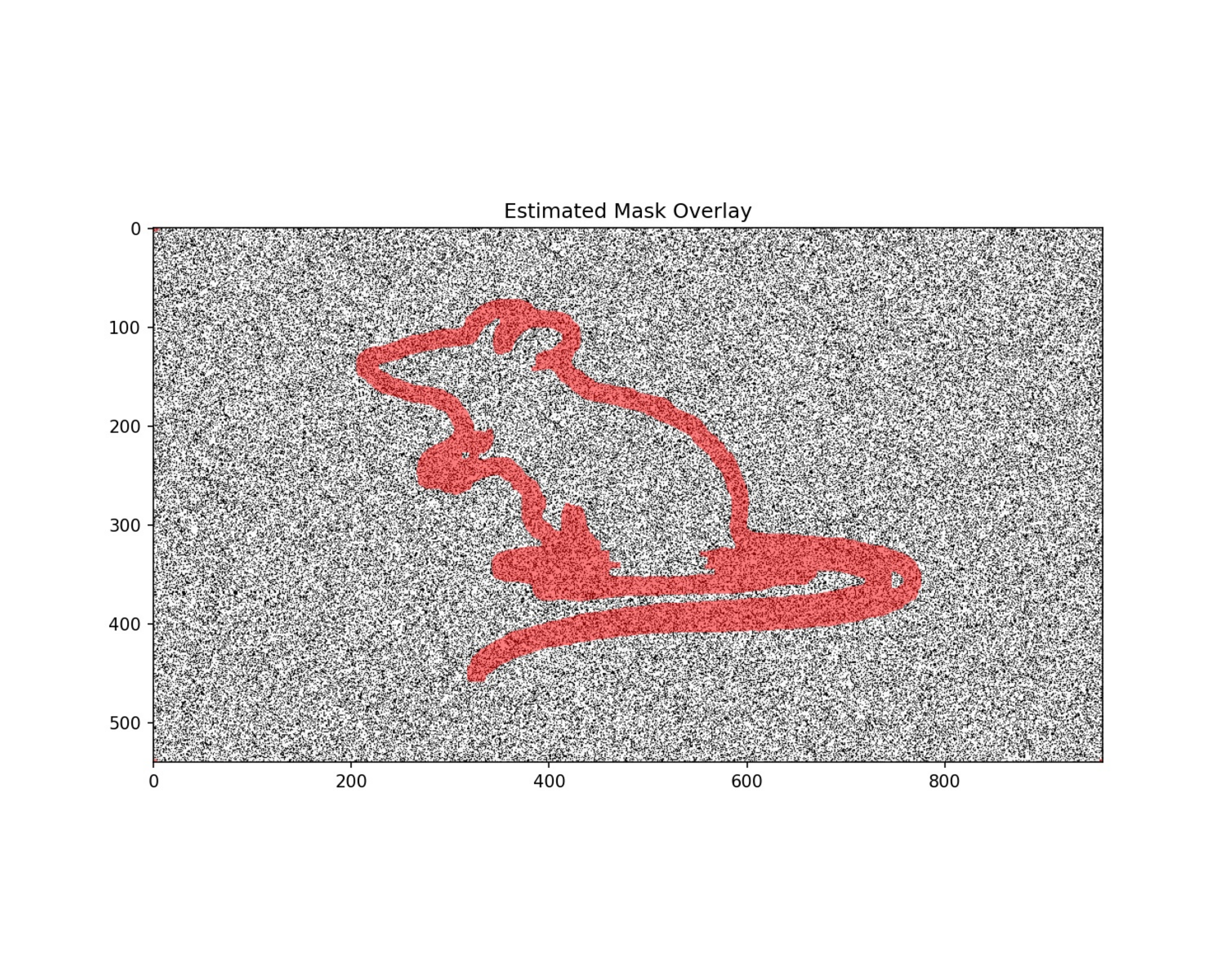} \\
    \end{tabular}
    \caption{Temporal motion coherence analysis for Images category (Part 2). Each row shows motion boundaries, boundary overlay, estimated mask, and mask overlay for: Kangaroo, Hammer, T-Rex, and Mouse (top to bottom).}
    \label{fig:images-temporal-analysis-2}
\end{figure*}

\begin{figure*}[t]
    \centering
    \begin{tabular}{@{}c@{\hskip 4pt}c@{\hskip 4pt}c@{\hskip 4pt}c@{}}
        \includegraphics[width=0.22\textwidth]{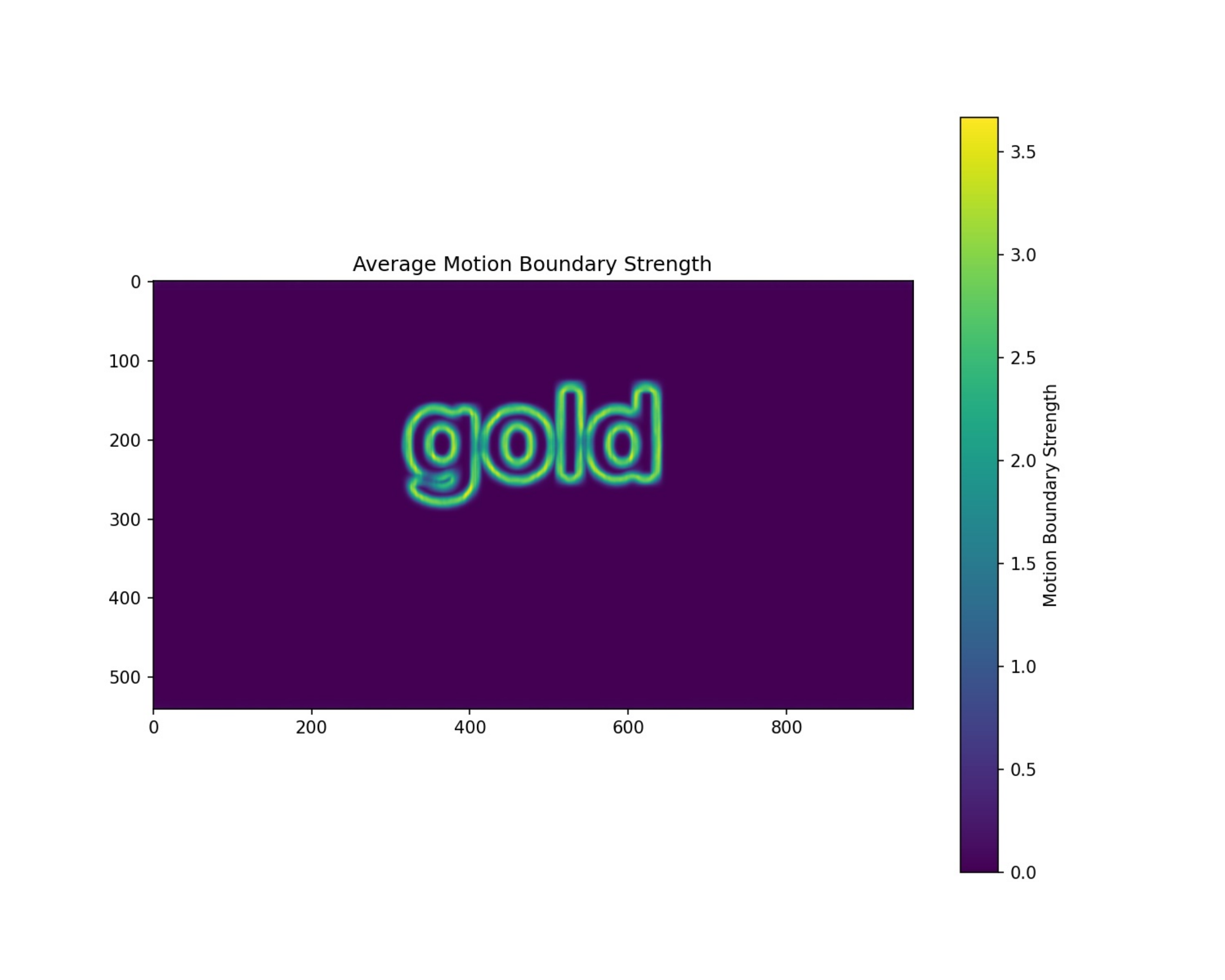} &
        \includegraphics[width=0.22\textwidth]{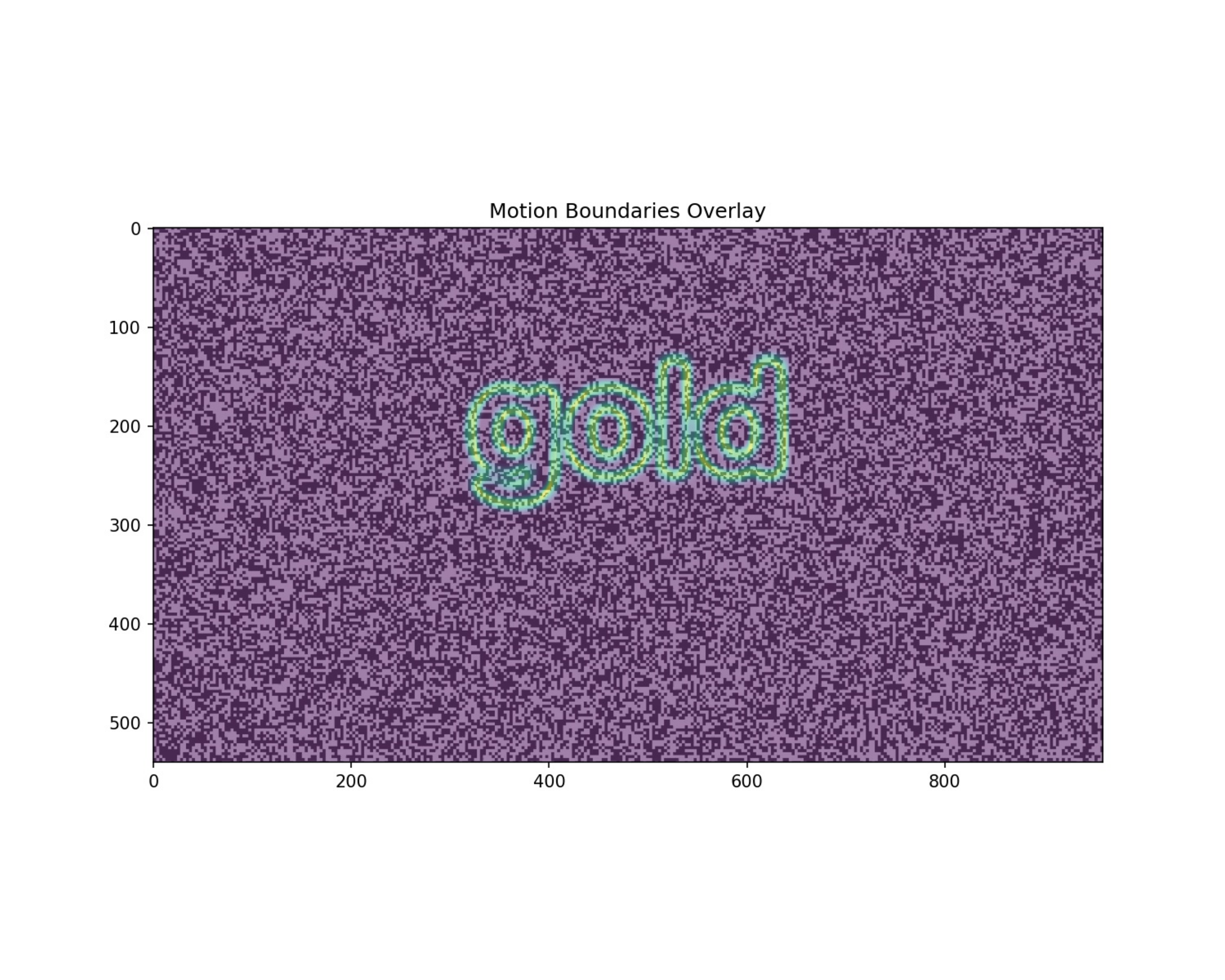} &
        \includegraphics[width=0.22\textwidth]{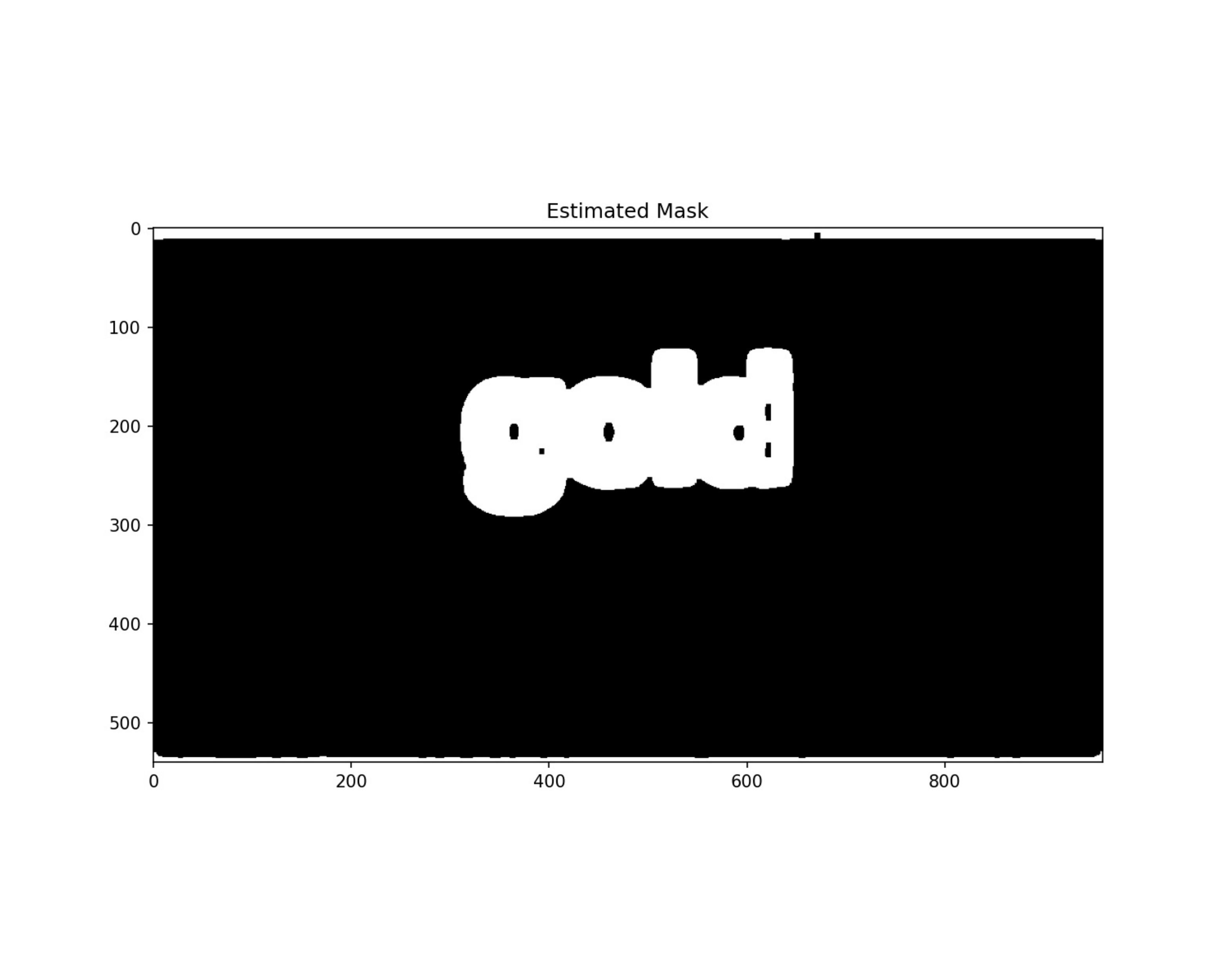} &
        \includegraphics[width=0.22\textwidth]{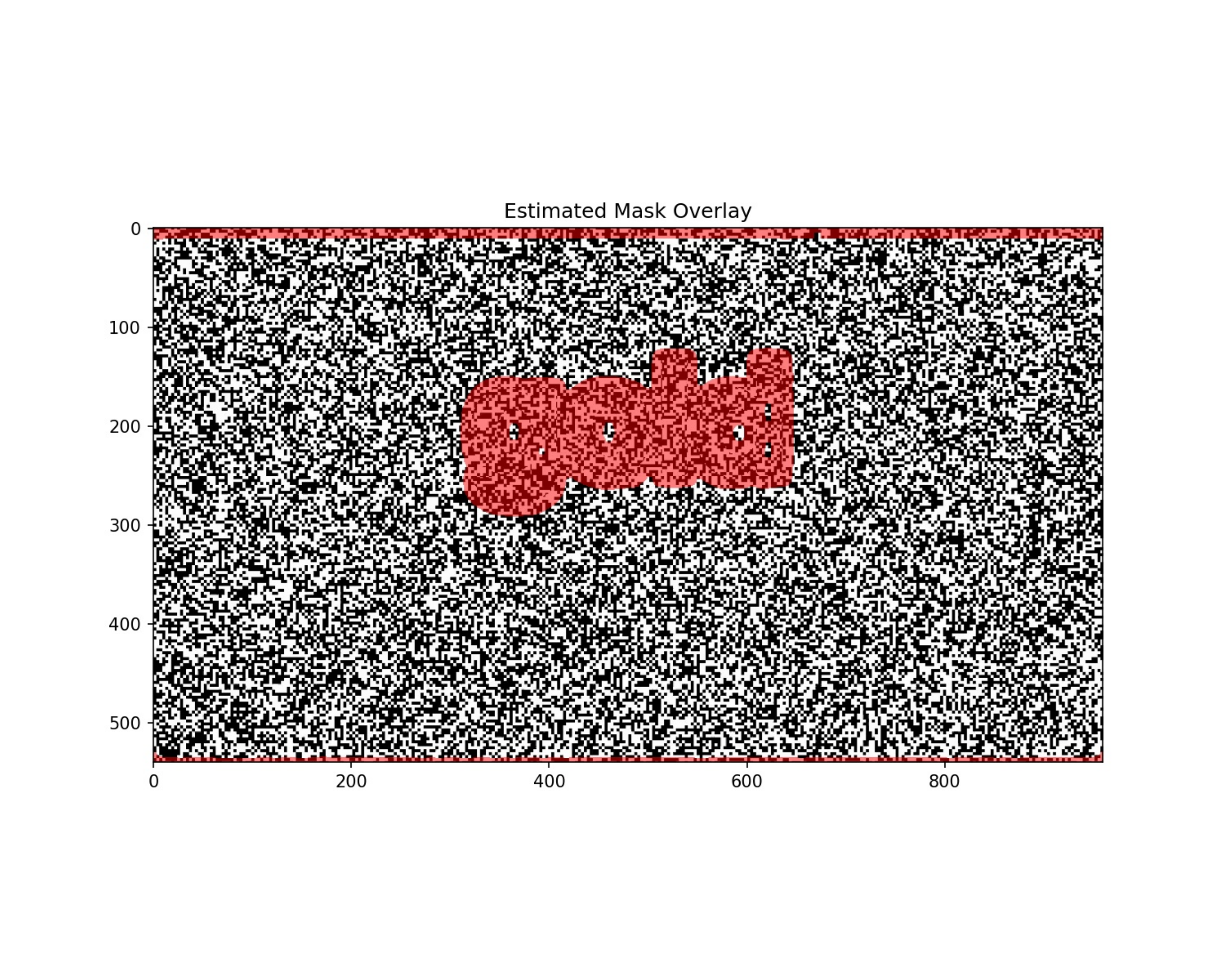} \\
        
        \includegraphics[width=0.22\textwidth]{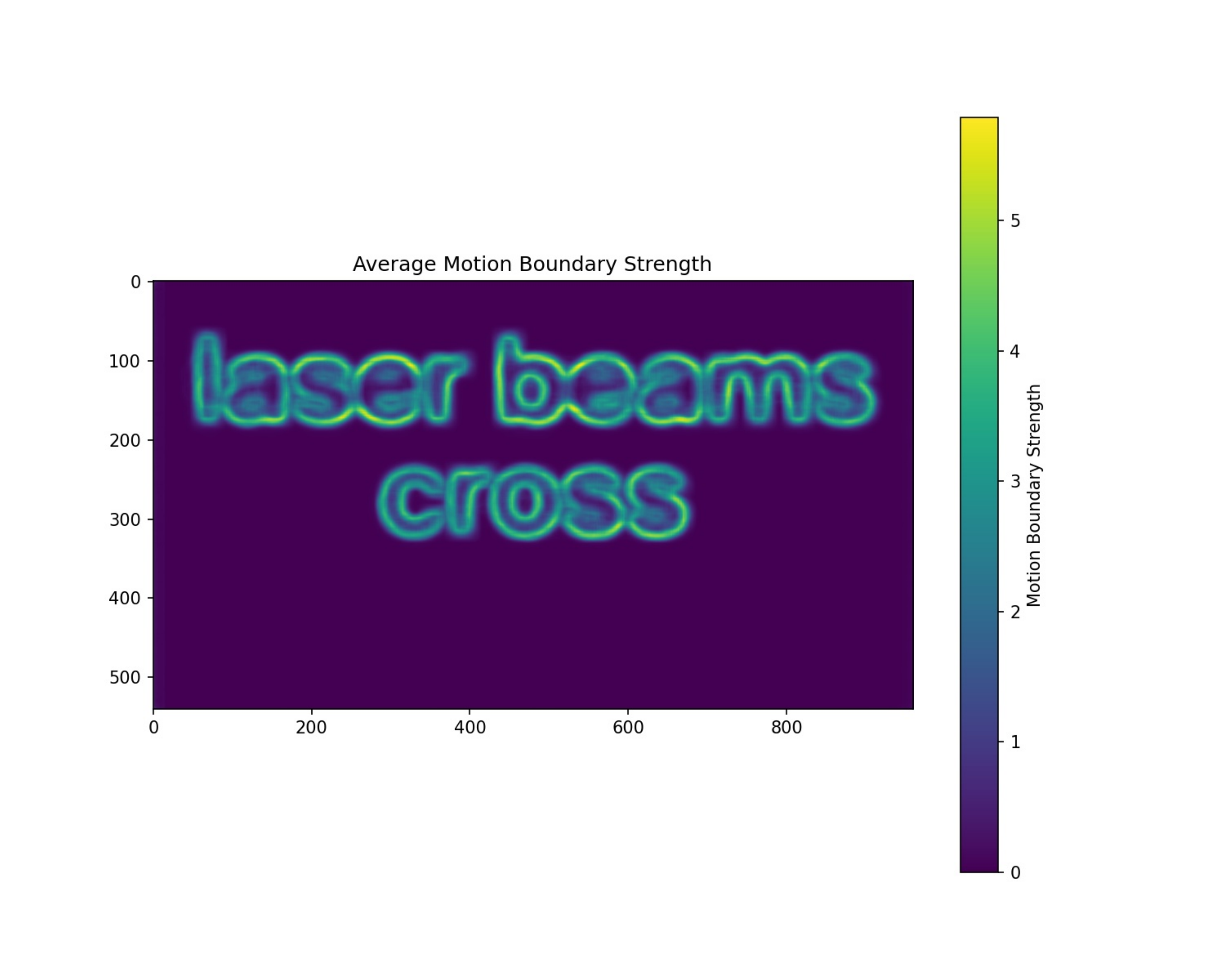} &
        \includegraphics[width=0.22\textwidth]{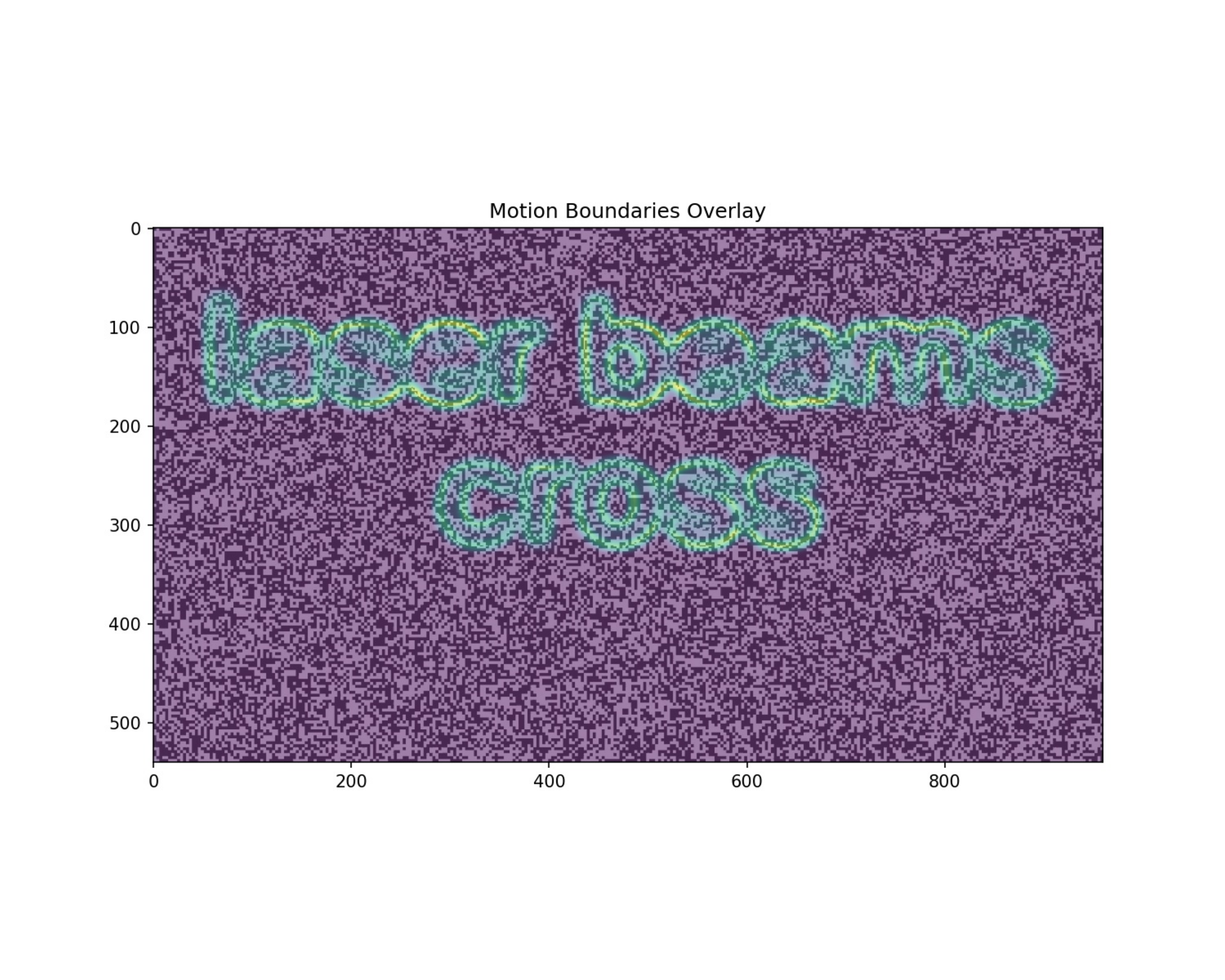} &
        \includegraphics[width=0.22\textwidth]{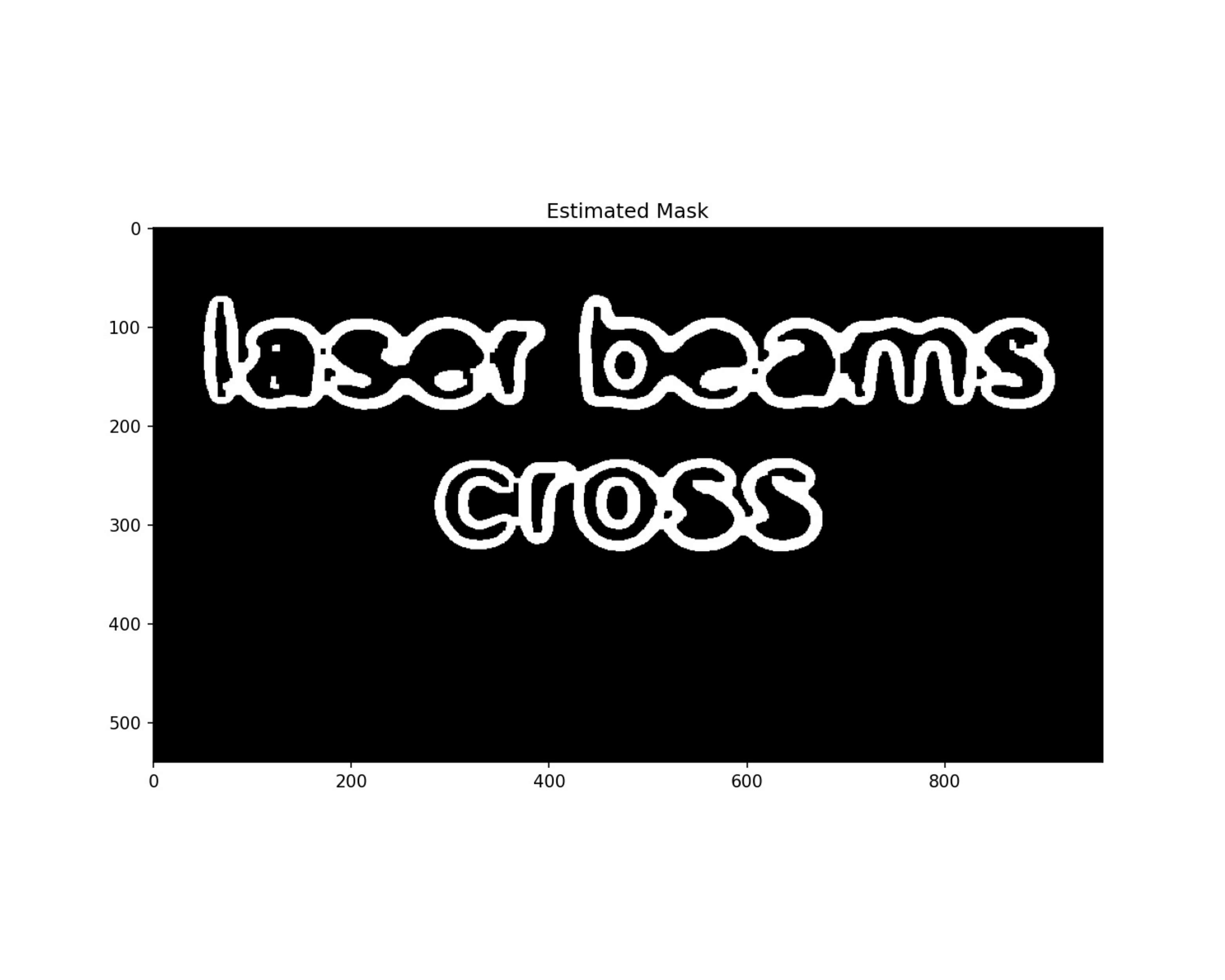} &
        \includegraphics[width=0.22\textwidth]{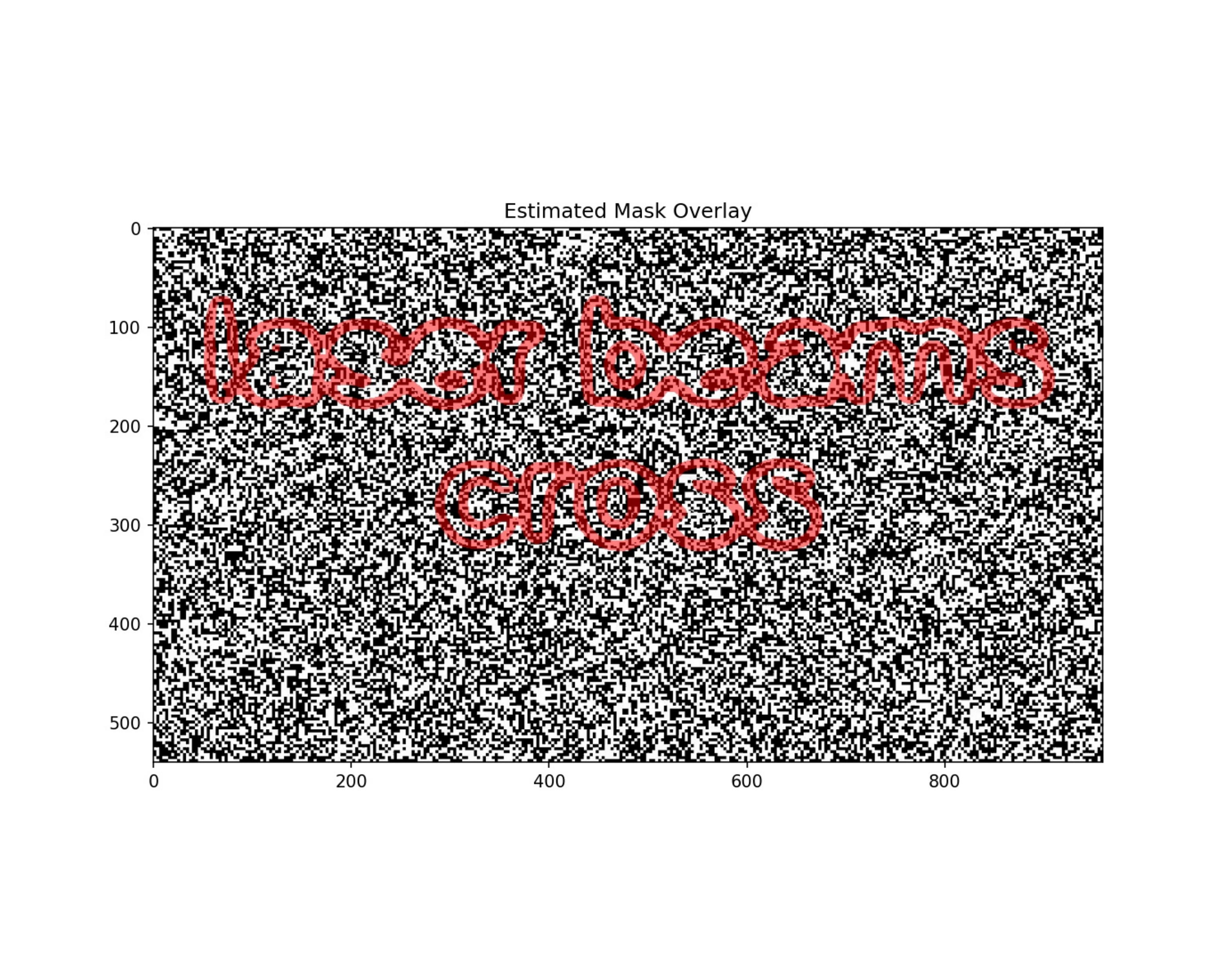} \\
        
        \includegraphics[width=0.22\textwidth]{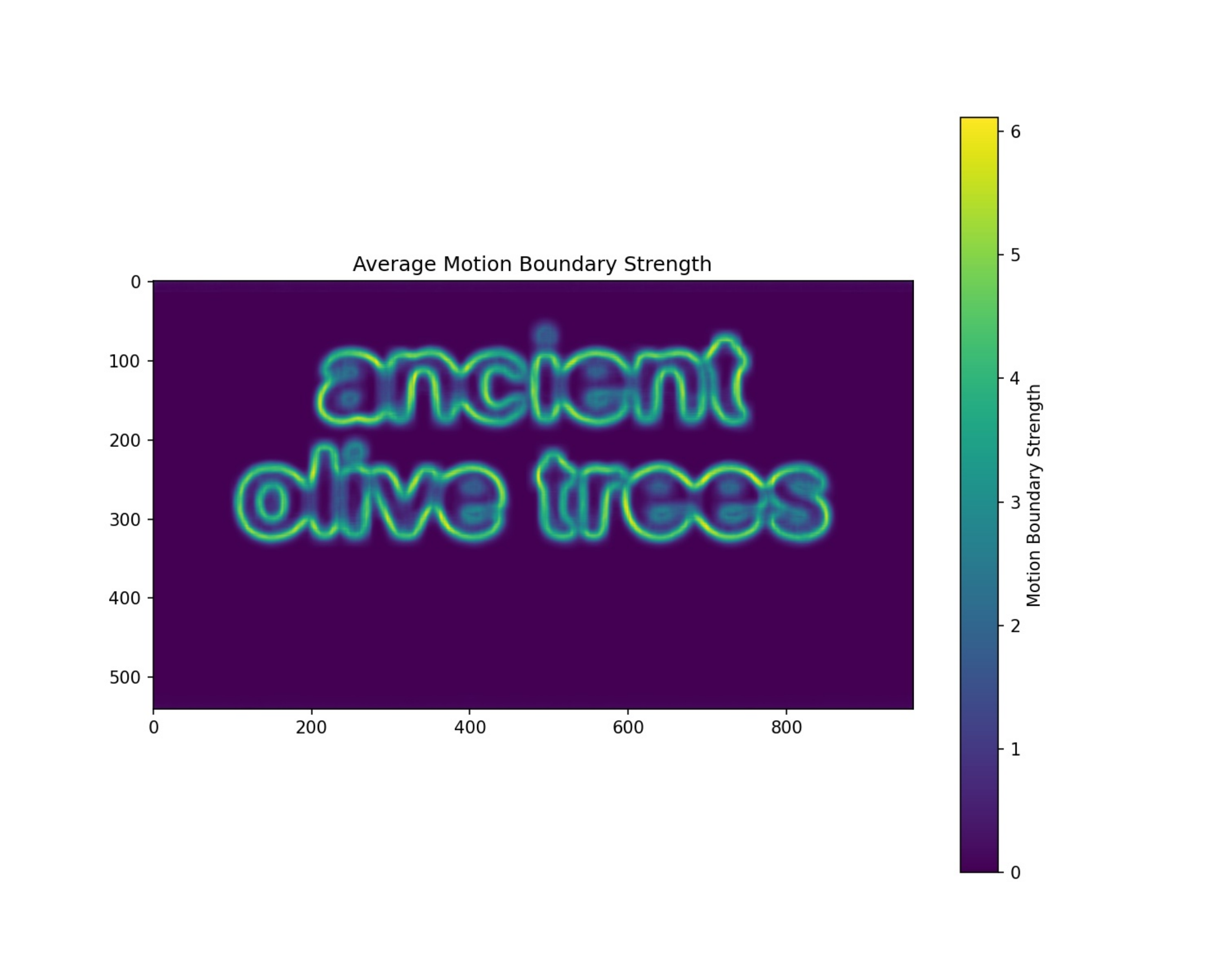} &
        \includegraphics[width=0.22\textwidth]{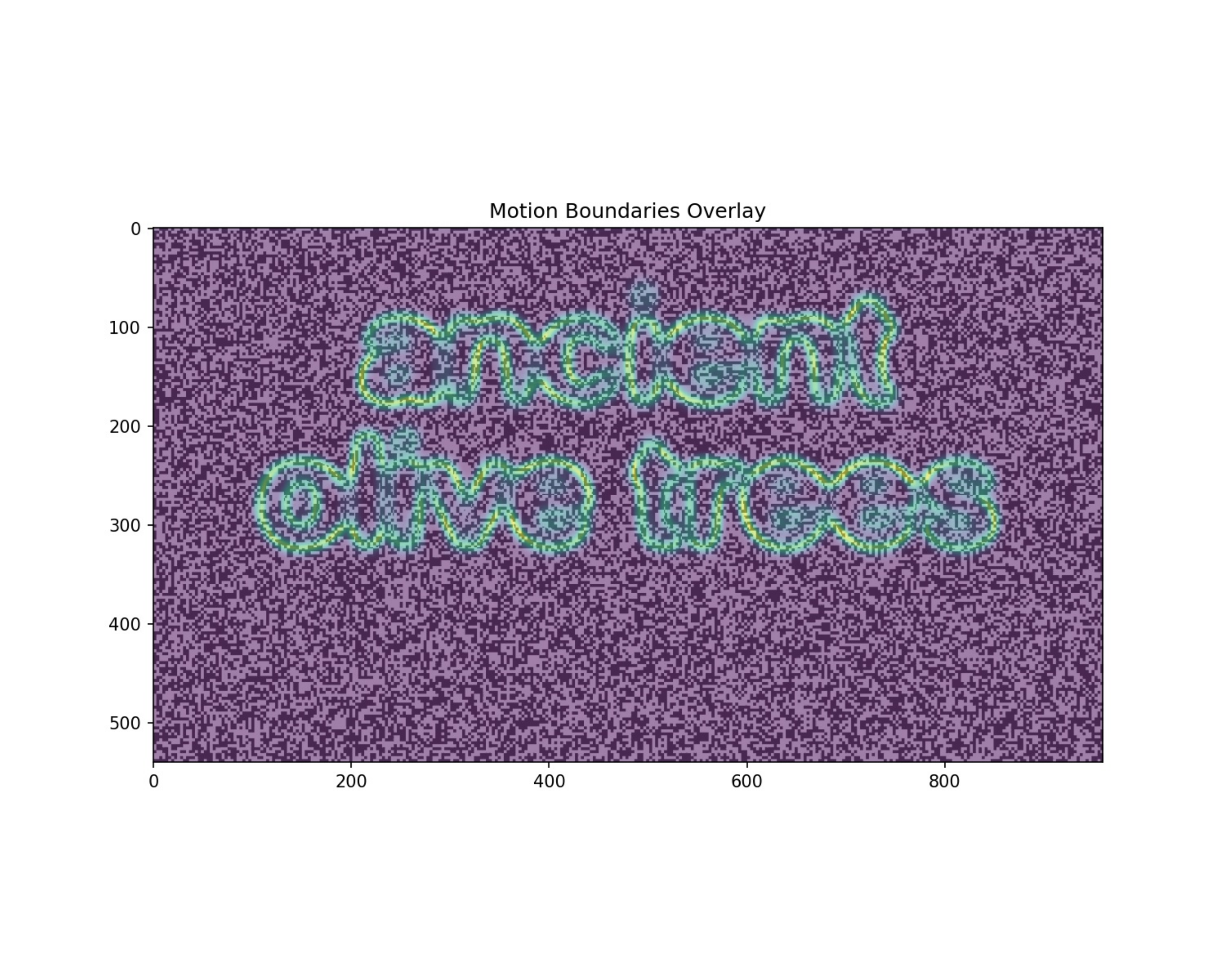} &
        \includegraphics[width=0.22\textwidth]{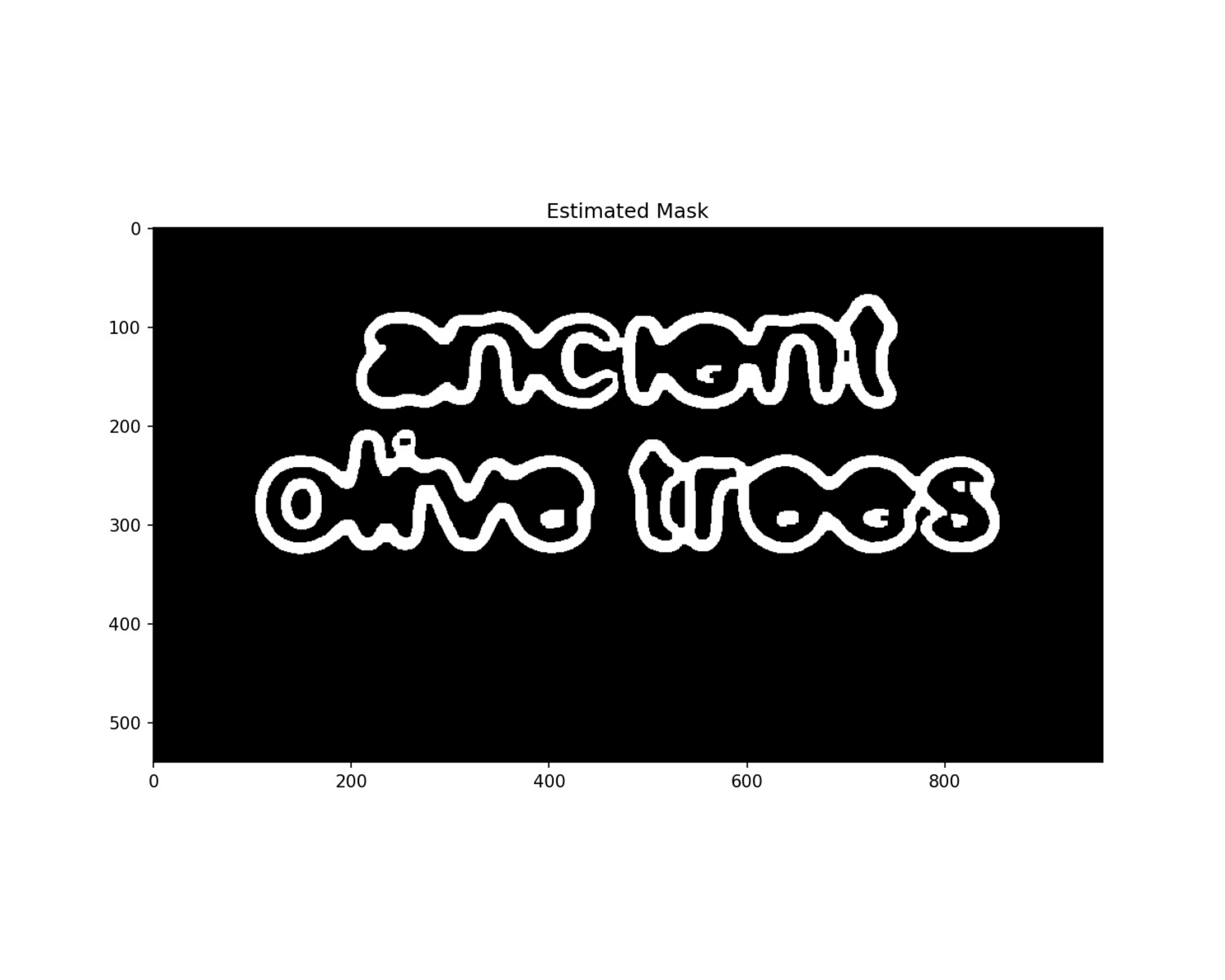} &
        \includegraphics[width=0.22\textwidth]{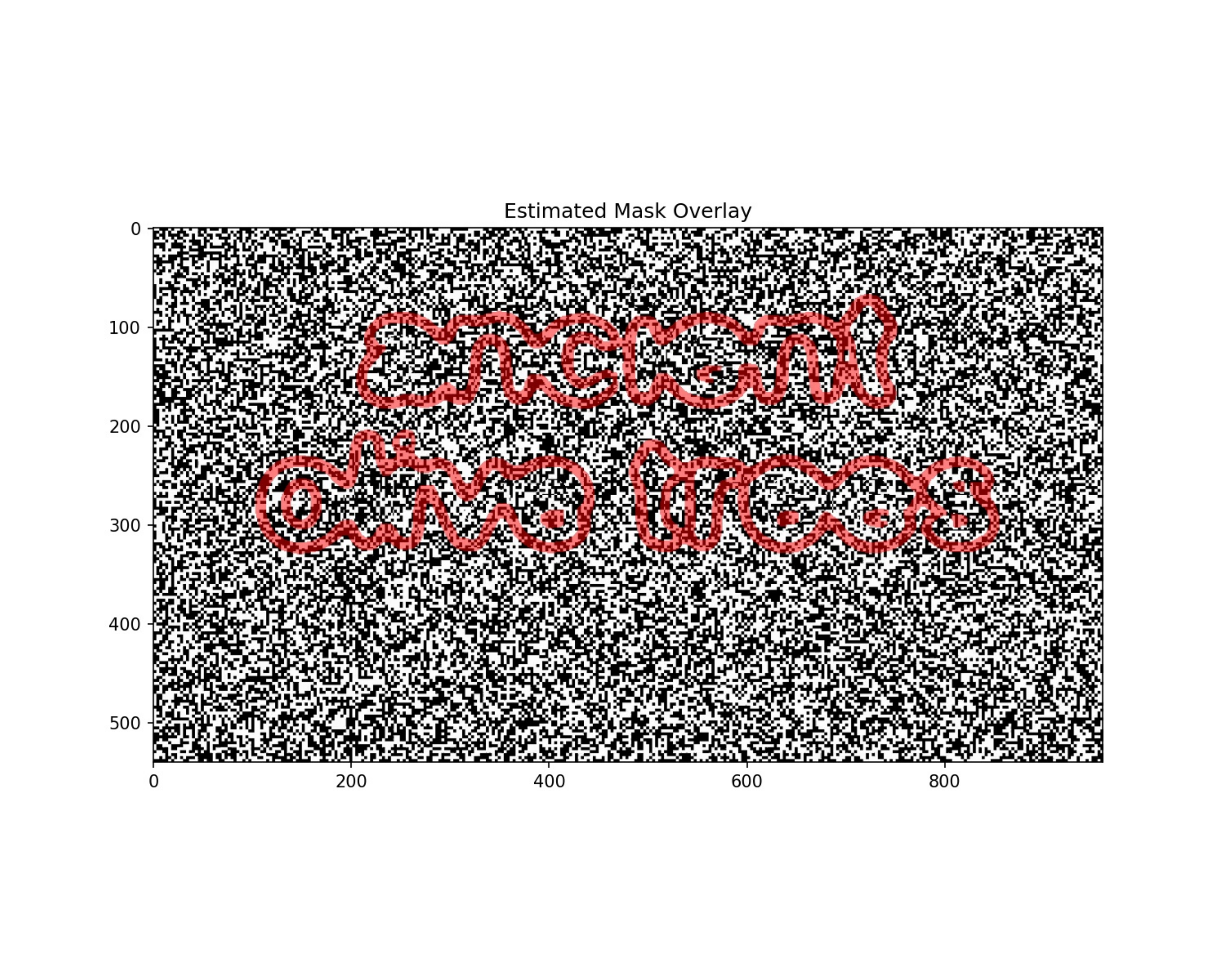} \\
        
        \includegraphics[width=0.22\textwidth]{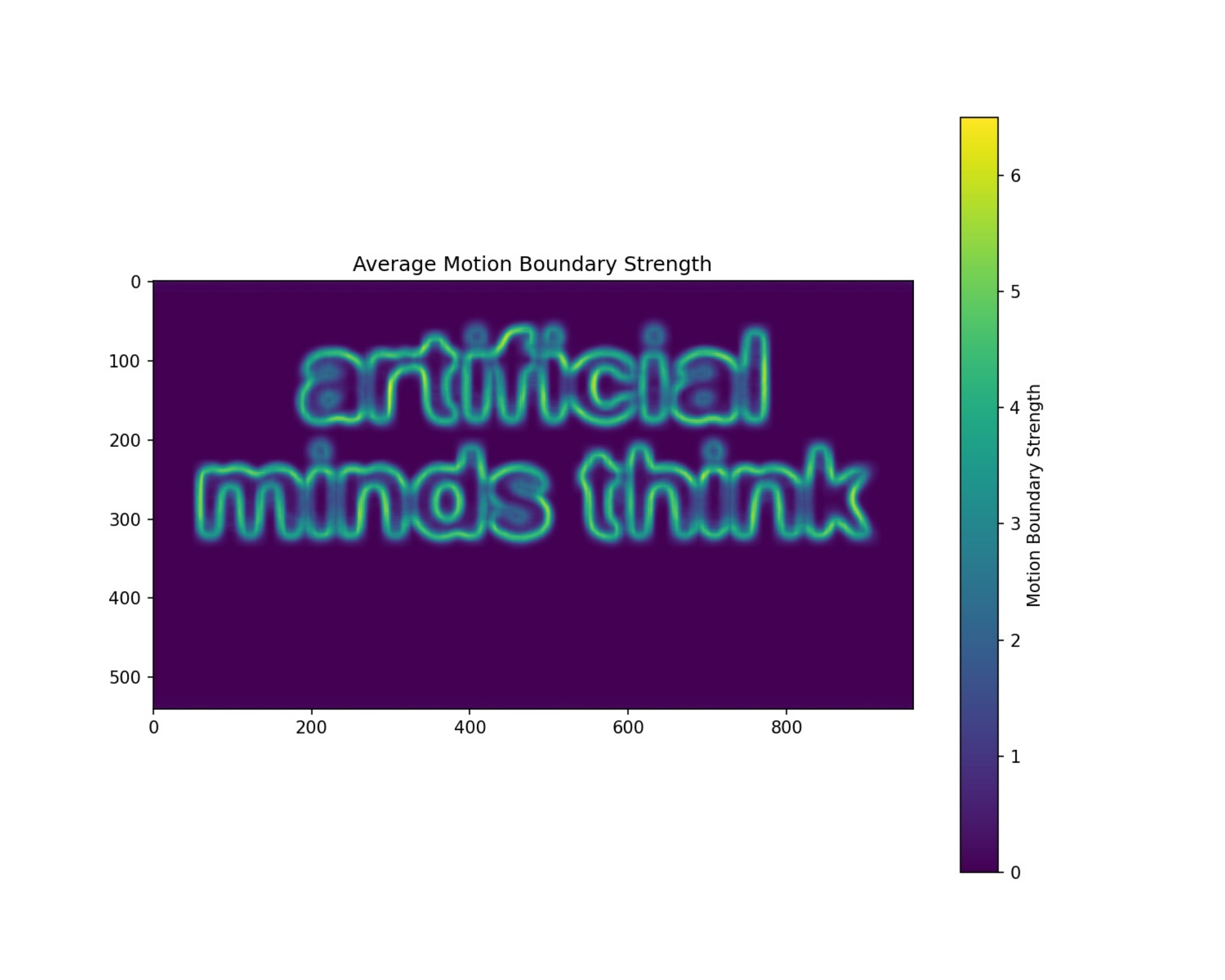} &
        \includegraphics[width=0.22\textwidth]{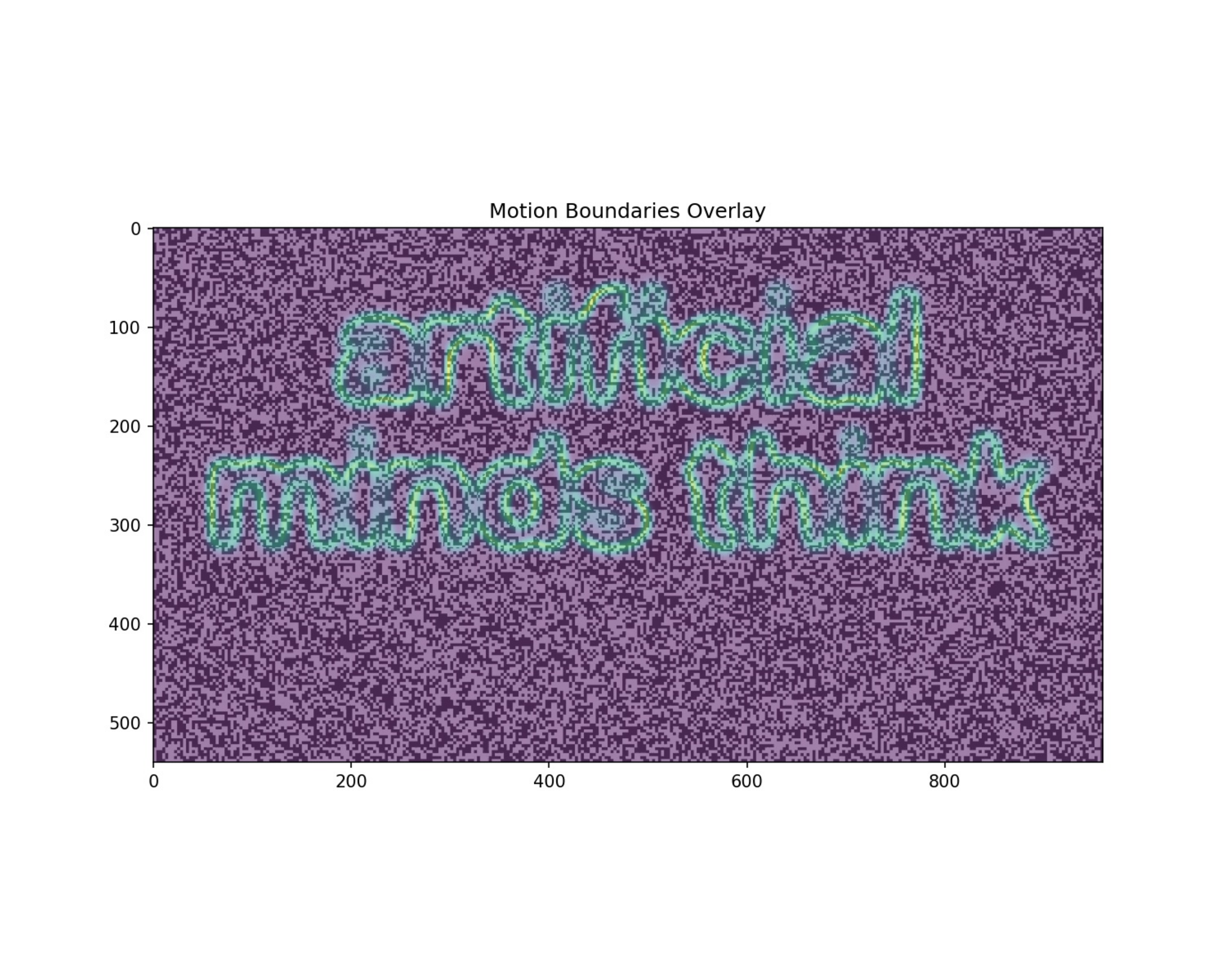} &
        \includegraphics[width=0.22\textwidth]{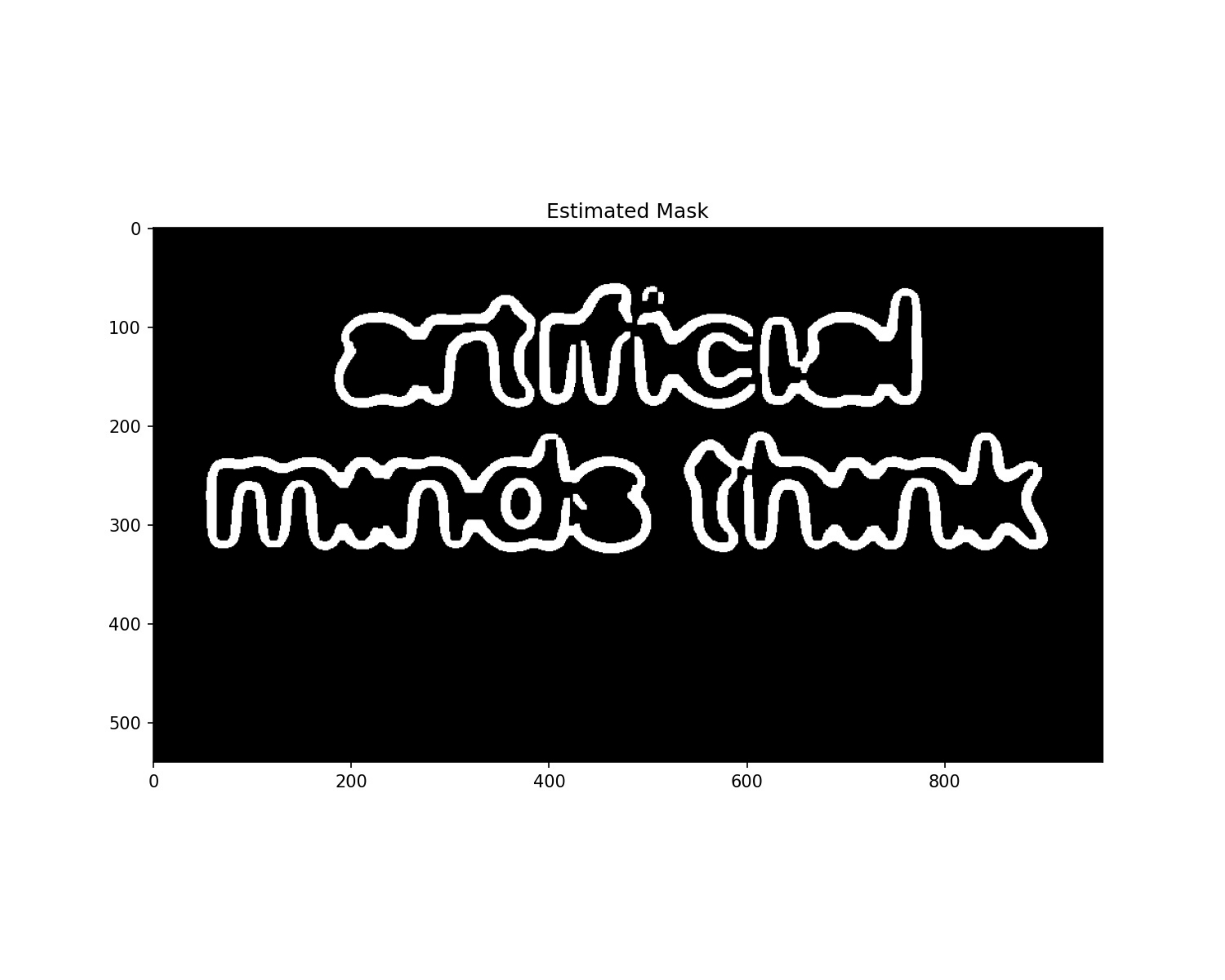} &
        \includegraphics[width=0.22\textwidth]{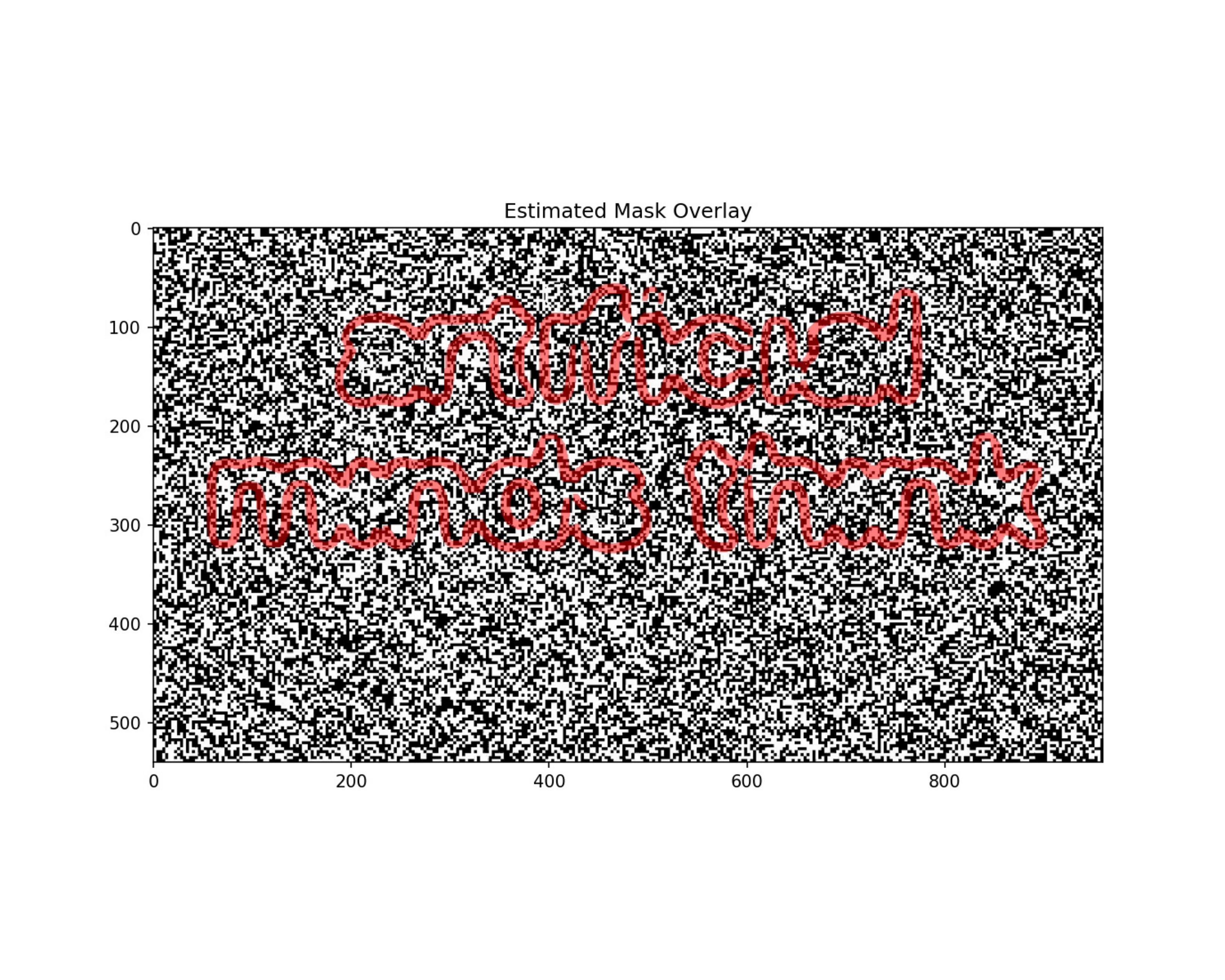} \\
    \end{tabular}
    \caption{Temporal motion coherence analysis for Words category. Each row shows motion boundaries, boundary overlay, estimated mask, and mask overlay for: Gold, Laser Beams Cross, Ancient Olive Trees, and Artificial Minds Think (top to bottom).}
    \label{fig:words-temporal-analysis}
\end{figure*}

\begin{figure*}[t]
    \centering
    \begin{tabular}{@{}c@{\hskip 4pt}c@{\hskip 4pt}c@{\hskip 4pt}c@{}}
        \includegraphics[width=0.22\textwidth]{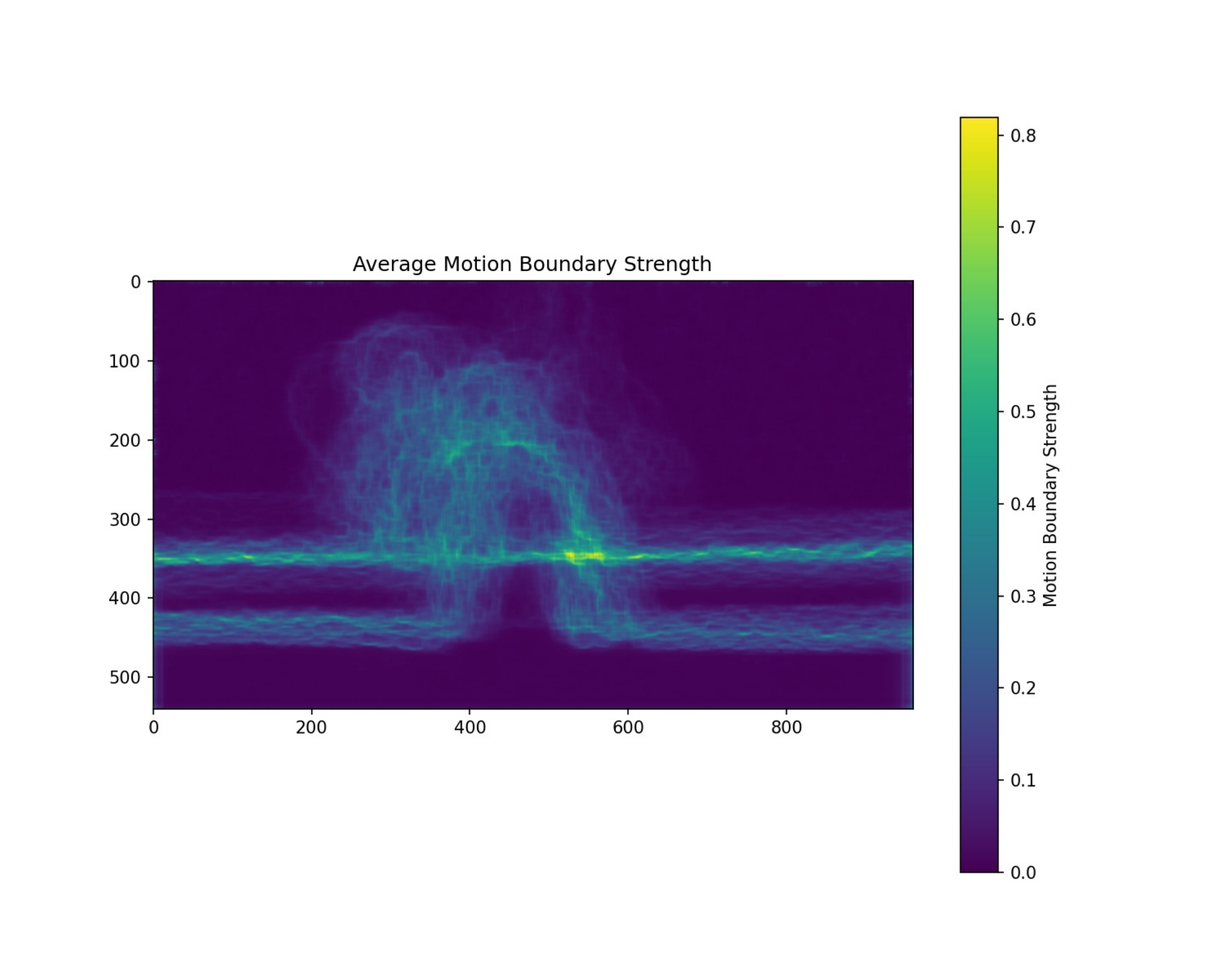} &
        \includegraphics[width=0.22\textwidth]{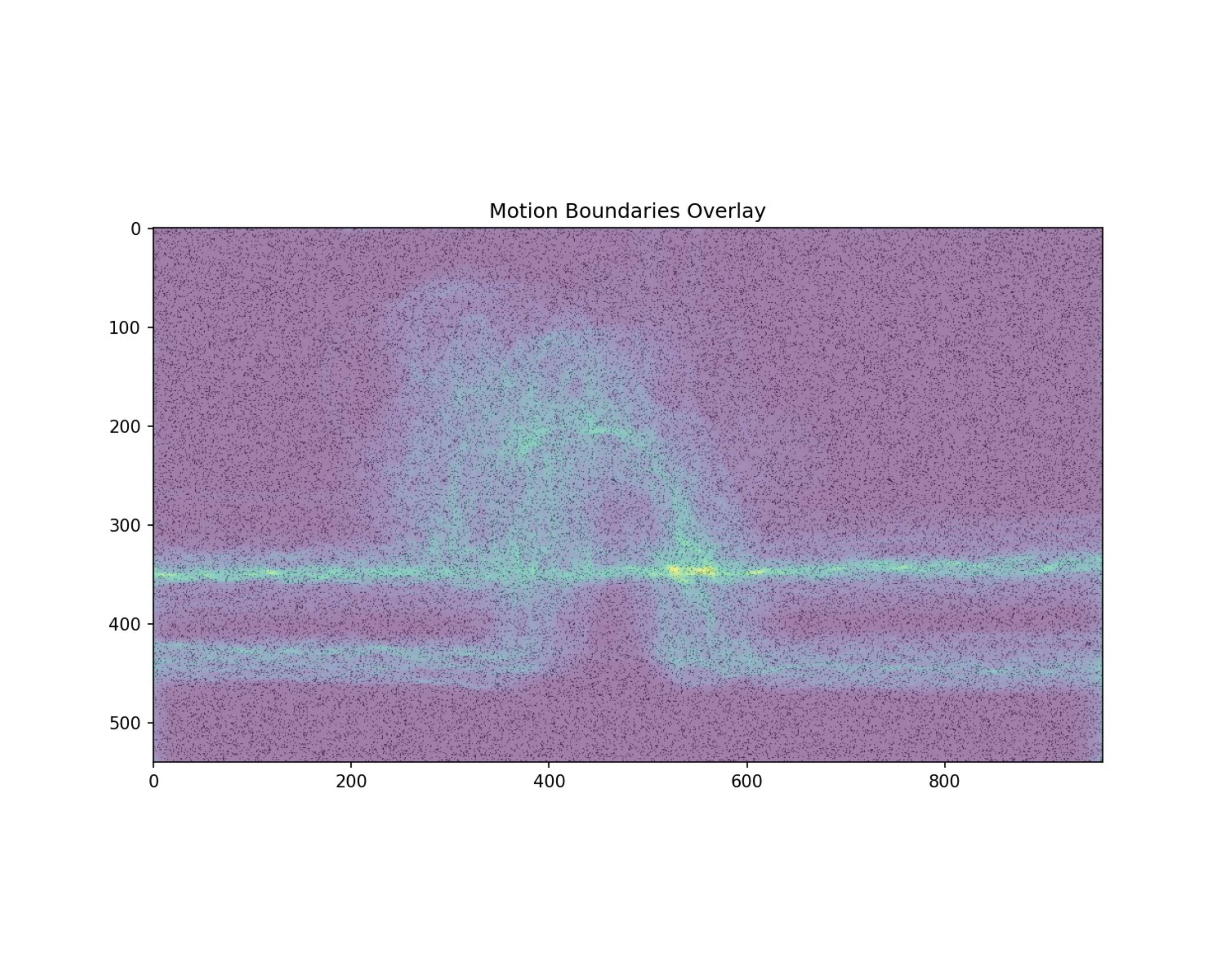} &
        \includegraphics[width=0.22\textwidth]{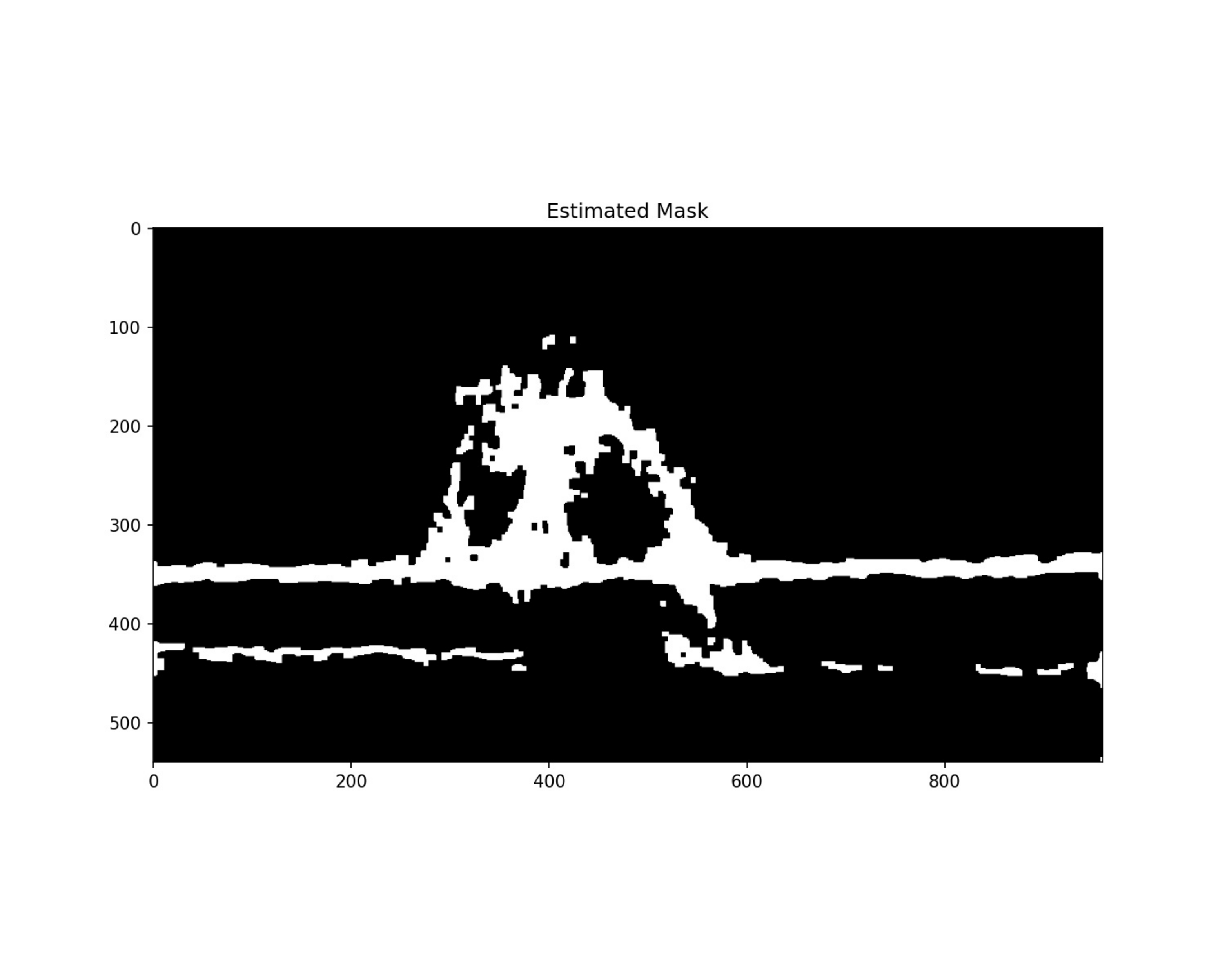} &
        \includegraphics[width=0.22\textwidth]{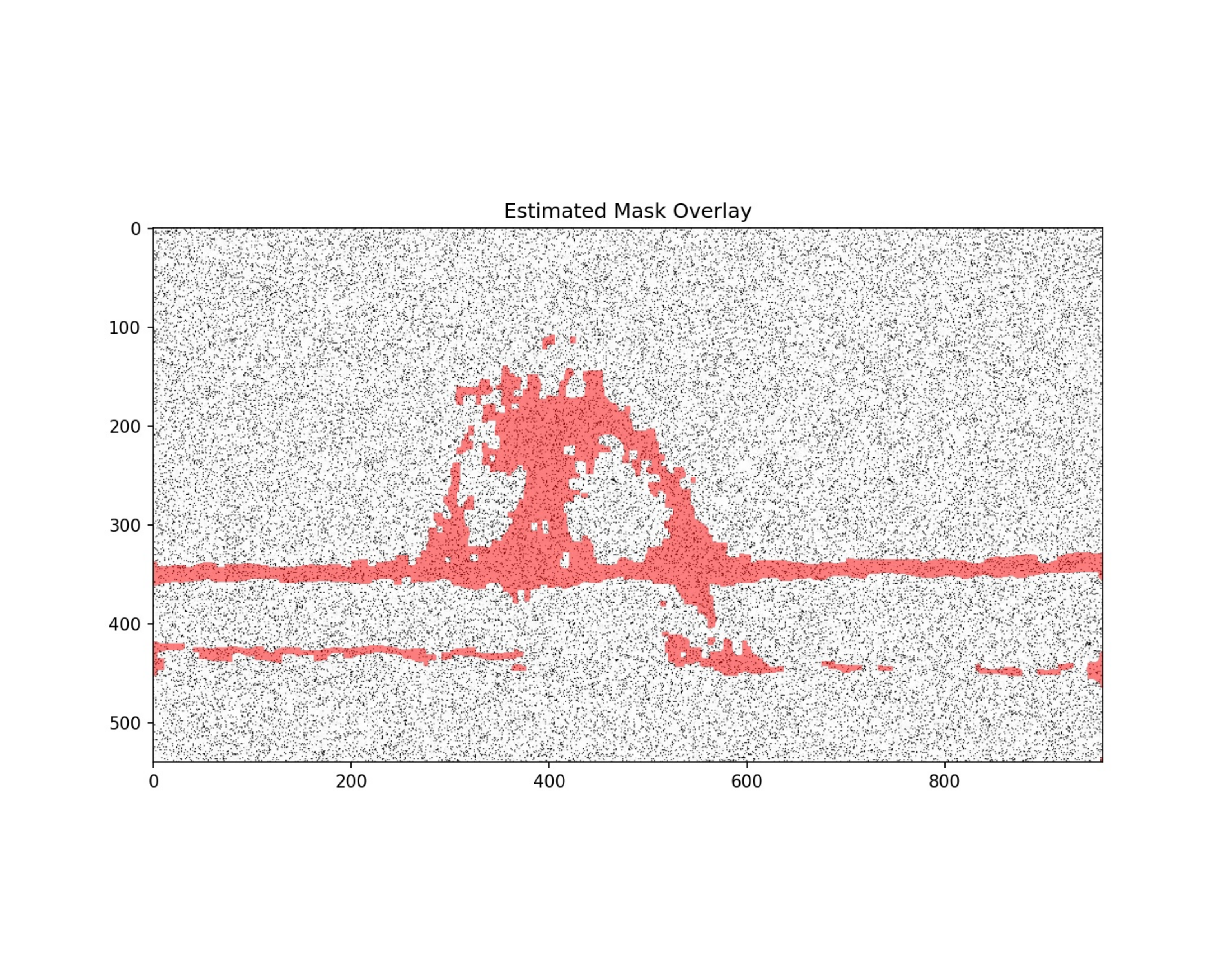} \\
        
        \includegraphics[width=0.22\textwidth]{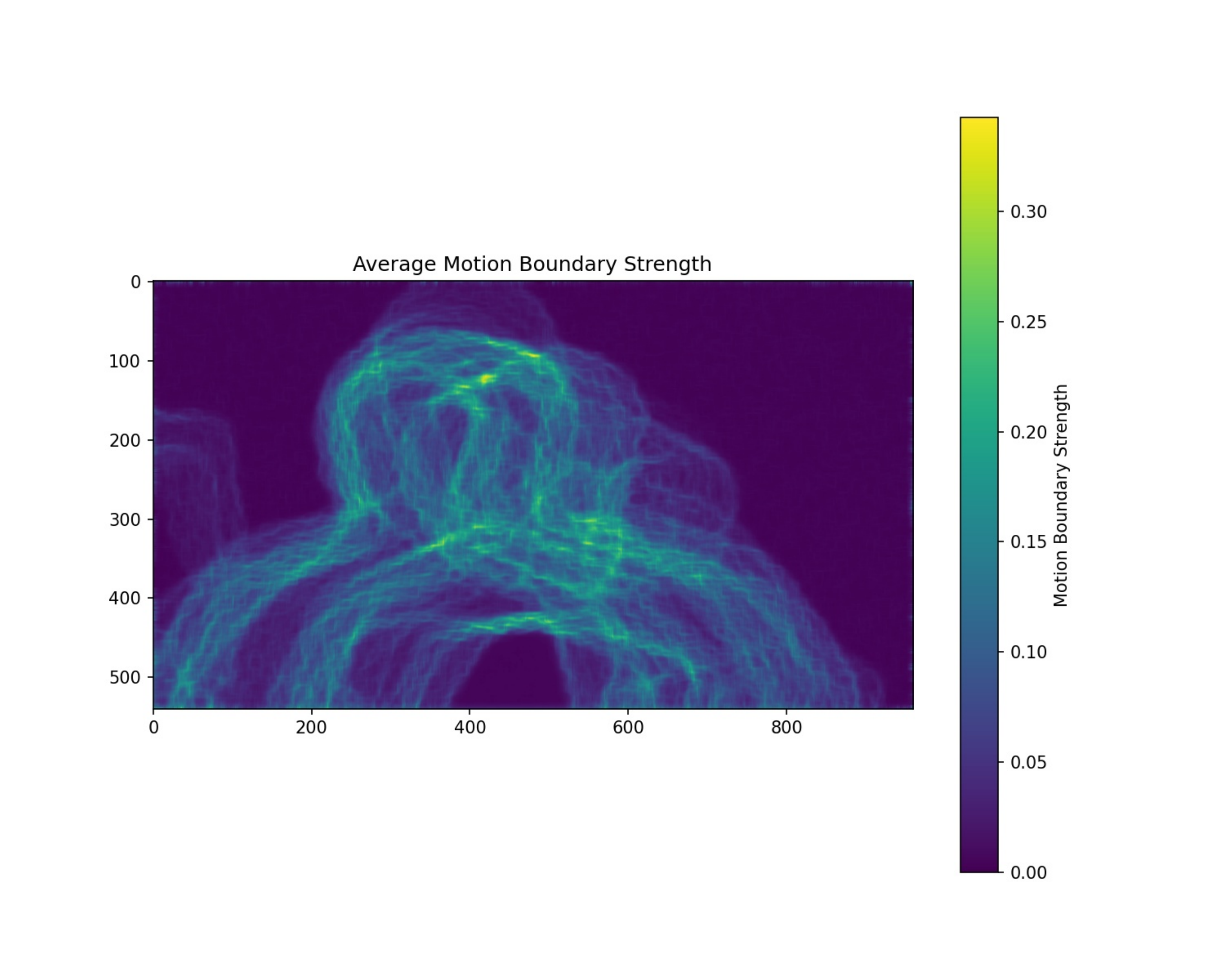} &
        \includegraphics[width=0.22\textwidth]{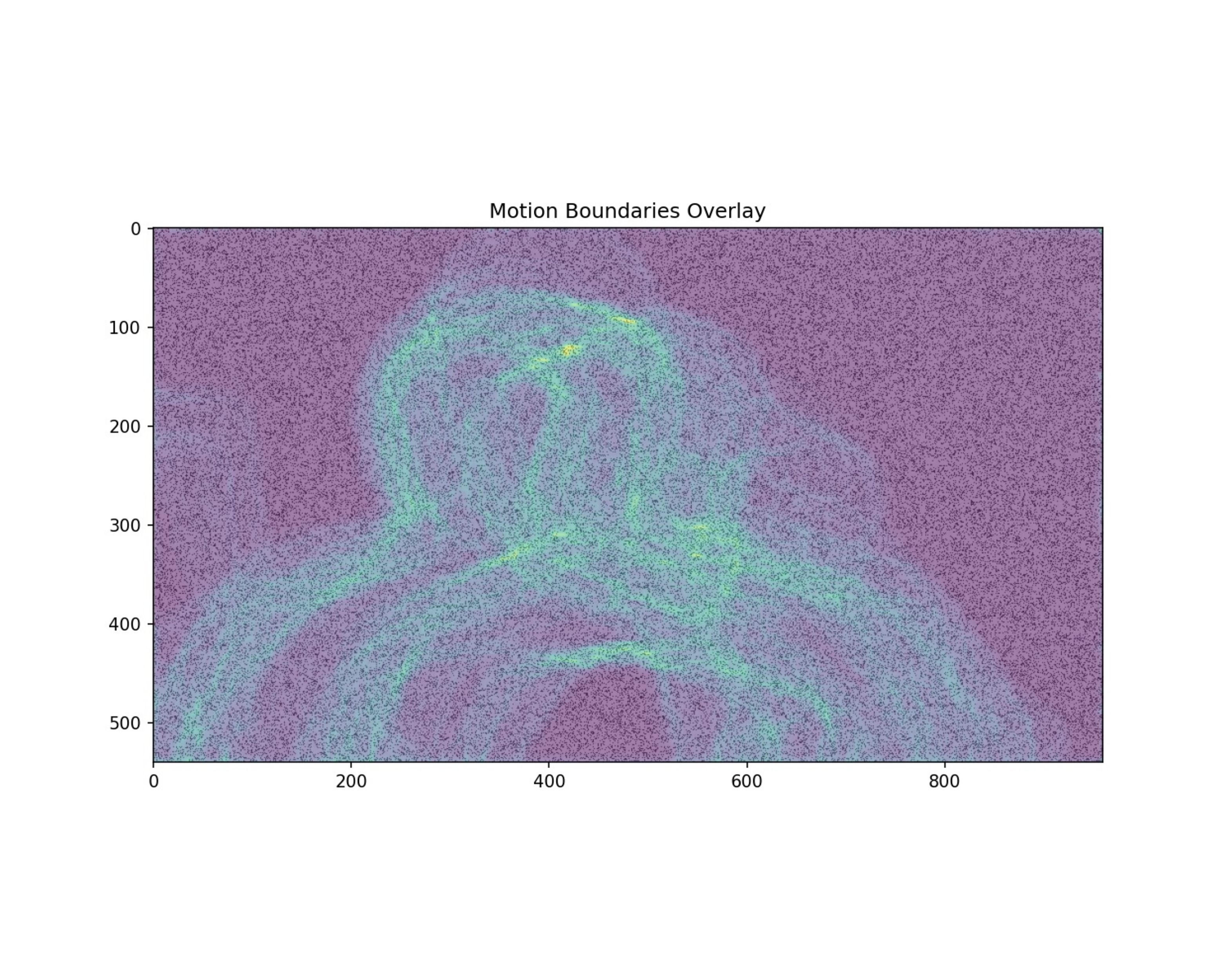} &
        \includegraphics[width=0.22\textwidth]{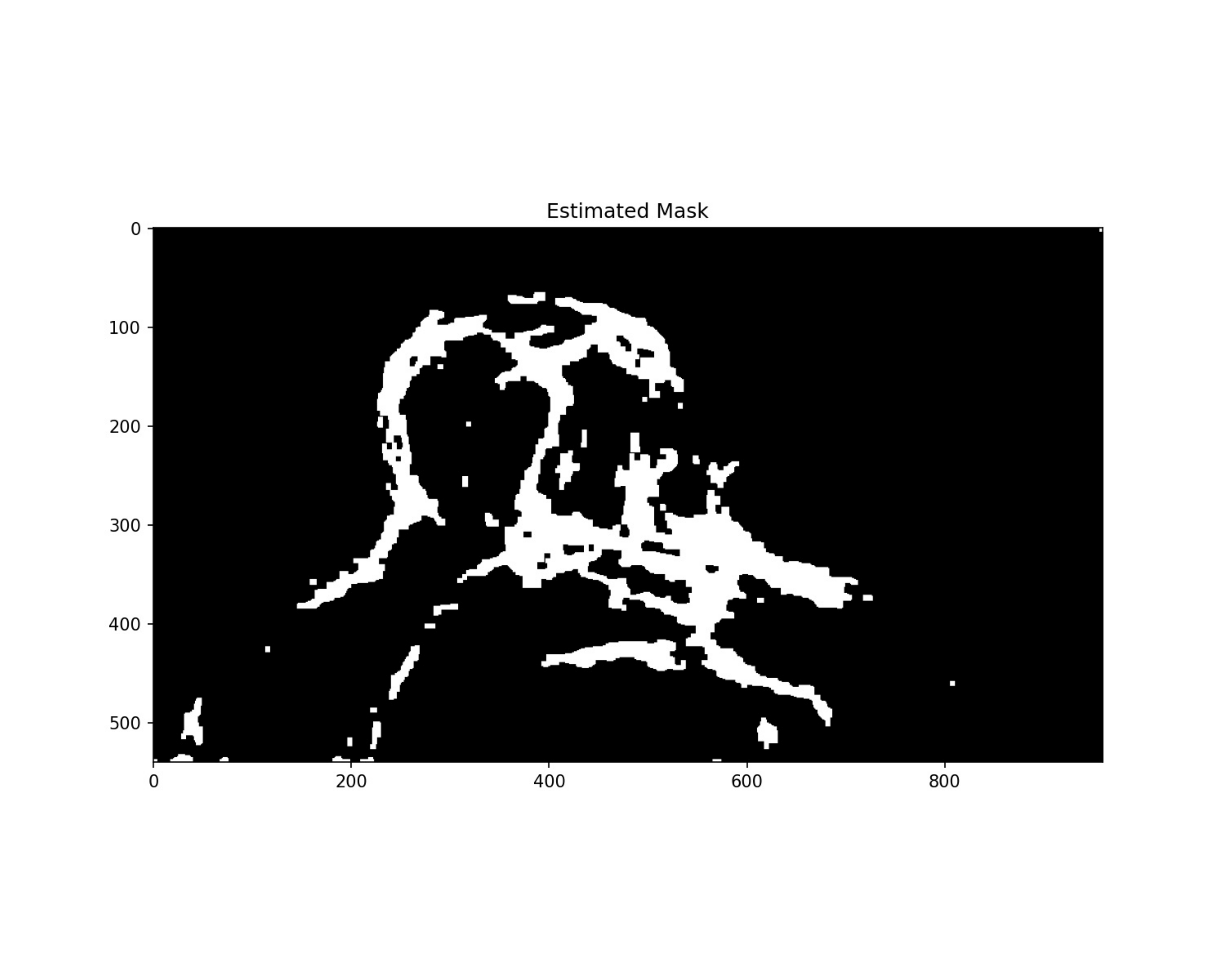} &
        \includegraphics[width=0.22\textwidth]{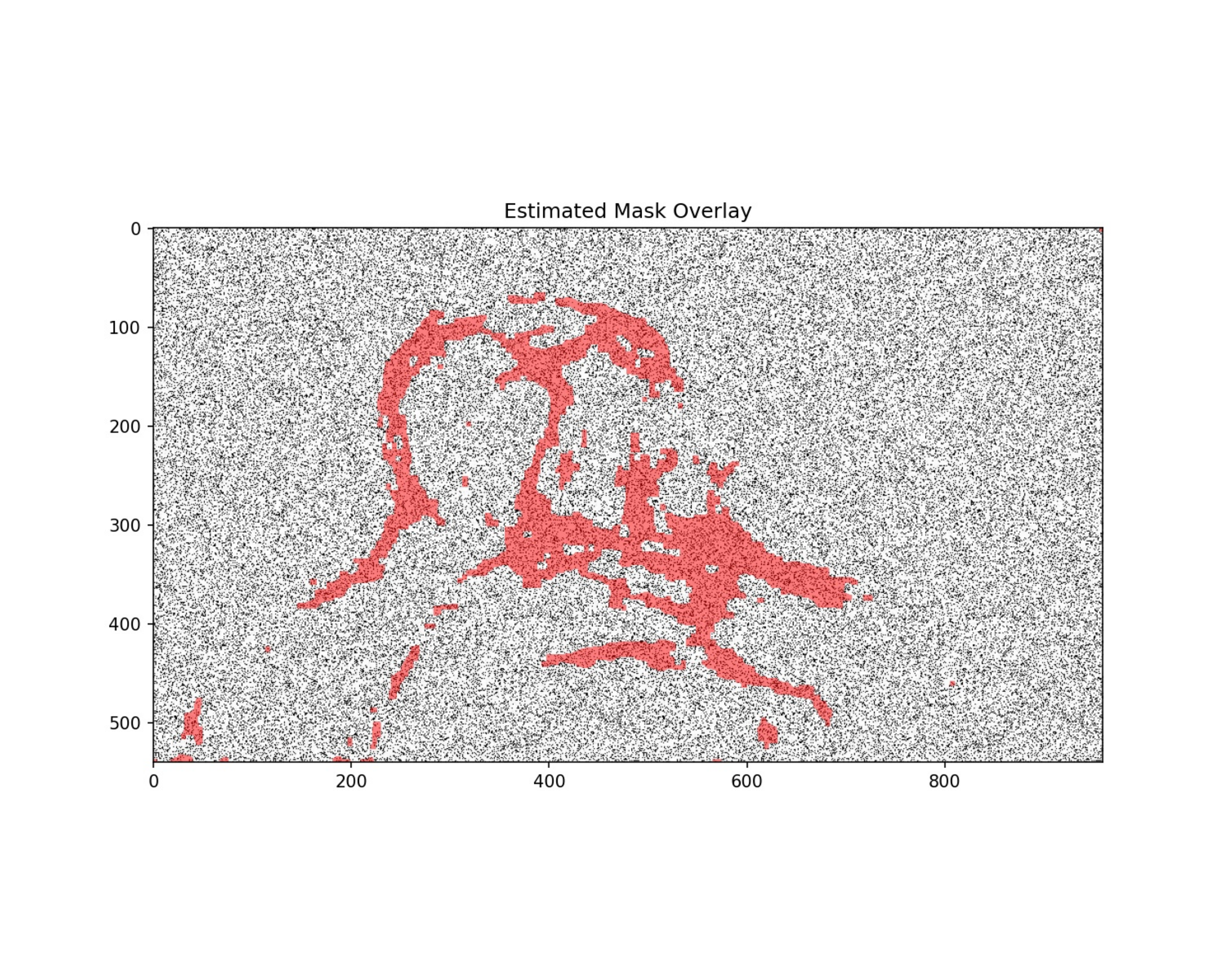} \\
        
        \includegraphics[width=0.22\textwidth]{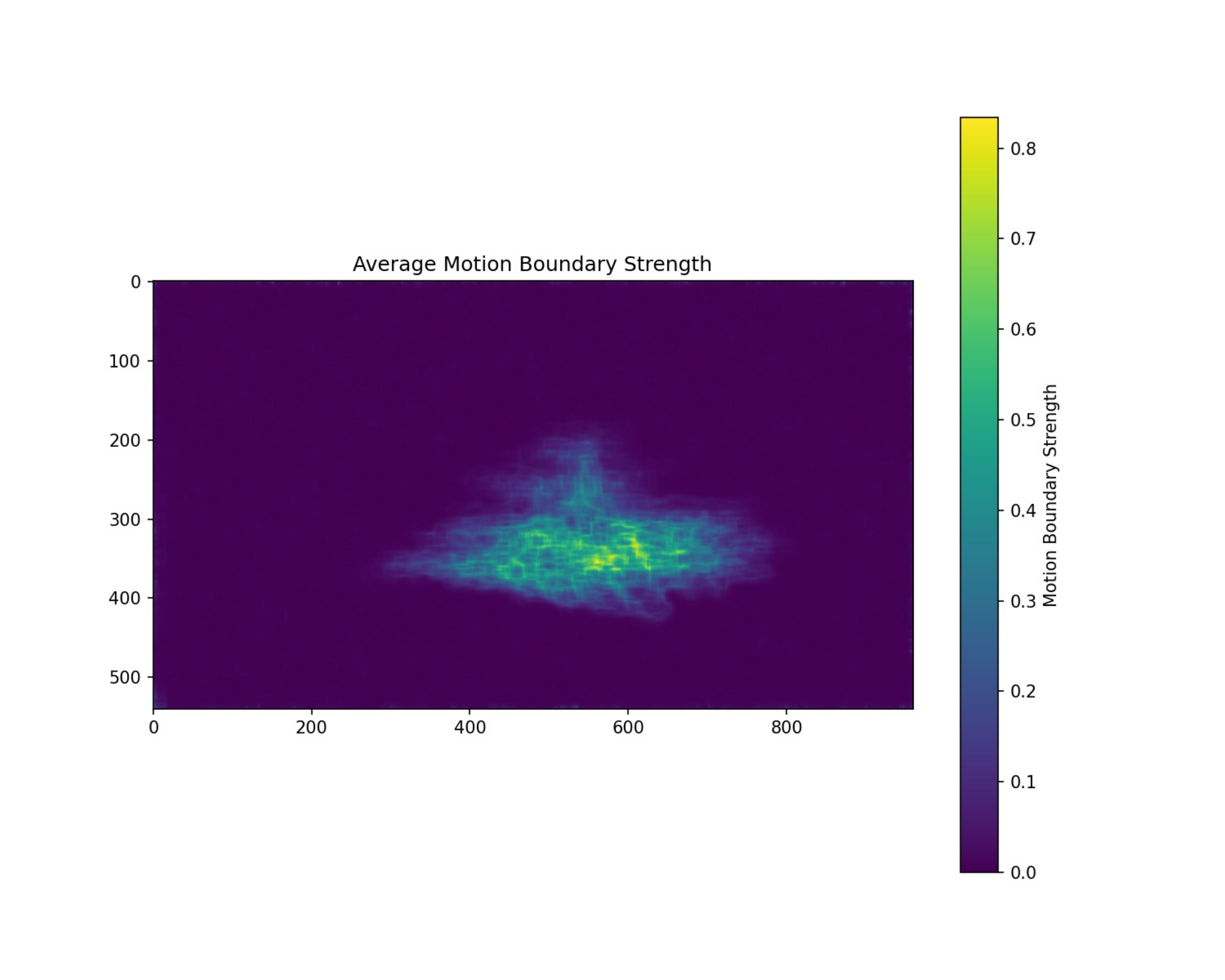} &
        \includegraphics[width=0.22\textwidth]{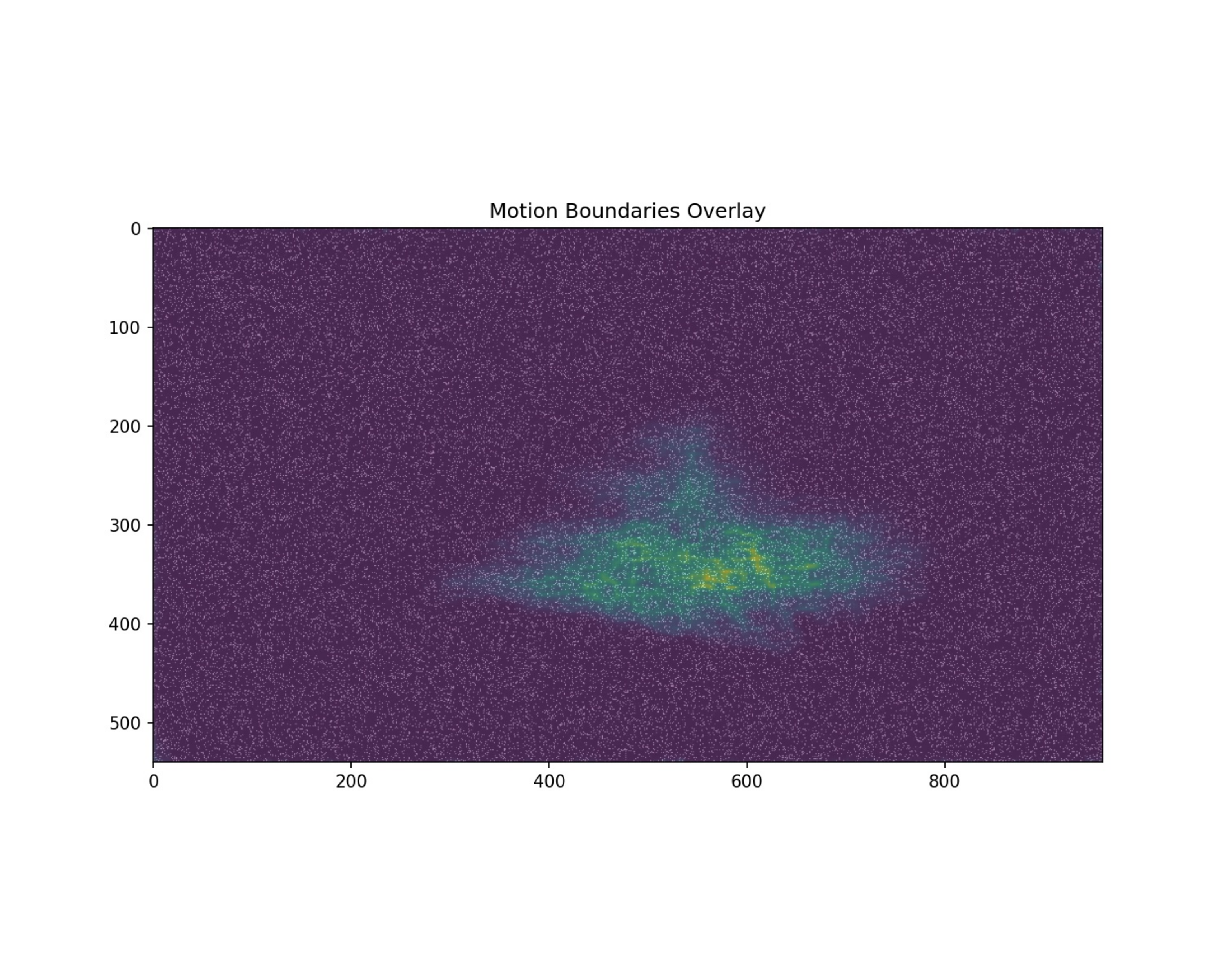} &
        \includegraphics[width=0.22\textwidth]{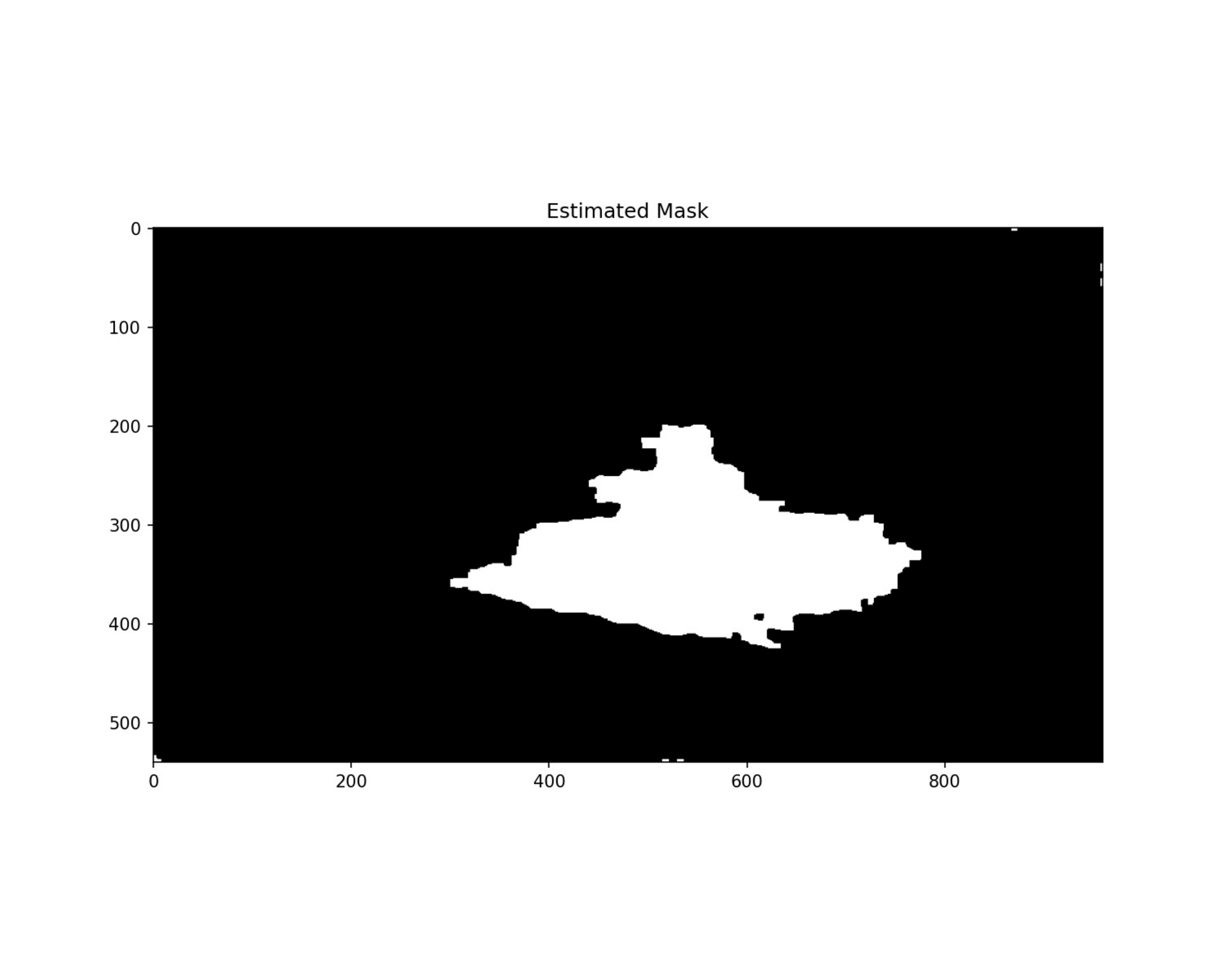} &
        \includegraphics[width=0.22\textwidth]{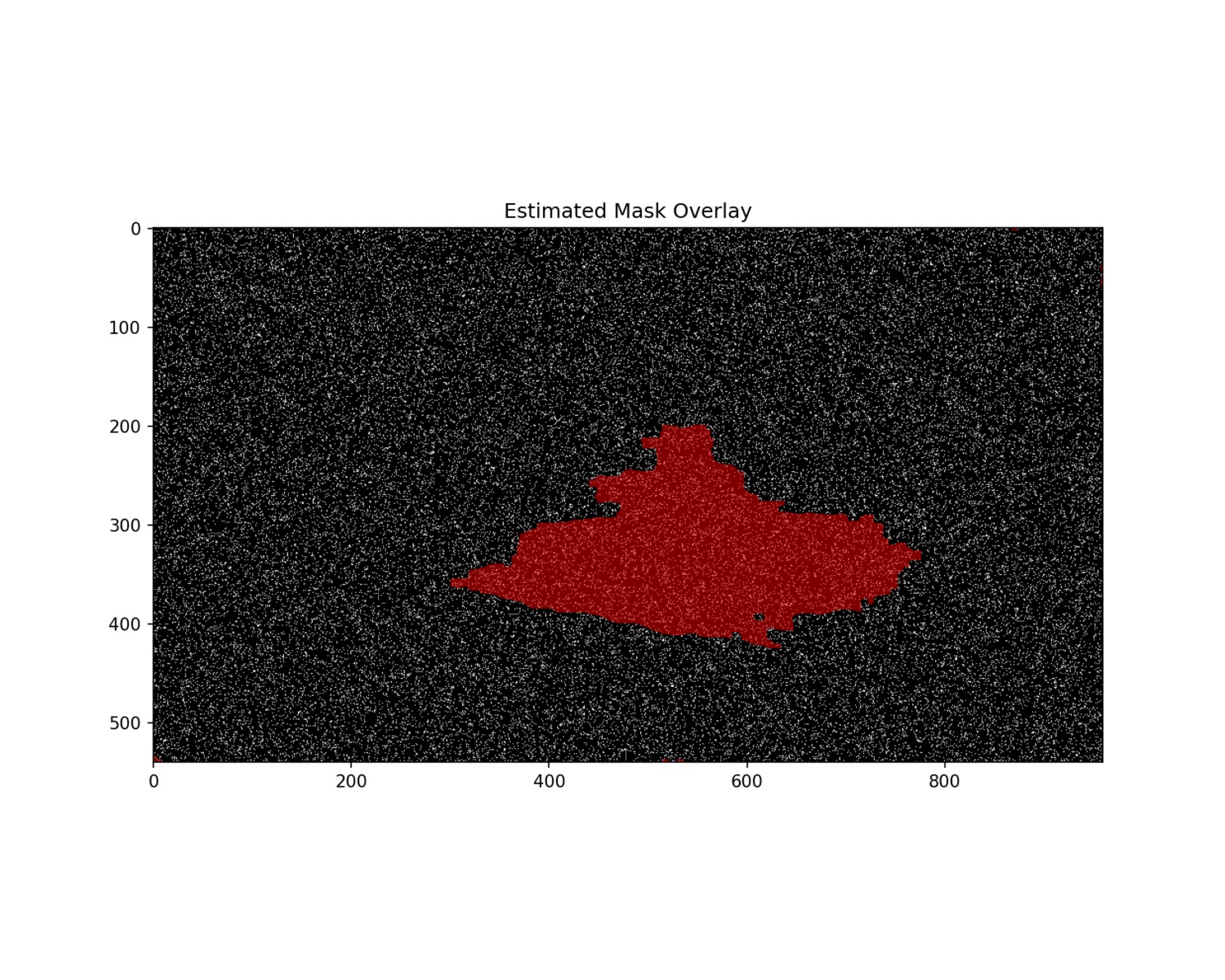} \\
        
        \includegraphics[width=0.22\textwidth]{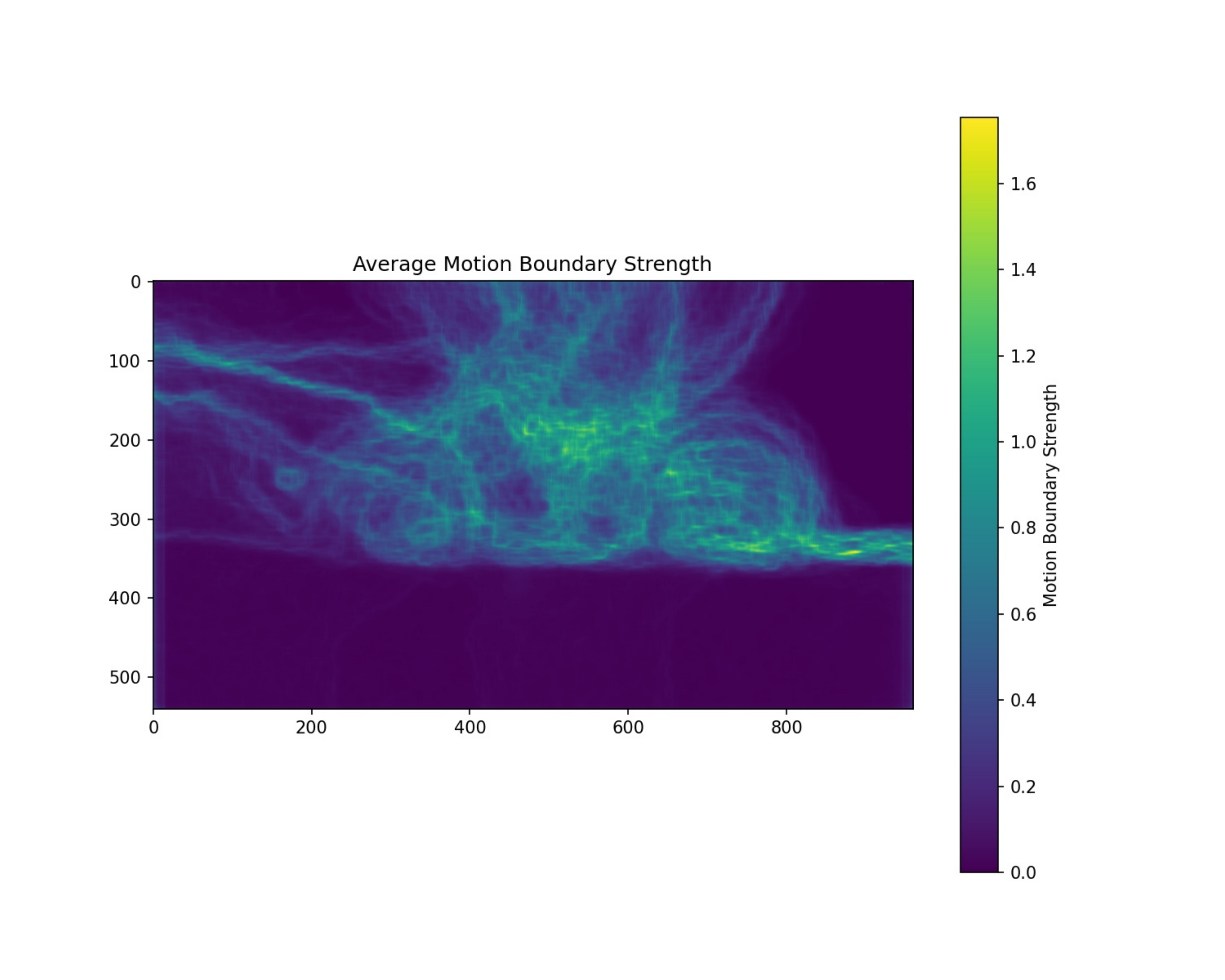} &
        \includegraphics[width=0.22\textwidth]{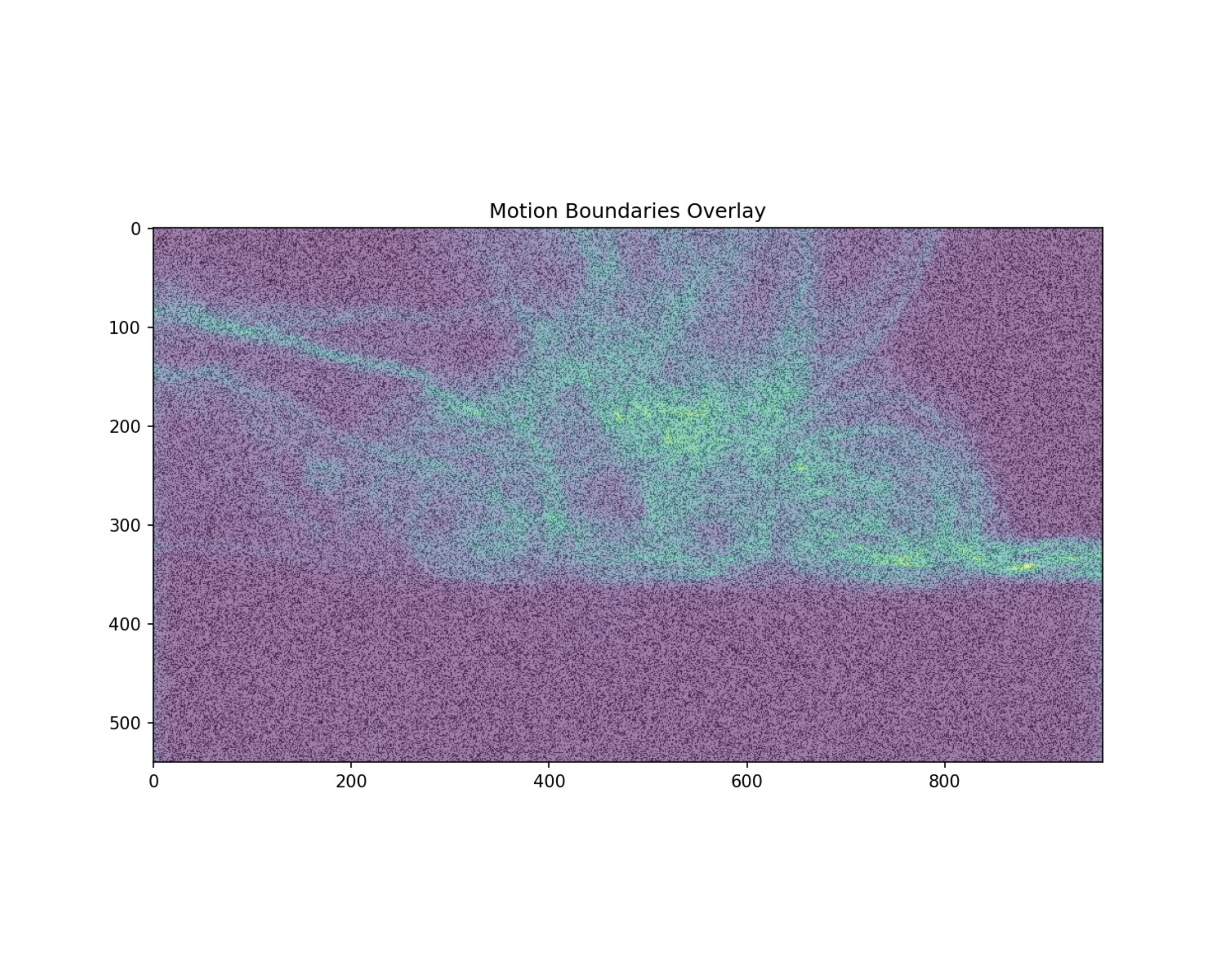} &
        \includegraphics[width=0.22\textwidth]{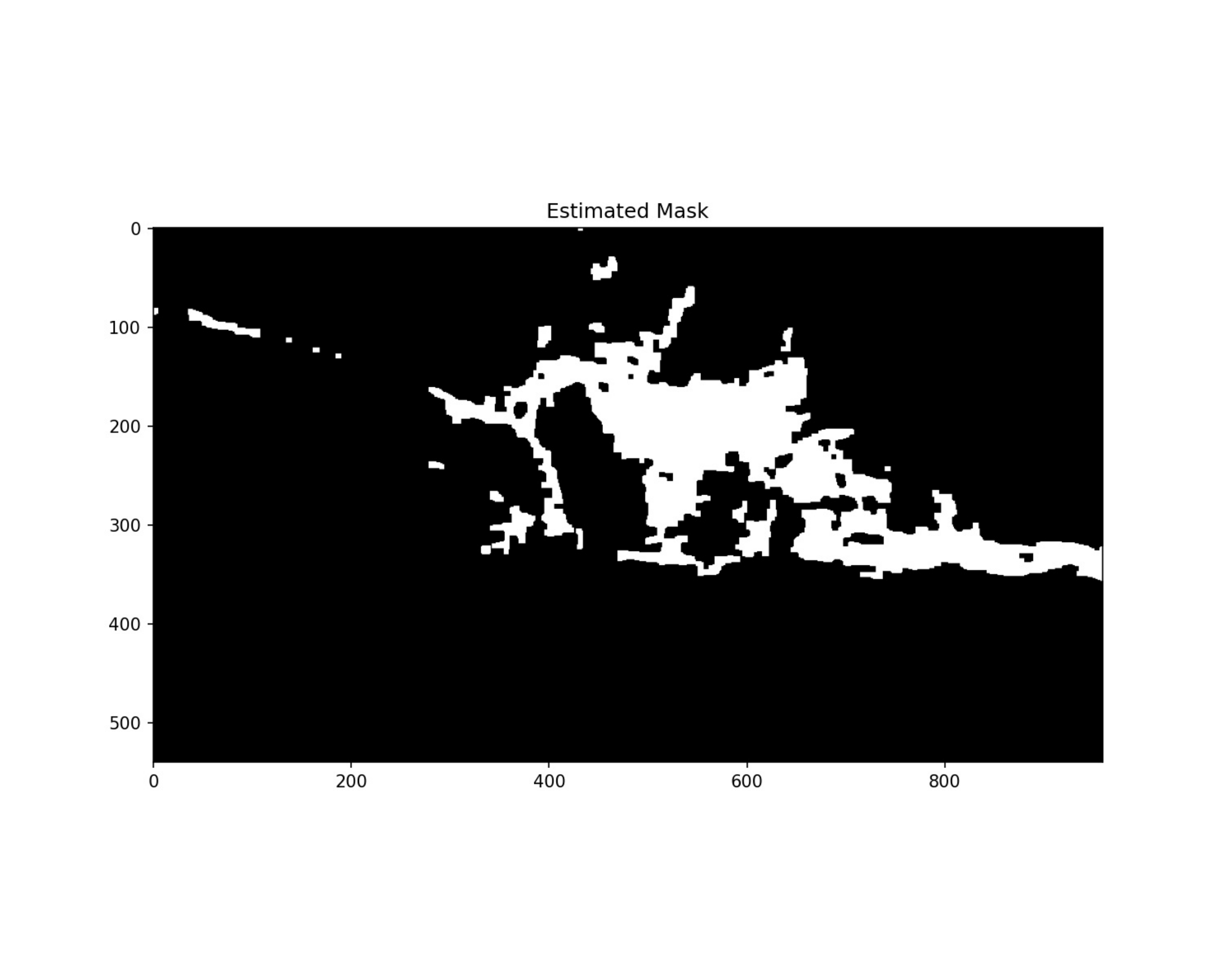} &
        \includegraphics[width=0.22\textwidth]{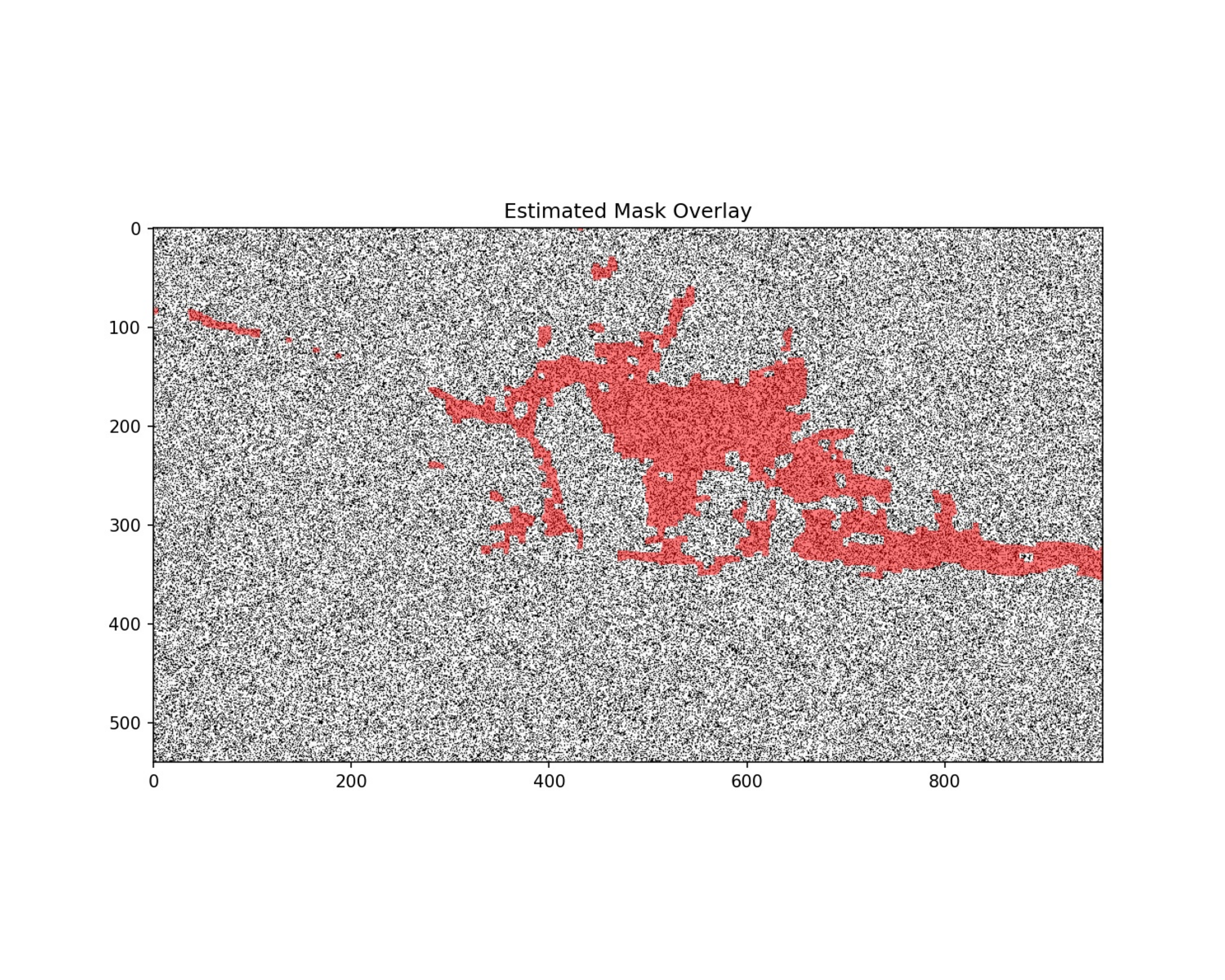} \\
    \end{tabular}
    \caption{Temporal motion coherence analysis for Videos category. Each row shows motion boundaries, boundary overlay, estimated mask, and mask overlay for: Human\_6, Man\_1, Plane\_2, and Bicycle\_10 video sequence (top to bottom).}
    \label{fig:videos-temporal-analysis}
\end{figure*}

We present additional images from the analysis of temporal motion coherence across all categories in SpookyBench. Figures \ref{fig:images-temporal-analysis-1}, \ref{fig:images-temporal-analysis-2}, \ref{fig:words-temporal-analysis}, and \ref{fig:videos-temporal-analysis} show motion boundaries, boundary overlays, estimated masks, and mask overlays for different examples in our dataset, demonstrating the varying effectiveness of temporal integration across different content types. In Figure \ref{fig:videos-temporal-analysis}, the temporal motion coherence method does not perform effectively for dynamic scenes. Since this category contains real-life videos with complex motion patterns, several factors reduce clarity:

\begin{enumerate}
    \item \textbf{Distributed Motion Patterns}: Human movement involves multiple articulated body parts moving in different directions simultaneously, creating competing motion signals that fragment coherent boundaries.
    
    \item \textbf{Non-rigid Deformation}: Dynamic content involves continuous shape changes throughout motion sequences, making consistent boundary extraction more challenging than for static objects.
    
    \item \textbf{Complex Temporal Dynamics}: Mechanical motions in examples like Plane\_2 and Bicycle\_10 create temporal discontinuities that disrupt the motion coherence essential for shape perception.
\end{enumerate}

\clearpage
{
    \small
    \bibliographystyle{ieeenat_fullname}
    \bibliography{main}
}